%% file: ms.tex
\documentclass[times,twocolumn,final]{elsarticle}
\pdfoutput=1
\usepackage{framed,multirow}

\usepackage{amsmath,amssymb}
\usepackage{latexsym}
\usepackage[textfont=rm]{subcaption}

\usepackage{url}
\usepackage{xcolor}
\definecolor{newcolor}{rgb}{.8,.349,.1}

\usepackage{hyperref}

\usepackage[switch,pagewise]{lineno}

\usepackage{algorithm}
\usepackage{algpseudocode}
\usepackage{listings}

\DeclareMathOperator{\arctantwo}{arctan2}

\begin{document}

\begin{frontmatter}

\title{Neural inverse procedural modeling of knitting yarns from images}%

\author[1]{Elena Trunz}
\ead{trunz@cs.uni-bonn.de}
\author[1,2]{Jonathan Klein} 
\author[1]{Jan Müller} 
\author[1]{Lukas Bode} 
\author[1]{Ralf Sarlette}
\author[3]{Michael Weinmann\corref{cor1}}
\ead{M.Weinmann@tudelft.nl}
\author[1]{Reinhard Klein} 

\cortext[cor1]{Corresponding author}
\address[1]{University of Bonn, Germany}
\address[2]{KAUST, Saudi Arabia}
\address[3]{Delft University of Technology, Netherlands}

\begin{abstract}
We investigate the capabilities of neural inverse procedural modeling to infer high-quality procedural yarn models with fiber-level details from single images of depicted yarn samples.
While directly inferring all parameters of the underlying yarn model based on a single neural network may seem an intuitive choice, we show that the complexity of yarn structures in terms of twisting and migration characteristics of the involved fibers can be better encountered in terms of ensembles of networks that focus on individual characteristics.
We analyze the effect of different loss functions including a parameter loss to penalize the deviation of inferred parameters to ground truth annotations, a reconstruction loss to enforce similar statistics of the image generated for the estimated parameters in comparison to training images as well as an additional regularization term to explicitly penalize deviations between latent codes of synthetic images and the average latent code of real images in the encoder’s latent space.
We demonstrate that the combination of a carefully designed parametric, procedural yarn model with respective network ensembles as well as loss functions even allows robust parameter inference when solely trained on synthetic data.
Since our approach relies on the availability of a yarn database with parameter annotations and we are not aware of such a respectively available dataset, we additionally provide, to the best of our knowledge, the first dataset of yarn images with annotations regarding the respective yarn parameters. For this purpose, we use a novel yarn generator that improves the realism of the produced results over previous approaches.
\end{abstract}


\end{frontmatter}

%\linenumbers

\input{introduction.tex}
\input{related_work.tex}
\input{modeling.tex}
\input{fitting.tex}
\input{results.tex}
\input{conclusions.tex}

 \appendix
\input{supplemental.tex}

\end{document}

%% file: introduction.tex
\section{Introduction}
Due to their ubiquitous presence, fabrics have a great importance in domains like entertainment, advertisement, fashion and design.
In the era of digitization, numerous applications rely on virtual design and modeling of fabrics and cloths.
Besides the use of fabrics in games and movies, further examples include online retail with its focus on more accurately depicting the appearance of the respective clothes in images, videos or even virtual try-on solutions, as well as virtual prototyping and advertisement applications to provide pre-views on respective product designs.\\
\indent The accurate digital reproduction of the appearance of fabrics and cloth relies on a fiber-level based modeling to allow accurately representing light exchange in the fiber and yarn levels.
However, due to their structural and optical complexity imposed by the arrangement of fibers with diverse characteristics within yarns and the interaction between yarns in the scope of weave and knitting patterns -- where small changes in the fiber and yarn arrangement may result in significant appearance variations -- as well as due to the numerous partial occlusions of the involved fibers and yarns, capturing and modeling the appearance of yarns, fabrics and cloth remains a challenge.
In the context of reconstructing yarns, Zhao et al.~\cite{zhao2016fitting} addressed the difficulty of scanning the self-occluding fiber arrangements based on computer-tomography (CT) scans to get accurate 3D reconstructions of the individual yarns.
However, this imposes the need for special hardware.
Instead, in this paper, we aim at the capture and modeling of the appearance of yarns by inferring individual yarn parameters from a single photograph depicting a small part of a yarn.\\
\indent To address this goal, we investigate the capabilities of neural inverse procedural modeling.
Whereas directly optimizing all the parameters that determine a yarn's geometry (including flyaways i.e. fibers that migrate from the yarn, contributing to the fuzziness of the yarn) with a single neural network may seem an intuitive choice, the complexity of the depictions of yarns, where twisting characteristics dominate the appearance in the yarn center and flyaway statistics dominate the appearance in the yarns' border regions, imposes that the network has the capacity to understand where which parameters can be predominantly inferred from.
This observation might indicate that other strategies such as training separate networks for inferring the structural parameters for the main yarn and the characteristics of flyaways or even using an ensemble of networks, where each of these networks is only responsible for estimating a single parameter of the underlying yarn model, could be reasonable alternatives.
Therefore, we investigate the potential of these approaches for the task of inverse yarn modeling from a single image.
Furthermore, we investigate the effect of different loss functions including a parameter loss to penalize the deviation of inferred parameters to ground truth annotations, a reconstruction loss to enforce similar statistics of the image generated for the estimated parameters in comparison to training images as well as an additional regularization term
to explicitly penalize deviations between latent codes of synthetic images and the average latent code of real images in the encoder’s latent space. Thereby, we also analyze to what extent such models can be trained from solely using synthetic training data.\\
\indent All of these models are trained based on synthetic training data generated from a high-quality yarn simulator that improves upon the generator by Zhao et al.~\cite{zhao2016fitting} in terms of a more realistic modeling of hair flyaways, fiber cross-section characteristics and the orientation of the fibers' twisting axis.
As our approach relies on the availability of a dataset of yarn images with respective annotations regarding characteristic yarn parameters, such as the number of plys, the twisting length etc., we introduce -- to the best of our knowledge -- the first dataset of synthetic yarns with respective yarn parameter annotations. Both the dataset and the yarn generator used for the automatic generation of this dataset will be released upon acceptance of the paper.
Our approach for neural inverse procedural modeling of yarns exhibits robustness to 
variations in appearance induced by varying capture conditions such as different exposure times as long as strong over-exposure and under-exposure are avoided during capture.\\
\indent In summary, the key contributions of our work are:
\begin{itemize}
  \item We present a novel neural inverse modeling approach that allows the inference of accurate yarn parameters including flyaways from a single image of a small part of a yarn.
	\item We investigate the effect of different loss formulations on the performance based on different configurations of a (yarn) parameter loss to penalize deviations in the inferred parameters with respect to the ground truth, a reconstruction loss to enforce the statistics of a rendering with the estimated parameters to match the statistics of given images, and regularization term to explicitly penalize deviations between latent codes of synthetic images and the average latent code of real images in the encoder’s latent space.
	\item We provide, to the best of our knowledge, the first dataset of realistic synthetic yarn images with annotations regarding the respective yarn parameters.
	\item We present a yarn generator that supports a large range of input parameters as well as a yarn sampler that guides the selection of parameter configurations for the automatic generation of realistic yarns.
\end{itemize}

%% file: related_work.tex
\section{Related Work}\label{sec:relatedwork}
Respective surveys~\cite{Schroeder_SA2012Course,castillo2017challenges,castillo2019recent,amor2021classification,pagan2020silk,noor2021review,mohammadi2021smart} indicate the opportunities of computational approaches for the cloth and apparel industries as well as challenges regarding the capture, modeling, representation and analysis of cloth. 
Some approaches approximate fabrics as 2D sheets. Wang et al.~\cite{wang2008modeling} and Dong et al.~\cite{dong2010manifold} leverage spatially varying BRDF (SVBRDFs) based on tabulated normal distributions to represent the appearance of captured materials including embroidered silk satin, whereas others focused on appearance modeling in terms of bidirectional texture functions (BTFs)~\cite{dana1997reflectance,weinmann2014material,filip18evaluating}.

For scenarios with a focus on efficient simulation and editing or respective manipulation, yarn-based models~\cite{kaldor2008simulating,cirio2014yarn,martin2018toward} have been shown to be more amenable~\cite{yuksel2012stitch}, however, at the cost of not offering the capabilities to accurately capture details of fiber-level structures and the resulting lack of realism.
Drago and Chiba~\cite{drago2004painting} focused on simulating the macro- and microgeometry of woven painting canvases based on procedural displacement for modeling the arrangement of the woven yarns (i.e. a spline-based representation) and surface shading.
The model by Irawan and Marschner~\cite{irawan2012specular} also predicts yarn geometry (in terms of curved cylinders made of spiralling fibers) and yarnwise BRDF modeling to represent the appearance of different yarn segments within a weaving pattern. However, this approach does not model shadowing and masking between different threads.
The latter has been addressed with the appearance model for woven cloth by~\cite{sadeghi2013practical} that relies on extensive measurements of light scattering from individual threads, thereby taking into account for shadowing and masking between neighboring threads.
However, these approaches are suitable for scenarios where cloth is viewed from a larger distance, since reproducing the appearance characteristics oberservable under close-up inspection would additionally require the capability to handle thick yarns or fuzzy silhouettes as well as the generalization capability to handle fabrics with strongly varying appearances.
To increase the degree of realism, Guarnera \textit{et al.}~~\cite{guarnera2017woven} augment the yarns extracted for woven cloth in terms of micro-cylinders with adjustments regarding yarn width and misalignments according to the statistics of real cloth in combination with the simulation of the effect of yarn fibers by adding 3D Perlin Noise \cite{perlin1985image} to the micro-cylinder derived normal map.
Several approaches focus on fitting an appearance model like Bidirectional Reflectance Distribution Functions (BRDFs) \mbox{\cite{dobashi2019inverse,jin2022woven}} to inferred micro-cylinder yarn models or Bidirectional Curve Scattering Distribution Function (BCSDF) \mbox{\cite{schroder2011volumetric}} to simulate the appearance from the fibers within each ply curve extracted for a pattern without explicitly modeling each individual fiber or applying a pre-computed fiber simulation~\cite{montazeri2021systems}.
Extracting yarn paths from image data can be approached by leveraging the prior of perpendicularly running yarns for woven cloth (e.g., ~\cite{Schroeder2015}) as well as based on knitting primitive detection inspired by template matching with a refinement according to an underlying knitting pattern structure~\cite{trunz2019inverse} or deep learning based program synthesis~\cite{pmlr-v97-kaspar19a}.
While such approaches allow the modeling of the underlying yarn arrangements, the detailed yarn modeling with yarn widths, yarn composition, yarn twisting, hairiness, etc. is not explicitly modeled.\\
\indent Following investigations on the geometric structure of fabrics in the domain of the textile research community~\cite{morris1999modelling,tao1996mechanical,keefe1994solid,shinohara2010extraction}, several works focused on a more detailed modeling of the underlying cloth micro-appearance characteristics to more accurately model the underlying cloth characteristics such as thickness and fuzziness.
This includes volumetric cloth models~\cite{xu2001photorealistic,jakob2010radiative,zhao2011building,zhao2012structure,zhao2013modular}, that describe cloth in terms of 3D volumes with spatially varying density, as well as fiber-based cloth models~\cite{Schroeder2015,khungurn2015matching} that infer the detailed 3D structure of woven cloth at the yarn level with its fiber arrangement.
Zhao et al.~\cite{zhao2011building,zhao2012structure,zhao2016fitting} leveraged a micro-computed tomography (CT) scanner to capture 3D volumetric data.
The detailed volumetric scan allows them to trace the individual fibers and, hence, provides a an accurate volumetric yarn model that captures high-resolution volumetric yarn structure.
For instance, Zhao et al.~\cite{zhao2016fitting} present an automatic yarn fitting approach that allows creating high-quality procedural yarn models of fabrics with fiber-level details by fitting procedural models to CT data that are additionally augmented by a measurement-based model of flyaway fibers.
Instead of involving expensive hardware setups such as based on CT scanning, others focused on inferring yarn parameters from images, thereby representing more practical approaches for a wide range of users.
Voborova et al.~\cite{voborova2004yarn} focused on estimating yarn properties like the effective diameter, hairiness and twist based on initially fitting the yarn's main axis based on an imaging system consisting of a CCD Camera, a microscope, and optical fiber lighting.
Furthermore, with a focus on providing accurate models at less computational costs and memory requirements than required for volumetric models, Schröder et al.~\cite{Schroeder2015} introduced a procedural yarn model based on several intuitive parameters as well as an image-based analysis for for the structural patterns of woven cloth.
The generalization of this approach to other types of cloth, such as knitwear, however, has not been provided but still needs further investigation.
Saalfeld et al.~\cite{saalfeld2018fitting} used gradient descent with momentum to predict some of the procedural yarn parameters used by Zhao et al.~\cite{zhao2016fitting} from images of synthetically generated yarns. Although the results were promising for some of the parameters, the approach still could not be applied to the real yarn images.
Wu et al.~\cite{wu2019modeling} estimate yarn-level geometry of cloth given a single micro-image taken by a consumer digital camera with a macro lens, leveraging prior information in terms of a given yarn database for yarn layout estimation.
Large-scale yarn geometry is estimated based on image shading, whereas fine-scale fiber details are obtained based on fiber tracing and generation algorithms.
However, the authors mention that the use of a single micro-image does not suffice for the estimation of all relevant yarn parameters of complex procedural yarn models like the ones by Zhao et al.~\cite{zhao2016fitting} or Schröder et al.~\cite{Schroeder2015}, and, hence, the authors only consider the two parameters of fiber twisting and fiber count.
Whereas our yarn generator is conceptually similar to the one by Zhao et al.~\cite{zhao2016fitting}, there is an important difference in how we model the orientation of the twisting axis of the fibers.
Instead of using the global $z$-axis, we align the twisting axis with the relative $z$-axis of the next hierarchy level, resulting in a more realistic yarn structure.
This relative implementation allows adding additional hierarchy levels, i.e. especially for the hand knitting it is common to twist different yarns if they are too thin, thus creating the next level. 
Furthermore, our model's realism is further increased by also considering elliptic fiber cross-sections as occurring for natural hair fibers like wool and by considering a more natural modeling of flyaways.\\
\indent In the context of inferring physical yarn properties from visual information, Bouman et al.~\cite{bouman2013estimating} estimated cloth density and stiffness from the video-based dynamics information of wind-blown cloth.
Others focused on a neural network based classification of cloths according how stretching and bending stiffness influence their dynamics.
Furthermore, Rasheed et al.~\cite{rasheed2020learning} focused on the estimation of the friction coefficient between cloth and other objects.
Based on the combination of neural networks with physically-based cloth simulation, Runia et al.~\cite{runia2020cloth} trained a neural network to fit the parameters used for simulation to make the simulated cloth match to the one observed in video data.
Liang et al.~\cite{liang2019differentiable} and Li et al.~\cite{li2022diffcloth} presented approaches for cloth parameter estimation based on sheet-level
differentiable cloth models.
Gong et al.~\cite{gong2022fine} introduced a differentiable physics model at a more fine-grained level, where yarns are modelled individually, thereby allowing to model cloth with mixed yarns and different woven patterns. 
Their model leverages differentiable forces on or between yarns, including contact, friction and shear.

%% file: modeling.tex
\section{Generation of synthetic training data}\label{ssec:yarn_generation}
Our learning-based approach to infer yarn parameters from images relies on the availability of a database of images of yarns with respective annotations. However, to the best of our knowledge, no such database exists to this day. 
Since the exact measurements of the parameters of a real yarn is a complex task that requires experts as well as additional hardware such as a CT scanner, we overcome this problem by leveraging modeling and rendering tools from the field of computer graphics to create 
images of synthetic yarns with known parameters that can be directly used for learning applications.

To enable robust parameter inference from photographs of real yarns, the synthetic yarn images used for training the underlying neural model must be highly realistic, i.e. they must accurately model the yarn structure with its underlying arrangement of individual fibers.
Similar to Zhao et al.~\cite{zhao2016fitting}, we chose a fiber-based model rather than a volumetric one to gain more control over the generation and achieve higher quality. However, to increase the realism of the synthetic yarns, we extended the yarn model of Zhao et al. by including elliptical fiber cross-sections, local coordinate frame transformation for helix mapping and considering more complex modeling of hair flyaways.

We mimic the actual manufacturing process by introducing a hierarchical approach.
Multiple fibers are twisted together to form a ply, and, in turn, multiple plies are twisted together to form a yarn. If necessary, multiple thinner yarns can be twisted into a thicker yarn, which is sometimes the case in knitwear manufacturing. We denote the yarn resulting from this hierarchical procedure as \emph{raw yarn}.

In addition to capturing the characteristics of the fiber arrangement of the yarn structure, we must also consider that some of the fibers, referred to as \emph{flyaways}, may deviate from their intended arrangement within yarns and run outside the thread.
These deviations are caused by friction, aging or errors in the manufacturing process and play a central role in the overall appearance of yarns and the fabrics made from them.

Therefore, our yarn model is controlled by a number of parameters, which belong to two types: raw yarn parameters and flyaway parameters.
During generation, these parameters are stored along with the respective resulting images, and later serve as training labels for the network training.

The raw yarn is recursively built from multiple hierarchical levels (see Fig.~\ref{fig:twisting} and Alg. \ref{alg:raw_yarn}).
In the next step, flyaways are added (Alg. \ref{alg:flyaways}) and detailed fiber parameters such as material and cross section are defined. 
The yarn is then ready to be rendered.
We generate and render the synthetic yarn images using \emph{Blender}, 
which offers advanced modeling capabilities that can be fully controlled by Python scripts, making it suitable for procedural modeling. Especially, the high level mesh modifiers allow for relatively compact scripts.
Additionally, it also contains a path tracer capable of rendering photo-realistic images, which allows us to build an all-in-one pipeline.
\subsection{Hierarchical yarn model}\label{ssec:yarn_levels}
During the first step, yarns, plies and fibers are represented as polygonal lines, i.e. a tuple $\left(V,E\right)$ that stores the vertex positions $V=\left\{ v_{i}\in\mathbb{R}^{3}|i\in\mathbb{N}\right\}$ and their edges $E=\left\{ \left(i,j\right)|i,j\in\mathbb{N}\right\} $.
Note that our generation process allows for an arbitrary number of levels. However, in the rest of the paper, we will demonstrate the concept using three levels, fibers, plies, and yarns. Algorithm \ref{alg:raw_yarn} presents an overview over our recursive hierarchical generation of the raw yarn.
\begin{algorithm}
\caption{Recursive hierarchical generation of raw yarn}
\begin{algorithmic}[1]
\Require  level of raw yarn $level$
\Procedure{buildlevel}{$level$}
	\If {$level$ = 0}
		\State create straight polygonal line $line$
		\State \Return $line$ %(@\label{alg:make_line}@)
	\Else
		\State $template \gets$ \Call{buildlevel}{$level-1$}
	\EndIf

	\State $P \gets$ create $N$ instance positions using Eq. \ref{eq:small1} or Eq. \ref{eq:big1}-\ref{eq:big3}
	\State $output \gets \emptyset$ 
	\ForAll {$p\in P$}
		\State $I \gets copy(template)$
		\State $I \gets$ scale $x$ coordinate of $I$ with $e$ for elliptical cross-section
		\State $I \gets$ rotate $I$ using Eq. \ref{eq:rotate}
		\State center $I$ at position $p$
		\State generate helix at position $p$ using Eq. \ref{eq:center1}-\ref{eq:center3} and let $I$ follow the helix
		\State $output \gets output \cup I$ 
	\EndFor
	\State \Return $output$
\EndProcedure
\end{algorithmic}
\label{alg:raw_yarn}
\end{algorithm}
The input of the first level, the fiber level, is a simple straight polygonal line that must be chosen large enough to allow for the required resolution.
The vertices $v_i$ of the line are given by
\begin{linenomath}
\begin{equation}
\label{eq:vertex}
v_{i}=\left(\begin{array}{ccc}
0 & 0 & i\alpha_{f}\end{array}\right)^{T}
\end{equation}
\end{linenomath}
Here, $\alpha_{f}$ denotes the distance between two consecutive
vertices of a fiber.
\begin{figure}[]
	\center
	\includegraphics[scale=.18]{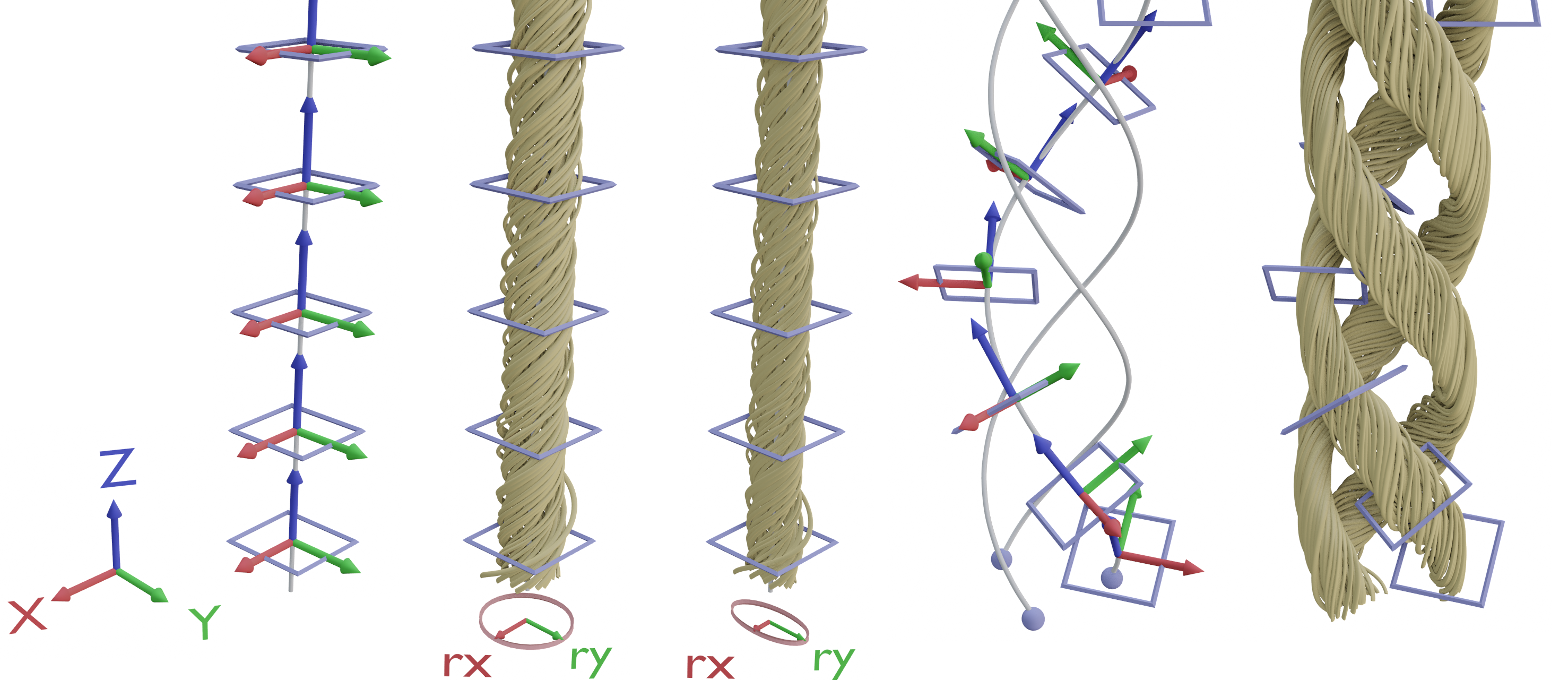}
	\hspace{1.0cm} a) \hspace{0.9cm} b) \hspace{0.8cm} c) \hspace{1.2cm} d) \hspace{1.3cm} e)
	\caption{Hierarchical twisting process. a) Level 1 corresponds to a straight polygonal line. b) Twisted fibers from level 1 form a ply on the second level. c) Before twisting plies into a yarn, the $x$-axis of each ply is downscaled to create an elliptical cross-section. d) Multiple initial positions (blue) are sampled, and a helix curve with the specified properties is created at each. These curves, called \emph{center lines}, represent the paths of the different plies. e) Deformed copies of the initial input follow each helix curve, resulting in the yarn on the third level and forming the input for the next step.}
	\label{fig:twisting}
\end{figure}
\begin{figure}[]
	\center
	\includegraphics[scale=.34]{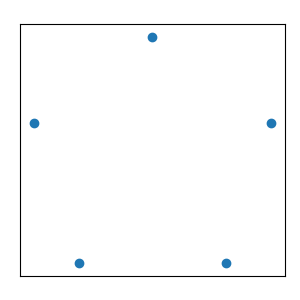}
	\includegraphics[scale=.34]{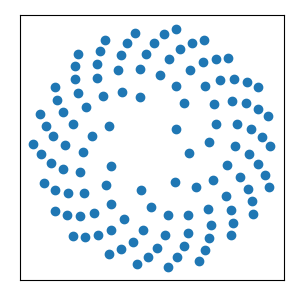}
	\includegraphics[scale=.34]{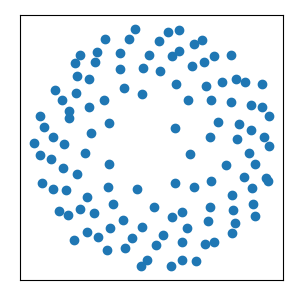}
	\caption{Ply and fiber distribution for the process explained in Fig.~\ref{fig:twisting}. For each level, multiple instances of the previous level are created and placed at initial positions according to a specified distribution. We use more randomness (middle) and jitter (right) on the fiber level and more structure on the ply level (left).}
	\label{fig:distribution}
\end{figure}
In each of the higher levels, we start by creating a set of $N$ 2D instance start positions $p_i$.
We define two variations of this procedure, one for small amounts of instances ($\sim 7$), and one for larger (in practice up to 200).
Both are illustrated in Fig.~\ref{fig:distribution}.
In both cases, we add some jitter $j_{xy}$ to the sample positions. For small numbers $N$ of instances, we generate a regular pattern on a circle with radius $r$:
\begin{linenomath}
\begin{equation}
p_{i} 
=r\left(\begin{array}{c}
\sin\theta_{i}\\
\cos\theta_{i}
\end{array}\right)+j_{xy}R\left(\begin{array}{c}
1\\
1
\end{array}\right)\label{eq:small1},\quad 
\theta_{i}  =2\pi\frac{i}{N} 
\end{equation}
\end{linenomath}
Here and in the following, $R$ is a zero-mean, normal distributed random variable with a standard deviation of 1 that is redrawn for each occurrence.

For larger numbers of instances, we sample the whole area of a disc.
We distribute fewer samples towards the center, as instances in the middle are mostly occluded by the outer ones.
\begin{linenomath}
\begin{align}
p_{i}  
&=r_{i}\left(\begin{array}{c}
\sin\theta_{i}\\
\cos\theta_{i}
\end{array}\right)+j_{xy}R\left(\begin{array}{c}
1\\
1
\end{array}\right) \label{eq:big1}\\ 
r_{i}  &=r\frac{i^{0.3}}{N^{0.3}}\label{eq:big2}\\ 
\theta_{i}  &=2\pi\cdot0.137\cdot i \label{eq:big3}
\end{align}
\end{linenomath}
The heuristically chosen constants create a slightly pseudo-random distribution that is enhanced by the added jitter.

Next, for each sample point, we copy the instance template from the previous level, and then each instance is transformed as follows: 
Since the sampling patterns are roughly circular, we downscale the template along the $x$-axis, transforming its cross-section into an ellipse (see Fig.~\ref{fig:twisting}, c).
The rotation ensures that the smaller radius of the ellipse is oriented toward the center, which simulates the squeezing of the individual fibers for dense packing.
As a last step, the template is translated to the position of the sampling point.

To simulate the twisting that occurs during the production of real yarns, we create a helix in the $z$-direction at each sample point $p = (p_{x},p_{y})^T$ and transform the template to follow it accordingly (Fig.~\ref{fig:twisting}).
The helix is given by:
\begin{linenomath}
\begin{align}
\theta_{i} & =\frac{i}{H}2\pi+\arctantwo\left(p_y, p_x\right) \label{eq:rotate}\\
r_{h} & =\sqrt{p_{x}^{2}+p_{y}^{2}} \label{eq:center1}\\
s_{i} & =1+\max\left(0,R_{h}\right)\cdot\cos\left(2R_{h}+\frac{i}{H}2\pi R_{h}\right)\label{eq:center2}\\
v_{i} & =\left(\begin{array}{ccc}
r_{h}s_{i}\sin\theta_{i} &
r_{h}s_{i}\cos\theta_{i}&
\frac{i}{H}\alpha_h+j_{z}R_{i} \label{eq:center3}
\end{array}\right)^{T}
\end{align}
\end{linenomath}
Here, $\alpha_h$ is the height of each complete turn, called the \emph{pitch} of a helix. $H$ is the helix resolution, i.e. the number of vertices per turn. The number of turns for the helix depends on the desired total length of the generated yarn.
Since the helix is always curved around the center line, its radius $r_{h}$ is determined by the position of the sample point $p$.
The angle $\theta_i$ has an offset that ensures that the 0th vertex coincides with $p$.
The random variables $R_i$ and $R_h$ are drawn once per vertex and once per helix, respectively.
However, different occurrences of $R_i$ and $R_h$ are drawn independently.
$s_{i}$ is the fiber migration value, modulation of the helix radius that varies along the vertical axis.
It is realized by scaling the radius with a height-dependent cosine function with random amplitude, offset and phase speed.

Note that each template point is transformed to a local coordinate frame given by the helix at the corresponding height. We do not perform an actual physical simulation for the twisting process, as this would require a complex numerical simulation and thus increase computation time drastically.

For our recursive hierarchical raw yarn generation, we used generic variable names, such as $N$, $R$ and $\alpha_h$. The parameters of each level for the three-level fiber-ply-yarn model, used in the following, are summarised in Table \ref{tab_params}.
\subsection{Flyaway generation}\label{ssec:flyaways}
\begin{figure}[t]
	\center
	\includegraphics[scale=.16]{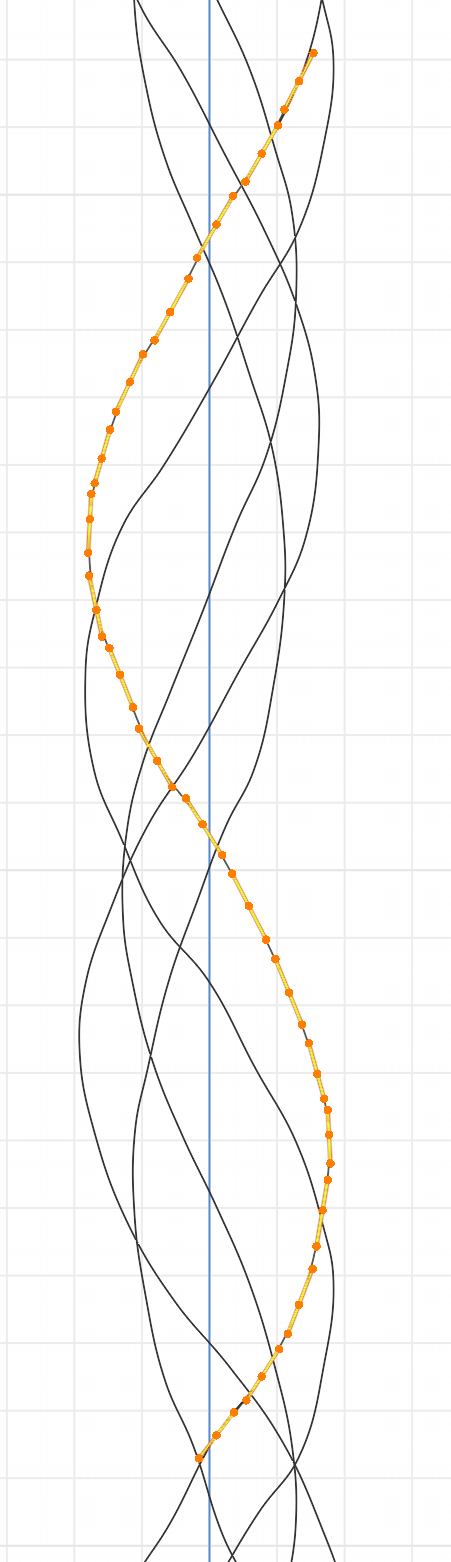}\hspace{0.3cm}
	\includegraphics[scale=.16]{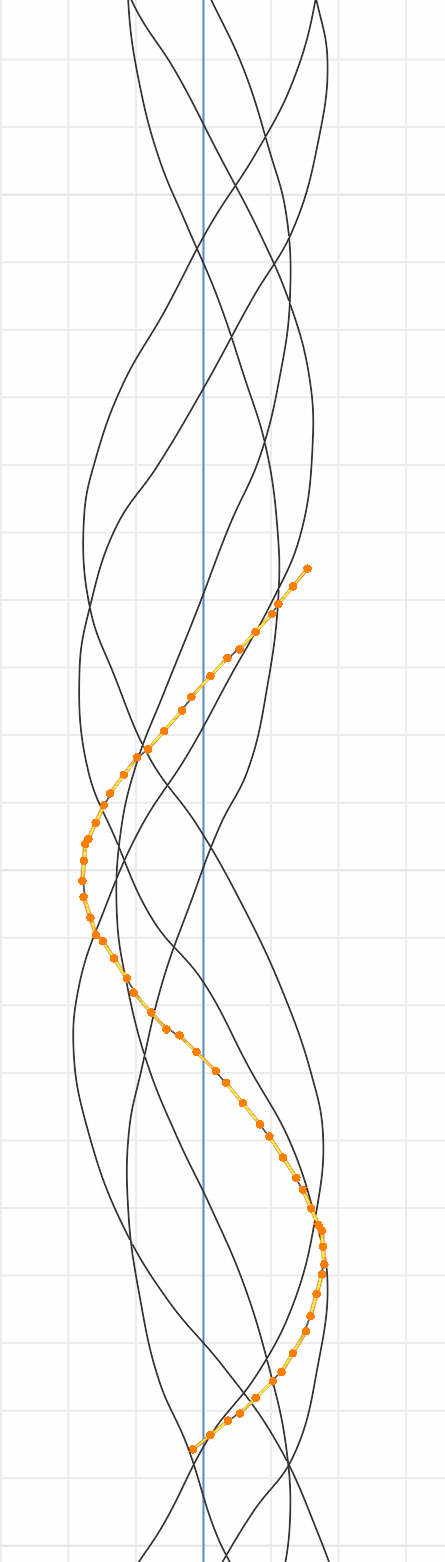}\hspace{0.3cm}
	\includegraphics[scale=.16]{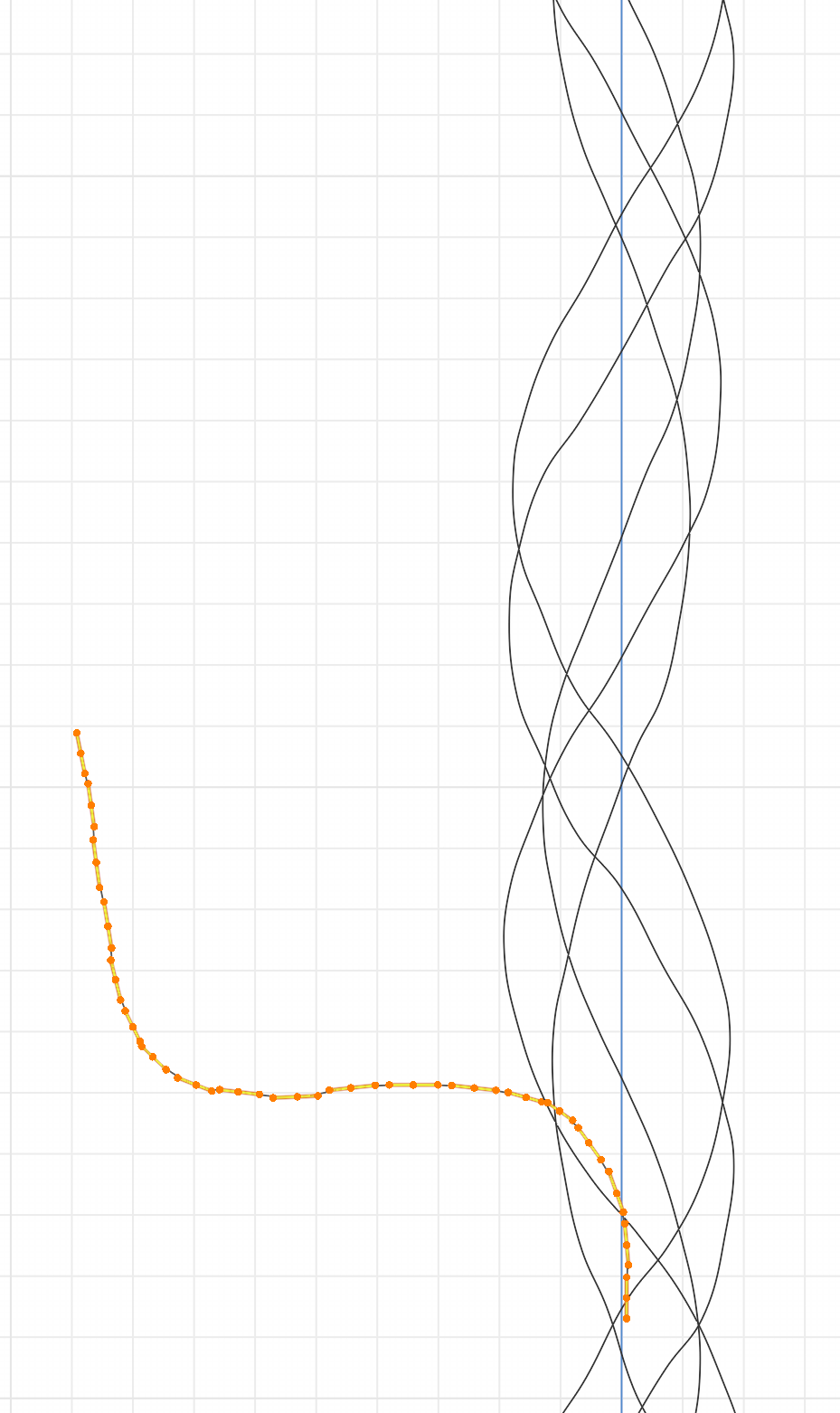}\hspace{0.3cm}
	\includegraphics[scale=.16]{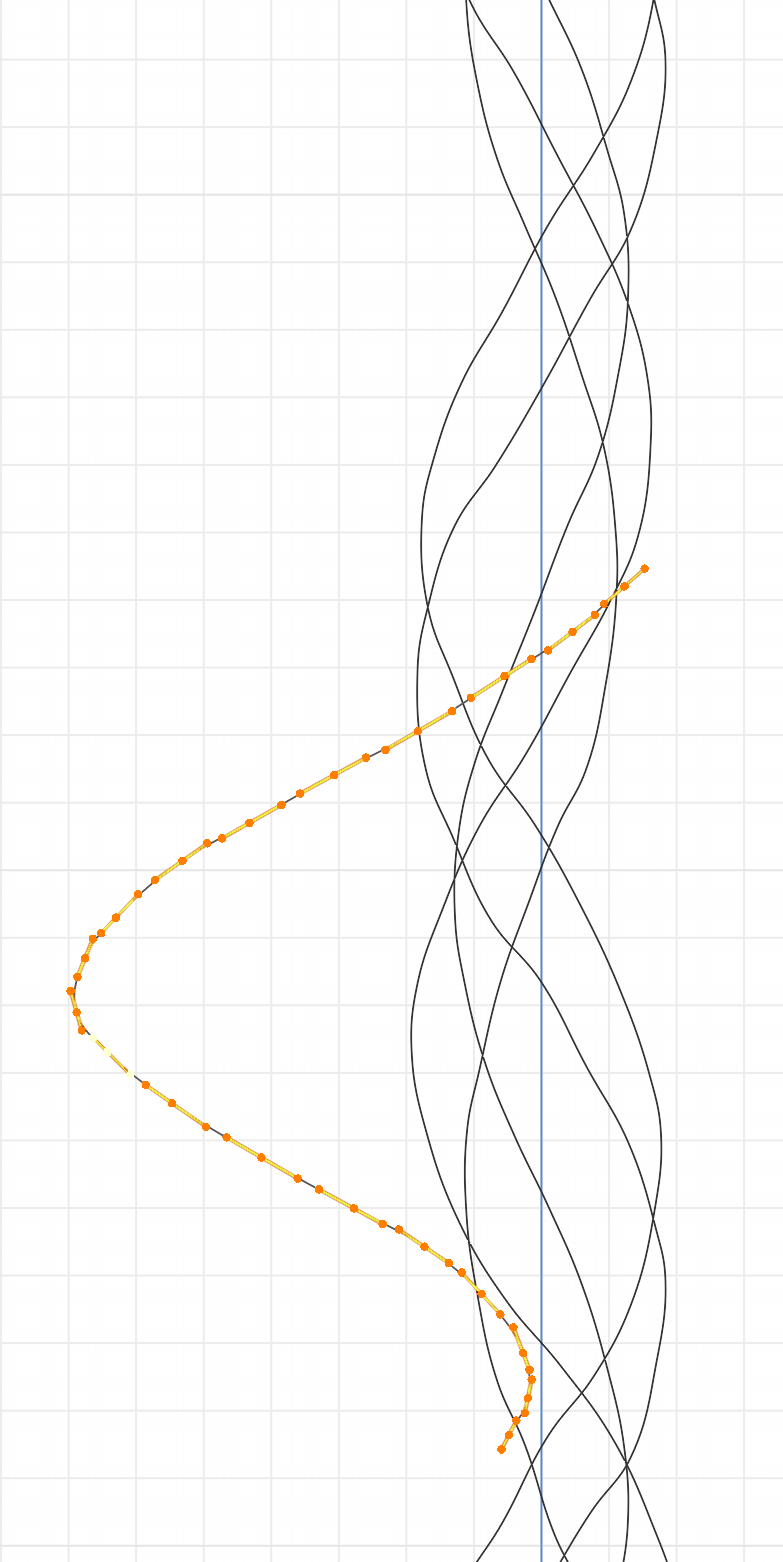}\\
		\hspace{0.05cm} a) \hspace{1.1cm} b) \hspace{1.6cm} c) \hspace{1.8cm} d) \hfil
	\caption{Generation of flyaways. a) A random vertex strip is selected and duplicated to become the new flyaway. b) The flyaway is scaled along its up-axis to exaggerate details. c) Hair flyaway: The flyaway from b) is rotated along its lowest point. d) Loop flyaway: The flyaway from b), where the vertices are moved radially according to a sine function, except for the first and last vertices, which remain at their previous locations, while the middle vertex is offset the most to simulate a loop.}
	\label{fig:flyawaygeneration}
\end{figure}

After creating the raw yarn structure according to the previous section, we now model the flyaways.
Flyaways are fibers that got displaced from their original position within the yarn.
Following the previous work~\cite{Schroeder2015, zhao2016fitting}, we distinguish between two different categories of flyaways.
\emph{Hair flyaways} are fibers where one side is completely outside the yarn, while \emph{loop flyaways} are fibers where both ends of the fiber are still inside the yarn, but the middle part is outside the main yarn.
Both types of flyaways and the key steps of their creation are shown in Figure~\ref{fig:flyawaygeneration}.
The generation of flyaways is summarized in Algorithm~\ref{alg:flyaways}.
Flyaways are created by copying and transforming parts of the yarn.
First, we determine whether the new flyaway will be a loop or a hair flyaway by drawing a uniformly distributed random number in $\left[0,1\right]$ and determining whether it is greater or less than the loop probability $p_l$.
\begin{algorithm}
\caption{Flyaway generation}
\begin{algorithmic}[1]
\Require  flyaway parameter $g$, $p_{l}$, $\beta$, $l_{hair}$, $s$, $l_{loop}$, $d_{mean}$, $d_{std}$
\Procedure{addflyaways}{$g$, $p_{l}$, $\beta$, $l_{hair}$, $s$, $l_{loop}$, $d_{mean}$, $d_{std}$}
	\For {$k\in \left[1,g\right]$}
		\State $fly \gets loop$ with prob. $p_{l}$, else $fly \gets hair$
	\EndFor
		\If {$fly = loop$}
			\State $length \gets l_{loop} + 0.01 R$ ($R$ as explained in \ref{ssec:yarn_levels})
		\Else
			\State $length \gets l_{hair} + 0.05 R$
		\EndIf
		\State find fiber segment $S$ of length $length$ via rejection sampling
		\If {$fly = loop$}
			\State create loop flyaway using Eq.~\ref{eq:loop1} on $S$
		\Else
			\State scale $z$ coordinates of $S$ and rotate by $\beta$ to create hair flyaways (Fig.~\ref{fig:flyawaygeneration})
		\EndIf
	\EndProcedure
\end{algorithmic}
\label{alg:flyaways}
\end{algorithm}
In both cases, the flyaway length is determined from a given mean and a fixed standard deviation.
Note that typical means are of the same order of magnitude as the standard deviations used.
To find a fiber segment for the new flyaway, a random vertex is selected and the chain of connected vertices is followed in a random direction.
If this chain ends before the desired length is reached, the process is repeated with a different starting vertex (rejection sampling).
Once a suitable segment is found, it is copied and transformed according to its type in the next step.
Copying a segment from the original yarn, rather than creating a new vertex line, preserves the deformation from the overlapping helixes from different levels, adding realistic detail.

Loop flyaways are created by overlaying the segment with a sine wave by adding an offset to each vertex:
\begin{linenomath}
\begin{equation}
o_{i} =d\sin\left(\frac{i\pi}{j}\right)\left(\begin{array}{ccc}
v_{x}&
v_{y}&
0 
\end{array}\right)^{T},\quad
d =d_{mean}+d_{std}R \label{eq:loop1}
\end{equation}
\end{linenomath}
The sine wave moves the vertex in a radial direction, keeping its vertical coordinate untouched.
$j$ is the total number of vertices in the segment, so exactly half a period of the sine wave is used, ensuring that the first and last vertices remain at their original positions, thus creating the loop shape.
The amplitude $d$ is chosen per flyaway, not per vertex.

Hair flyaways are created by rotating the segment by the angle $\beta$  (see Fig.~\ref{fig:flyawaygeneration}).
Prior to rotation, they are scaled along the vertical axis by a value of $s$ to amplify their shape.

Once all levels and flyaways are created, 
the bevel parameter is set to control the thickness and ellipticity of each fiber, giving the object a proper volume. All learnable parameters for the yarn and the flyaways are summarized in Table \ref{tab_params}.
\begin{table}
\centering
\scriptsize
\caption{Parameters of our procedural Blender yarn model. Top: Fiber parameters, Middle: Ply parameters, Bottom: Flyaway parameters. Although fiber distribution and migration are not technically flyaway parameters, we consider them as such for our parameter prediction due to their probabilistic nature.}
%\label{table}
\setlength{\tabcolsep}{7pt}
\begin{tabular}{|p{50pt}|p{30pt}|p{80pt}|}
\hline
Parameter type & Parameter name & Explained \\
\hline
Fiber amount &$m$ & Number of fibers in each ply \\
Fiber ellipse & $t_x$, $t_y$ & Radii of fiber ellipse \\
Fiber twist &$\alpha$ & Pitch of the ply helix \\
\hline
Number of plies &$n$ & Number of plies in the yarn \\
Ply ellipse & $r_x$, $r_y$ & Radii of ply ellipse \\
Ply twist &$\alpha_{ply}$, $R_{ply}$ &Pitch and radius of the yarn helix \\
\hline
Fiber migration & $j_z$,$j$ & Jitter of the fibers in $z$ and in radial direction\\
Fiber distribution & $j_{xy}$ & Jitter of fibers in $xy$ plane direction\\
Flyaway amount & $g$ & Number of flyaways\\
Loop probability & $p$ & Probability for loop type flyaway \\
Hair flyaways & $\beta$, $l_{hair}$, $s$& Angle, hair length, fuzziness\\
Loop flyaways & $l_{loop}$, $d_{mean}$, $d_{std}$& Loop length, Mean and std of distance from ply center\\
\hline
\end{tabular}
\label{tab_params}
\end{table}
\subsection{Further Implementation Details}
To increase the realism of the resulting yarn appearance, we apply a reflectance model to the individual fibers, which describes their view- and illumination-dependent appearance.
This allow us to obtain synthetic images of yarns by placing the yarn in a pre-built scene that resembles our measurement environment in the lab where we took the photos of real yarns.

We implement the yarn generation as a Python script inside the 3D modeling suite Blender, since it not only provides many of the operations needed during the generation, but also has powerful rendering capabilities.
In particular, we leveraged Blender's \emph{principled hair BSDF shader}, which is particularly relevant for our scenario of fiber-based yarn representation, as well as the built-in path tracer capable of rendering photo-realistic images with full global illumination to generate images depicting the synthesized yarns according to the conditions we expect to occur in photographs of real yarns.
\subsection{Extensions to State-of-the-art Yarn Generator}
\begin{figure}
	\center
	\includegraphics[height=1.6cm]{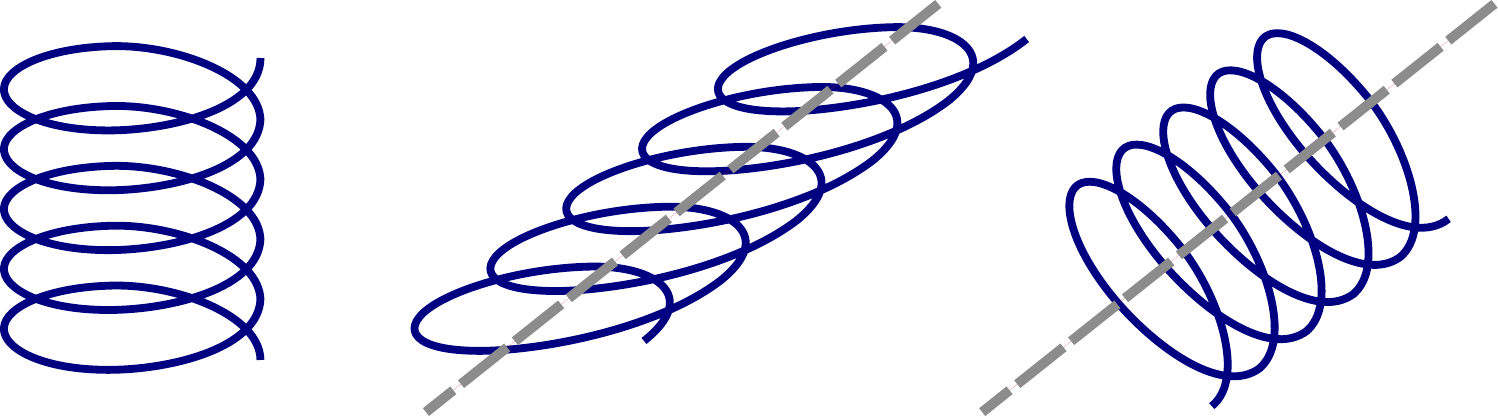}\\
	
	a) \hspace{1.7cm} b) \hspace{1.8cm} c)
	
	\caption{A ply (blue) is mapped to a helix segment (grey). The figure shows a very similar scene to Fig.~\ref{fig:twisting}, but drastically simplified and with exaggerated dimensions.
	a) The ply before mapping.
	b) Mapping by shifting orthogonal to the global vertical axis, as implemented in \cite{zhao2016fitting}.
	c) Mapping by applying a local coordinate frame transformation, as implemented in our generator.
	}
	\label{fig:helix_mapping}
\end{figure}

\begin{figure}
	\center
	\includegraphics[height=3.6cm]{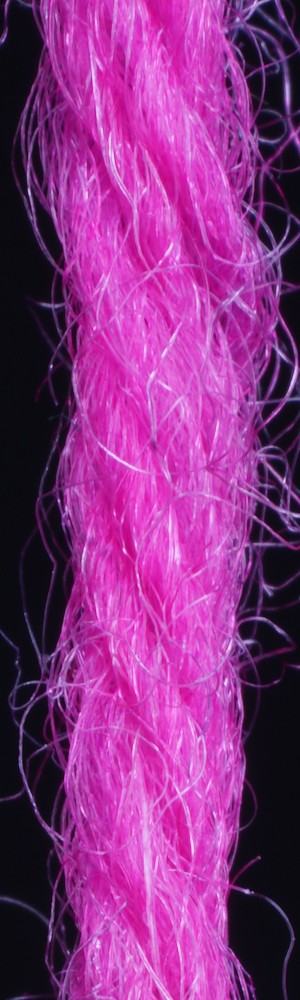}\hspace{0.17cm}
	\includegraphics[height=3.6cm]{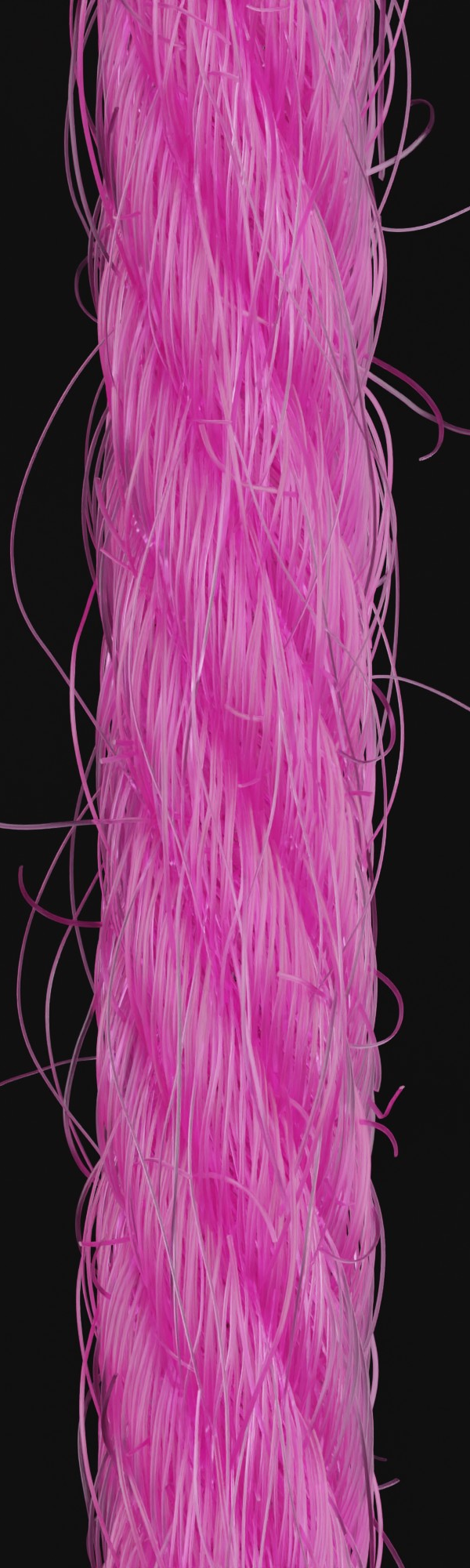}\hspace{0.17cm}
	\includegraphics[height=3.6cm]{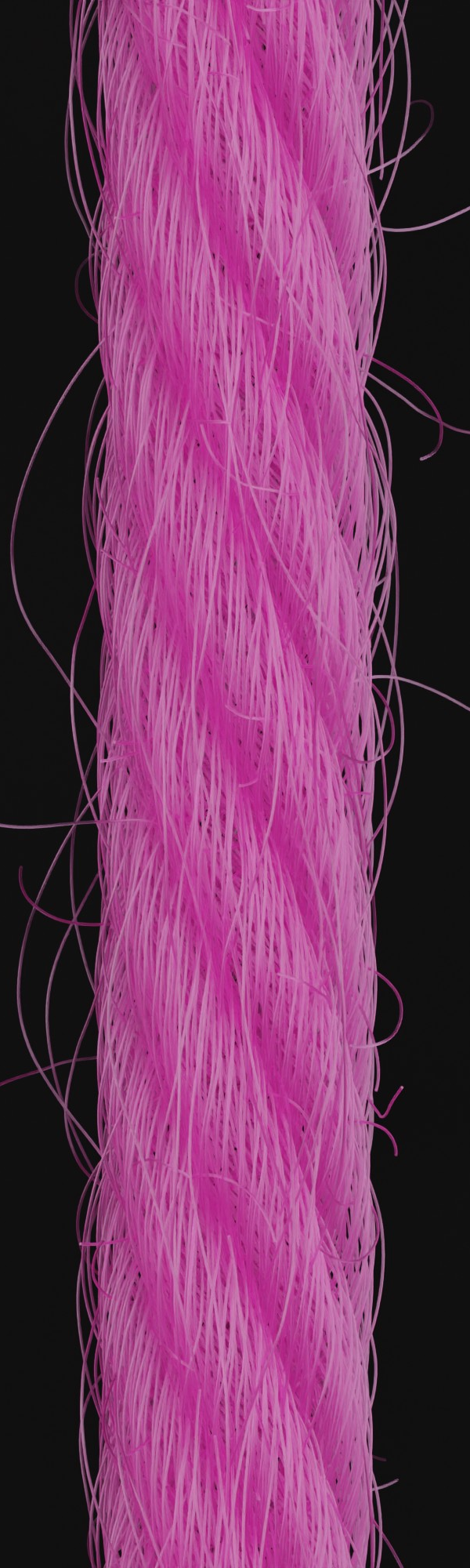} \hspace{0.35cm}
		\includegraphics[height=3.6cm]{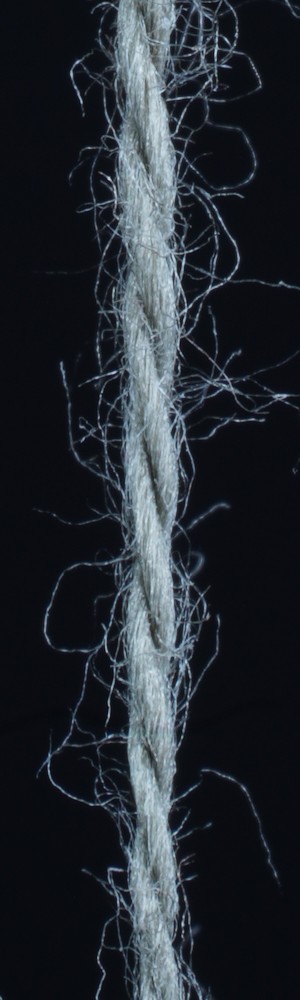}\hspace{0.17cm}
	\includegraphics[height=3.6cm]{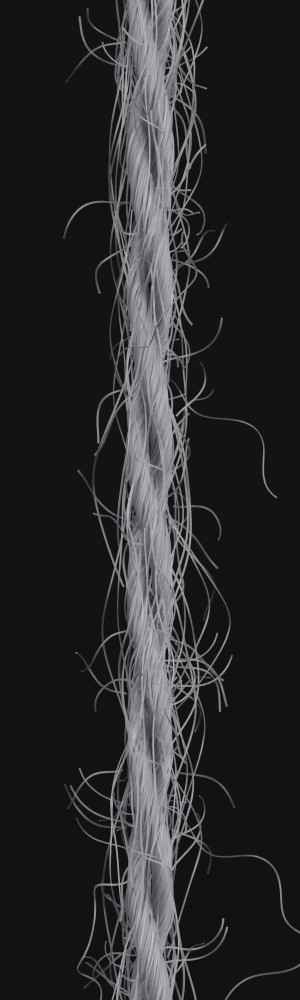}\hspace{0.17cm}
	\includegraphics[height=3.6cm]{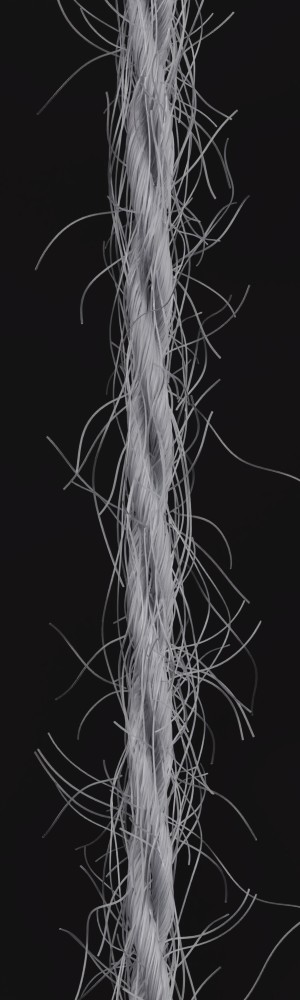} \\
	a) \hspace{1cm} b) \hspace{1cm} c) \hspace{1cm} d) \hspace{1cm} e) \hspace{1cm} f)
	\caption{Left: Comparisson of fiber cross section. a) Photograph of a wool yarn with elliptical cross section. b) Virtual yarn generated with elliptical fiber cross-section. c) Virtual yarn generated with circular fiber cross-section.
	The changes in geometry are hard to spot when zoomed out, however the shading and in particular the strength of the specular highlights is clearly affected by the cross-section shape. Right: Effect of the squeeze parameter $s$. d) Reference, e) With squeeze, f) Without squeeze.}
	\label{fig:fibers}
\end{figure}

Whereas Zhao et al.~\cite{zhao2016fitting} focused on woven cloth made of cotton, silk, rayon and polyester yarns, we observed that in addition to these fiber types, knitwear is often made of various types of natural wool (cashmere, virgin wool, etc.) and acrylic a as wool substitute, as they offer exceptional warming properties and knitwear is mainly worn or used in the colder months.
These and most other fiber types have longer flyaways, and their fibers exhibit elliptical cross-sections rather than circular ones, as assumed by Zhao et al.~\cite{zhao2016fitting}.
These observations inspired us to make the following extensions to the current state-of-the-art models\cite{Schroeder2015,zhao2016fitting}:
\begin{itemize}
\item Hair flyaways: Instead of implementing hair flyaways in terms of adding hair arcs, we simulate them similarly to loop flyaways in terms of being pulled out of the plies. 
Hence, the twist characteristics are preserved (see Fig.~\ref{fig:flyawaygeneration}, c)).
Furthermore, we leverage hair squeezing to simulate the effect that when flyaways are released from the twist, they are less stretched and contract slightly (see Fig.~\ref{fig:flyawaygeneration}, b).
These two steps make even the longer flyaways look realistic (see Fig.~\ref{fig:fibers}, d-f).
\item Elliptical fiber cross-sections:
We implement the ellipticity of the cross-section of many types of fibers, which is particularly prominent in natural hair fibers such as wool.
Although the geometric changes are too small to be seen directly, the shape of the cross-section affects the shading during the rendering, especially the prominence of specular highlights (see Fig.~\ref{fig:fibers}, a-c).
\item Local coordinate frame transformation for helix mapping:
Previously, in \cite{zhao2016fitting}, plies were twisted by sliding individual vertices orthogonal to the global vertical axis.
Instead, we introduce a proper coordinate system transformation, which leads to more plausible results. In some cases, the differences are small, in others they are much more obvious. Fig.~\ref{fig:helix_mapping} shows an exaggerated case to illustrate the difference.
%A realistic case is shown in Fig.~\ref{fig:discussion} d)-e).
\item Hierarchical generation: Sometimes, when multiple thinner threads are twisted into a thicker thread, yarns with more than three levels occur.
Our hierarchical generator allows for any number of levels.
\end{itemize}
\paragraph*{Evaluation of performance}
Although in its current implementation, the yarn generation process is more optimized for clarity and ease of use rather than efficiency,
the time for generating all fiber and flyaway curves (about 6-12 seconds per image) is significantly less than the rendering time (about 1 to 4 minutes). This makes it suitable for our purpose of generating a database of yarns, but further optimization of the generation process may be an aspect for future developement.
\subsection{Yarn dataset}\label{ssec:dataset}
\begin{figure}[!htb]
	\center
	\includegraphics[height=3.4cm]{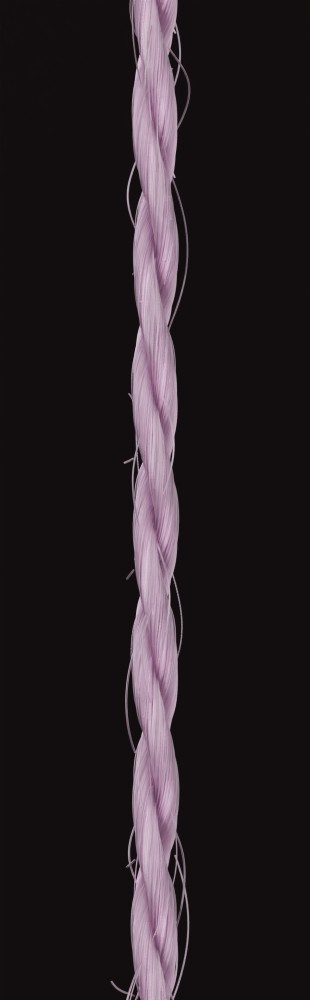}
	\includegraphics[height=3.4cm]{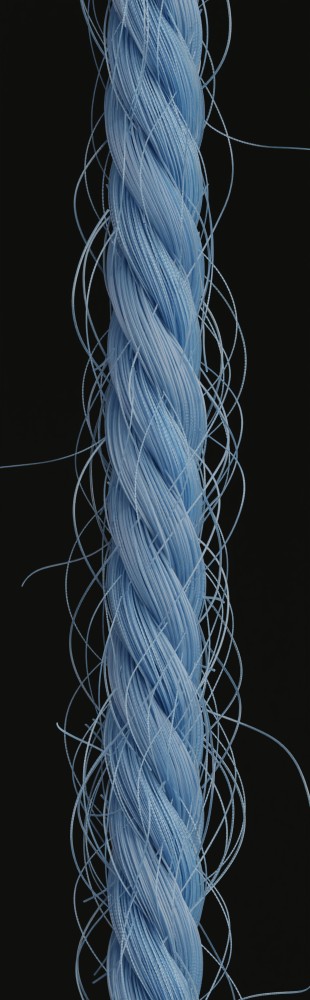}
	\includegraphics[height=3.4cm]{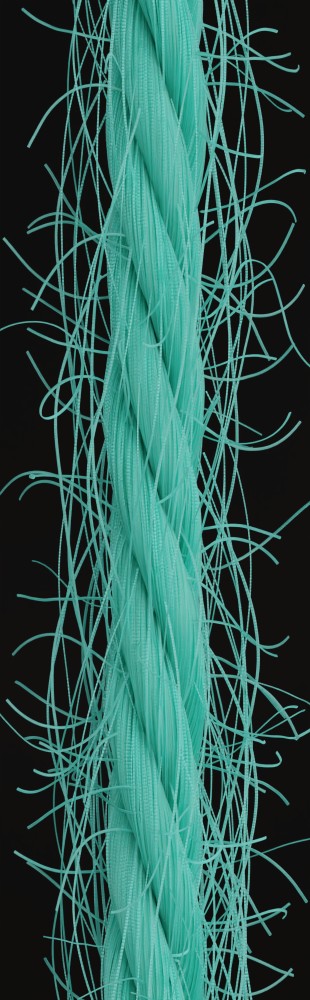}
		\includegraphics[height=3.4cm]{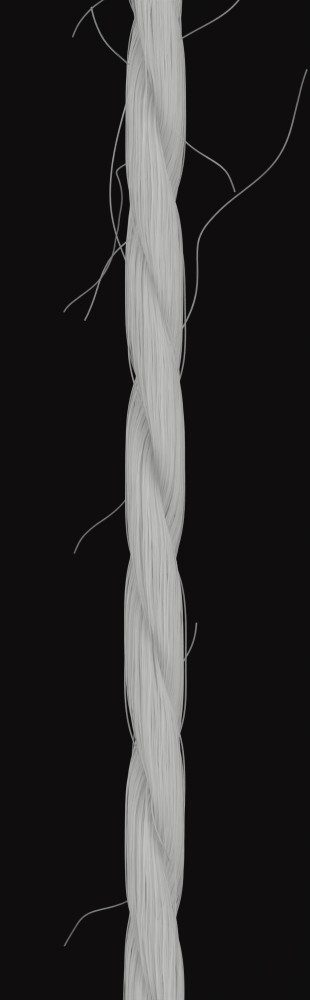}
	\includegraphics[height=3.4cm]{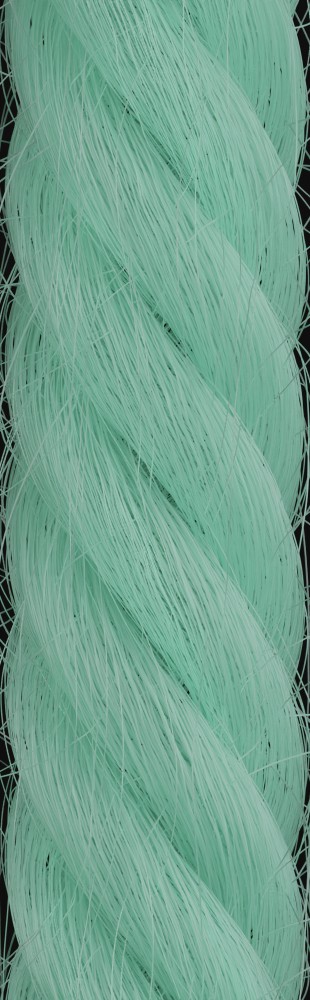}
	\includegraphics[height=3.4cm]{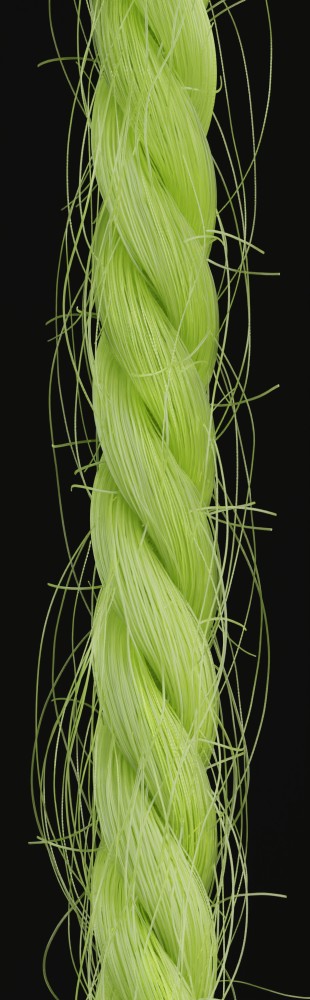}
	\includegraphics[height=3.4cm]{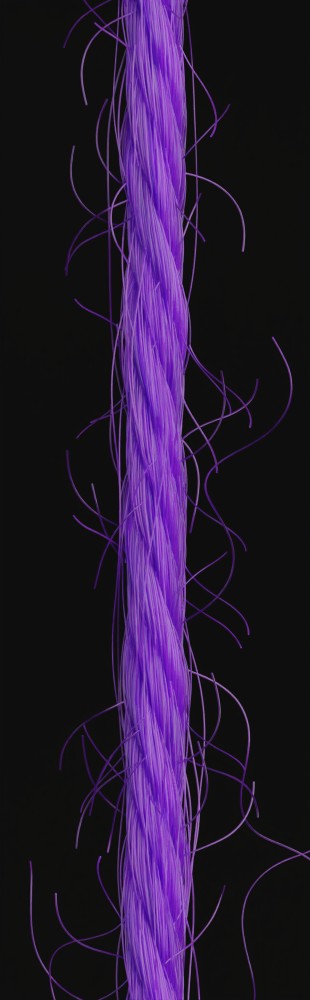}
	\caption{Examples of synthetic yarns in our database.}
	\label{fig:example_yarns}
\end{figure}

To represent the variations in color and reflective characteristics encountered in real yarns in our synthesized yarn dataset, we sample different of these parameter configurations by uniformly sampling the parameters within the corresponding, heuristically determined intervals shown in Table~\ref{tab_params} and then rendering the resulting yarns in different conditions that we expect to occur when considering photos of real yarns.
We provide details of our guided parameter sampling procedure in the supplementary material. All yarns in our database consist of two to six plies.
Note that our yarn generator allows the generation of yarns with more plies, but our observations indicate that three, four and five plies are the most common scenarios in the case of knitting yarns. Fig.~\ref{fig:example_yarns} depicts some of the yarns from the database.
In total, we sampled 4000 parameter configurations for the synthetic training set and 345 parameter configurations for the synthetic validation set, resulting in 4000 images with a resolution of 2000x600 pixels for training and 345 images with a resolution of 2000x600 pixels for validation.

Although our yarn generator can generate many levels of hierarchy, for proof-of-concept purposes, in this paper, we focused on yarns made up of plies and did not investigate learning the next level, where multiple thinner yarns are twisted into a thicker yarn. Therefore, our database does not include such yarns.
Furthermore, by rendering the yarn in different scenes, including various indoor and outdoor settings, training data for \emph{in-the-wild} yarn parameter estimation could be generated.

%% file: fitting.tex
\section{Inference of yarn characteristics from input images}
We model the prediction of the parameters for our procedural yarn model from images as a regression problem. Our training and validation dataset consists of annotated synthetic yarn images that we use to train a model that allows inferring yarn parameters from novel images of yarns not seen during training or validation. 
Before providing details of our respective approach (see Section~\ref{ssec:yarn_fitting}), we motivate our choice of a suitable network architecture that is capable of handling the challenging nature of the underlying problem.
We considered the saliency maps \cite{simonyan2013deep} of networks trained to predict the set of raw yarn's parameters and the set of the flyaway parameters with two independent models. 
An entry $m_{i,j}$ in a saliency map for a model $f$ that has been trained on a subset of $P$ parameters is the maximum derivative of the average value of the predicted parameters with respect to a pixel $x_{i,j,c}$ in the input image over the color channels $c$, i.e. 
\begin{linenomath}
\begin{equation}
    m_{i,j} = \max_c \left| \frac{\partial}{\partial x_{i,j,c}} \left( \frac{1}{P} \sum_p f_{p}(x) \right) \right|.
\end{equation}
\end{linenomath}
In contrast to saliency maps that have been proposed within the context of classification networks, we consider the derivative of the mean of the predicted parameters because we need to investigate the effect of a pixel on the entire subset on which the network was trained.

\begin{figure}[!htb]
	\center
	\begin{subfigure}{0.1\linewidth}
    \caption*{Yarn}
		\includegraphics[height=3cm] {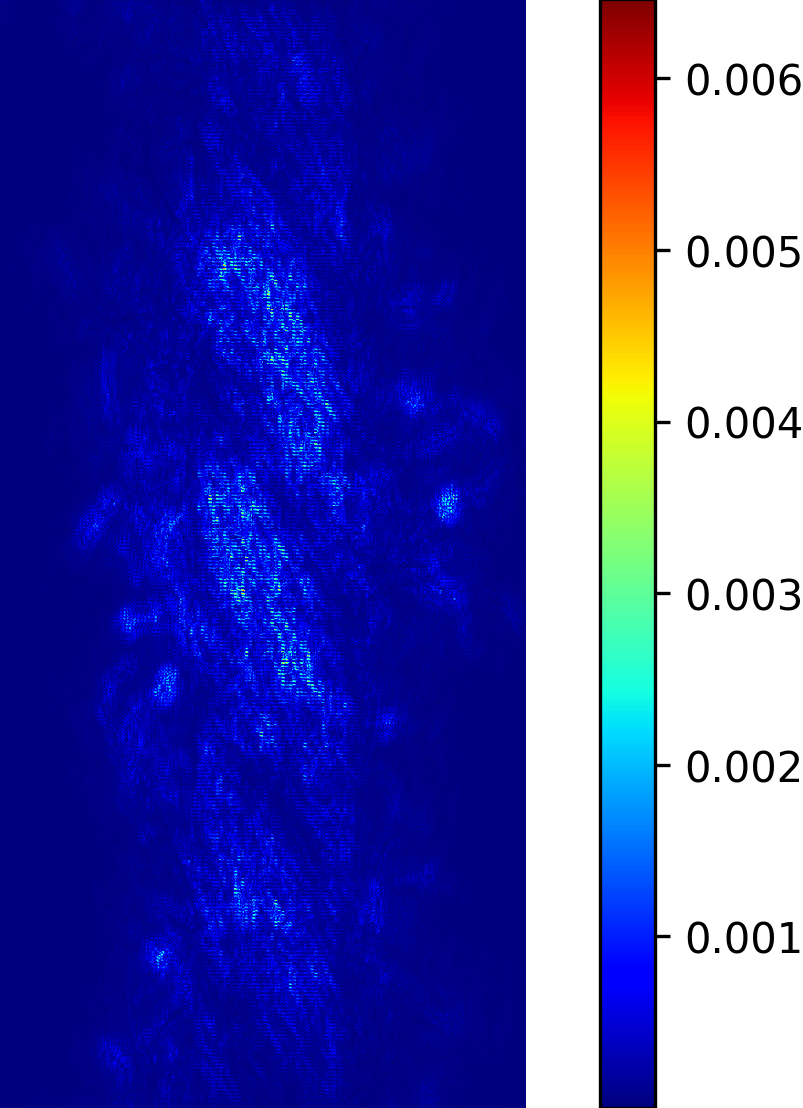} \\[3pt]
		\includegraphics[height=3cm] {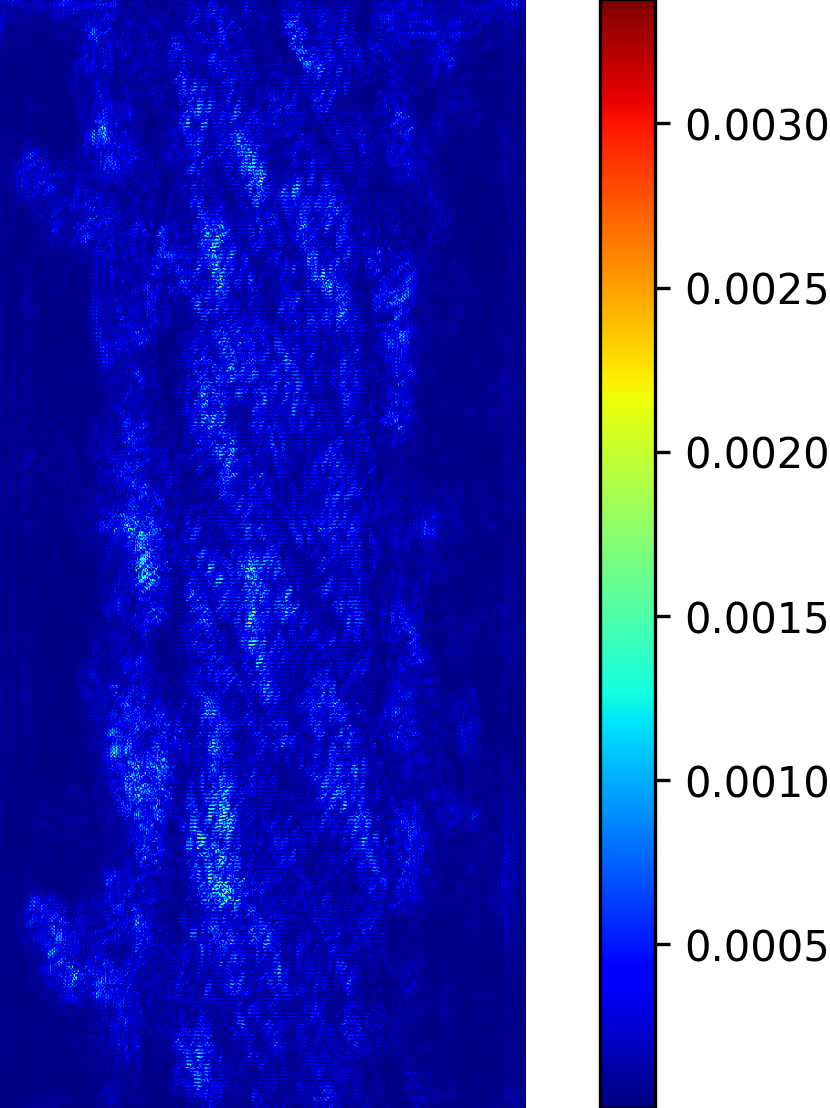} 
		\end{subfigure}
		\hfil
	\begin{subfigure}{0.1\linewidth}
		\caption*{Flyaways}
		\includegraphics[height=3cm] {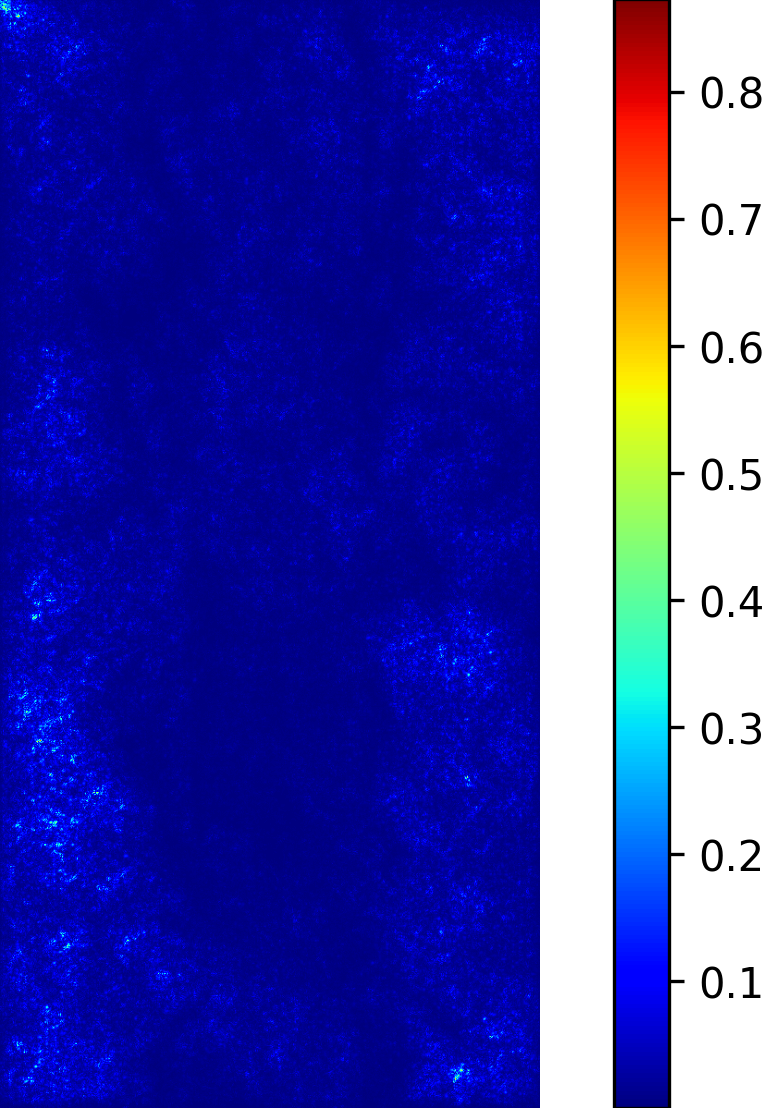}\\[3pt]
		\includegraphics[height=3cm] {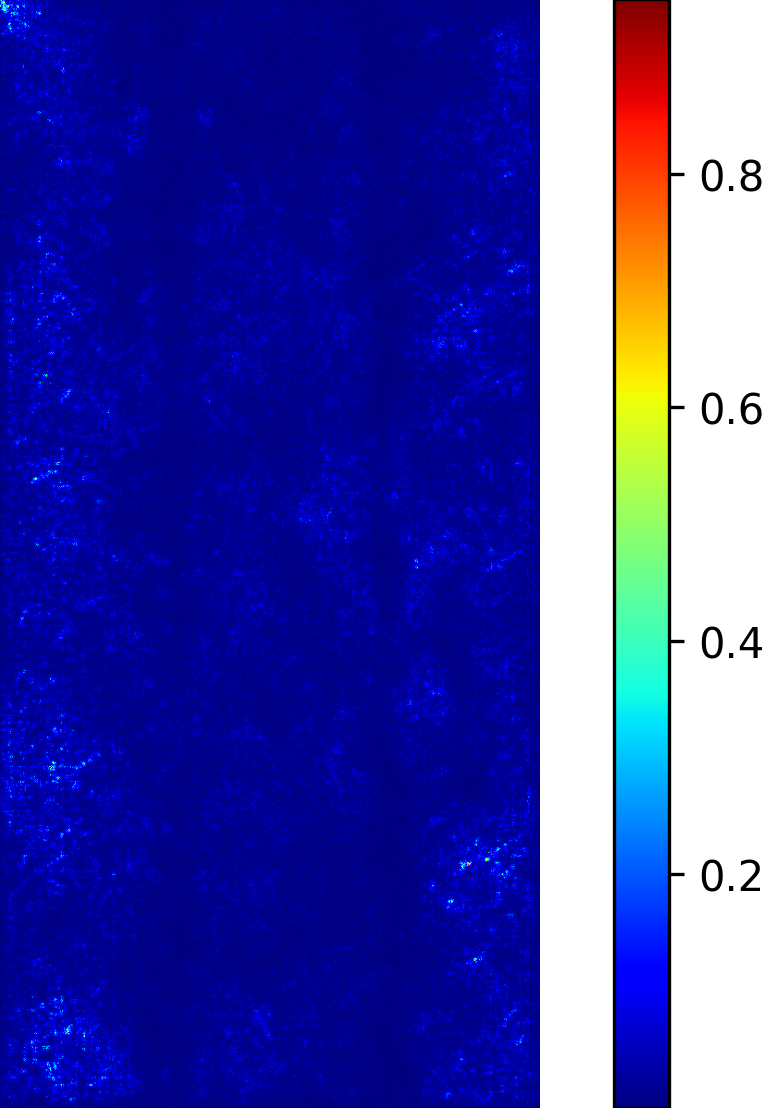} 
				\end{subfigure}
		\hfil
			\begin{subfigure}{0.1\linewidth}
		\caption*{Input}
		\includegraphics[height=3cm] {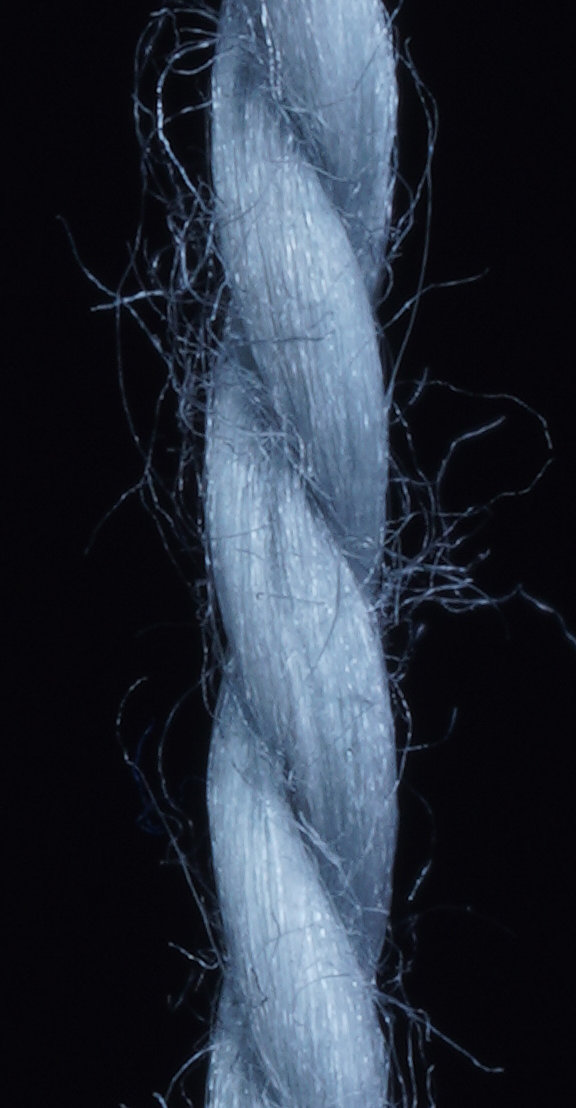}\\[3pt]
		\includegraphics[height=3cm] {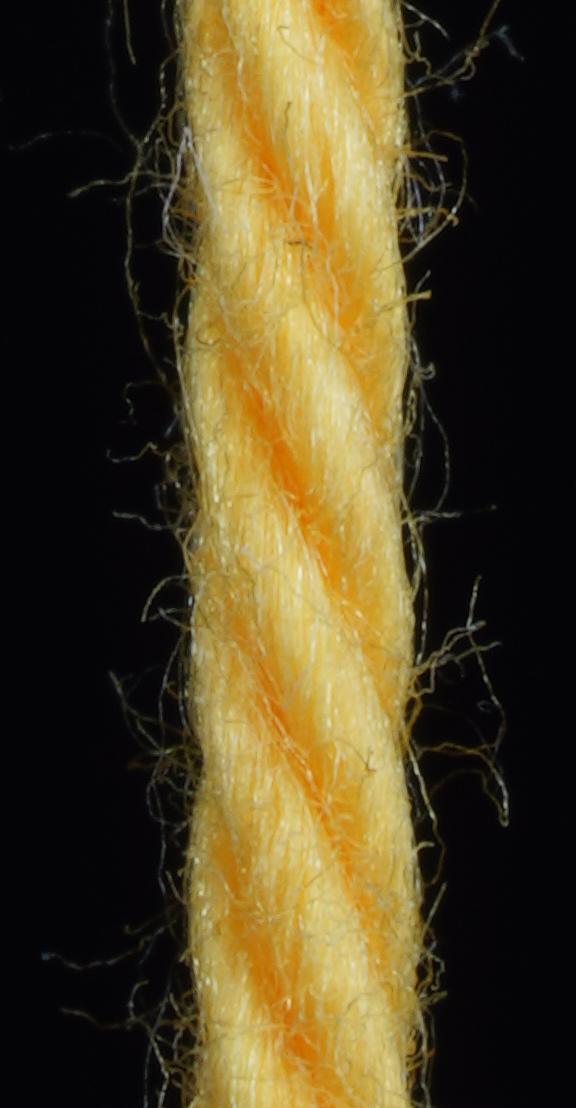} 
				\end{subfigure}
		\hfil
	\caption{Saliency maps computed for network configurations that are trained either to predict geometry (columns 1) or flyaway parameters (column 2) of the yarn model and either respective inputs (column 3). The color temperature in a saliency map indicates an input pixel's influence on the predicated parameter. Lighter/warmer colors correspond to a stronger influence.}
	\label{fig:saliency}
\end{figure}

The saliency maps (Figure~\ref{fig:saliency}) for the network trained to predict the raw yarn parameters illustrate a higher susceptibility to changes in the yarn center region of the image.
On the other hand, the saliency maps for the network that predicts the flyaway parameters indicate that such a network exhibits a higher sensitivity towards the border regions within the input images.
Motivated by these saliency maps, we concluded that it is better to train separate models for the raw yarn parameters and the flyaway parameters.
\subsection{Inference of yarn parameters}\label{ssec:yarn_fitting}
\begin{figure}[h]
    \centering
    \includegraphics[width=7cm]{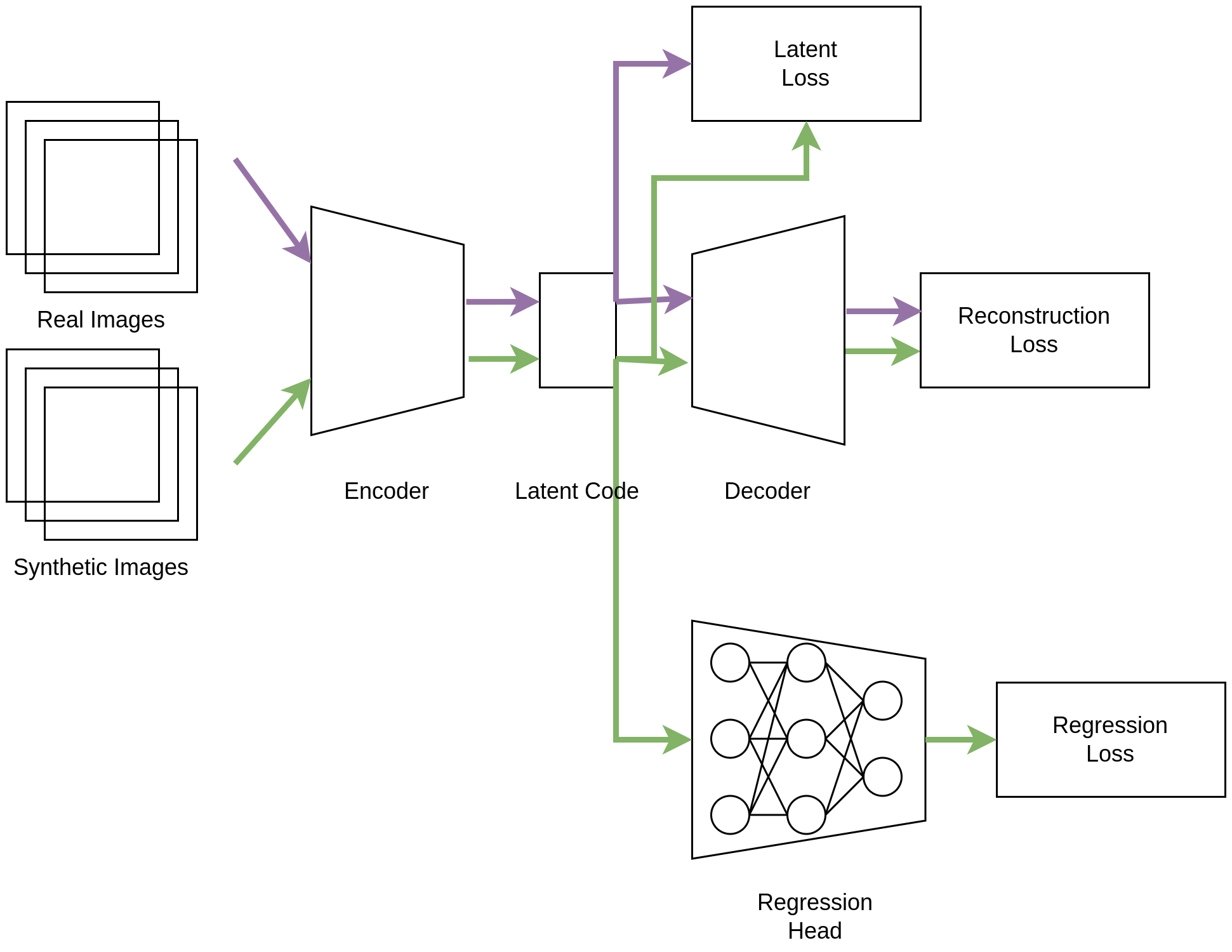}
    \caption{Overview of the interleaved training process. The depicted architecture combines all the different networks we investigated: The networks $Reg$ and $Reg_{latent}$ consist of the encoder and the regression head, while the networks $Reg^{ae}$ and $Reg_{latent}^{ae}$ additionally include the decoder.}
    \label{fig:training}
\end{figure}

As already mentioned before, we formulate the problem in terms of a regression problem.
Here an encoder $f$ maps an input image to a latent code which then becomes the input to a regression head $h$ (Fig.~\ref{fig:training}) which performs the parameter regression. 
This regression path within our model is trained to minimize an $L_1$ loss between the prediction obtained on synthetic yarn images $x^{(i)}$ and their ground truth parameter $y^{(i)}$ used in the generation model. 
\begin{linenomath}
\begin{equation}
    \mathcal{L}_\text{regress} = \mathbb{E} \left[ \left\lVert h(f(x^{(i)})) - \hat{y}^{(i)} \right\rVert_1 \right]
\end{equation}
\end{linenomath}
We will refer to this network simply as $Reg$.

Although the synthetic training data was carefully generated to match the appearance of real yarn photographs as accurately as possible, a domain gap between the synthetic and real images cannot be ruled out.
To address the domain gap, we investigated the impact of adding some not annotated real images to the training and utilized a semi-supervised training process which interleaves synthetic and real images in the training process to improve the extrapolation from synthetic images with known yarn parameters to real photographs.
For this purpose, we extended our aforementioned regression model into an autoencoder with an additional regression by adding a decoder model $d$.
The autoencoder of the path is trained to minimize a simple image reconstruction loss both on synthetic and real images:
\begin{linenomath}
\begin{equation}
    \mathcal{L}_\text{recon} = \mathbb{E} \left[\lVert d(f(x^{(i)})) - x^{(i)} \rVert_F^2 \right].
\end{equation}
\end{linenomath}
This unsupervised training process enables our encoder to be trained to map synthetic and real images into the same latent space from which the regression head predicts the yarn parameters for synthetic data points.
During inference, only the encoder and regression head are required to predict inputs for the parametric yarn model.
In an ideal case, the encoder maps the synthetic and real images of similar yarn to vectors that are within a close proximity in its latent space.
However, such a behavior is not guaranteed by the reconstruction loss.
On the contrary, the encoder might learn to distinguish between the synthetic and real images so that their latent vectors from two distinctive clusters.
We propose an additional regularization term which explicitly penalizes the distance between latent codes of synthetic images and the average latent code of real images in the encoder's latent space:  
\begin{linenomath}
\begin{equation}
    \mathcal{L}_\text{latent} = \mathbb{E} \left[\lVert f(x^{(i)}) - \text{sg}(\mu_\text{SMA}) \rVert_F^2 \right]
\end{equation}
\end{linenomath}
where $\text{sg}$ denotes the stop gradient function (i.e. $\mu_\text{SMA}$ is considered to be a constant in the backward step) and $\mu_\text{SMA}$ is a simple moving average of latent codes of real images for the last $5$ batches.
With this additional regularization term the average latent code of synthetic and real images are close to each other, and distinctive clusters become energetically less optimal.
Note that the evidence lower bound (ELBO) in the objective function of variational autoencoder could be considered as an alternative regularization.
However, it is more restrictive by forcing the latent code to be roughly normal distributed which is not necessary in this application.

In three different networks $Reg_{latent}$, $Reg^{ae}$ and $Reg_{latent}^{ae}$ we investigated the following three combinations of those losses:
\begin{itemize}
	\item Network $Reg_{latent}$: $\mathcal{L}_\text{reglat} = \mathcal{L}_\text{regress} + \lambda_\text{latent1} \mathcal{L}_\text{latent}.$ 
	
	\item Network $Reg^{ae}$:  $\mathcal{L}_\text{regrec} = \mathcal{L}_\text{regress} + \lambda_\text{recon1} \mathcal{L}_\text{recon}$.
	
	\item Network $Reg_{latent}^{ae}$: The combination of both previous variants, i.e.: $\mathcal{L}_\text{combined} = \mathcal{L}_\text{regress} + \lambda_\text{recon} \mathcal{L}_\text{recon} + \lambda_\text{latent} \mathcal{L}_\text{latent}$.
\end{itemize} 
where $ \lambda_\text{recon}, \lambda_\text{recon1},\lambda_\text{latent}, \lambda_\text{latent1}$
are hyper-parameters of the models.
The combined architecture is provided in Figure~\ref{fig:training}.
\paragraph{Network Architecture}
The encoder architecture is a pure CNN model based on ResNet (\cite{resnet}) were the average pooling has been moved to the regression head i.e. the latent codes are the tensors which result from the convolution stack.
We explore both the ResNet18 and ResNet34 configurations with the standard ResNet block as proposed in \cite{resnet} as well as the more recently proposed convnext blocks which also replaces the batch normalization with layer normalization (\cite{convnext}).
% Regression head
If the encoder uses a ResNet18 or ResNet34 architecture, the regression head $h$ is a two layer MLP with a hidden dimension of 512 and Exponential Linear Unit activation function after the first layer.
Otherwise, the regression head is a linear projection of the 512-dimensional input onto the required output dimension.

The decoder $g$ consists of four transposed convolutions with kernel sizes $k_1=2$, $k_2=2$, $k_3=2$, $k_4=2$, the stride $s_i = k_i/2$ and output kernel sizes $256, 128, 64, 3$.
The first three layers use ReLU activations, while the last layer uses the Tanh function to ensure that the output values are within the range of the pixel values.

\begin{figure}[h]
	\center
	\begin{subfigure}{0.1\linewidth}
    \caption*{Parameter \\networks}
		\includegraphics[height=3cm] {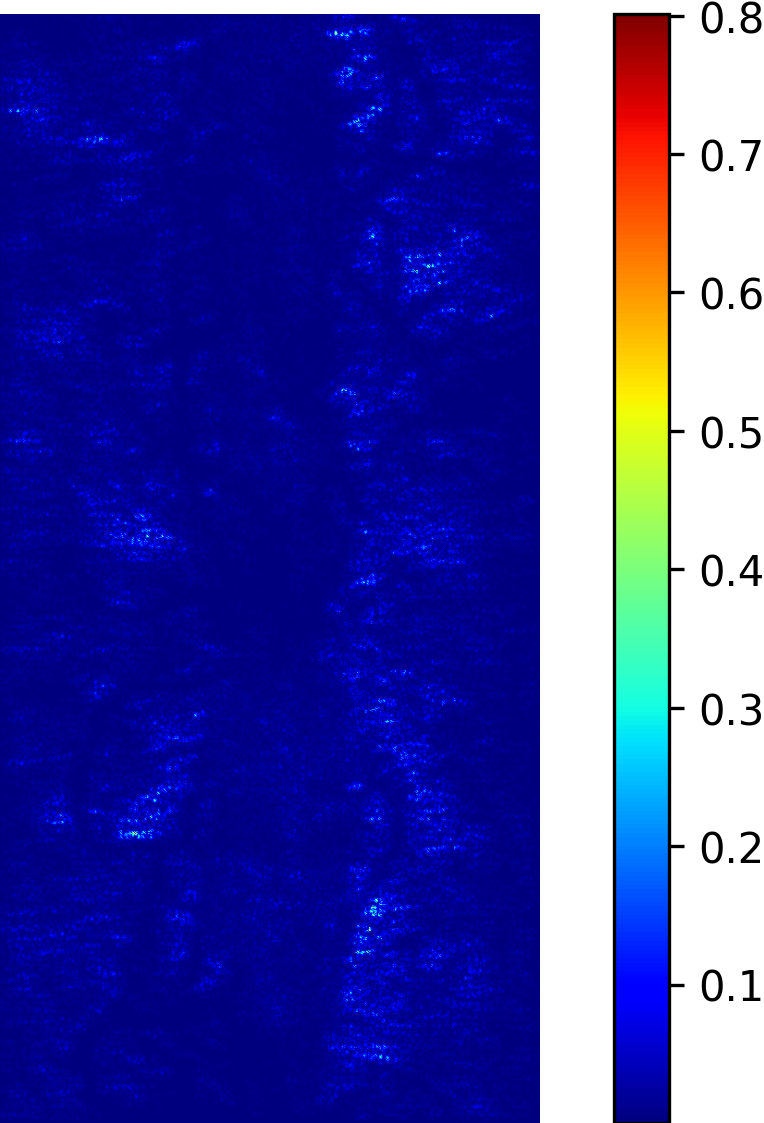} \\[3pt]
		\includegraphics[height=3cm] {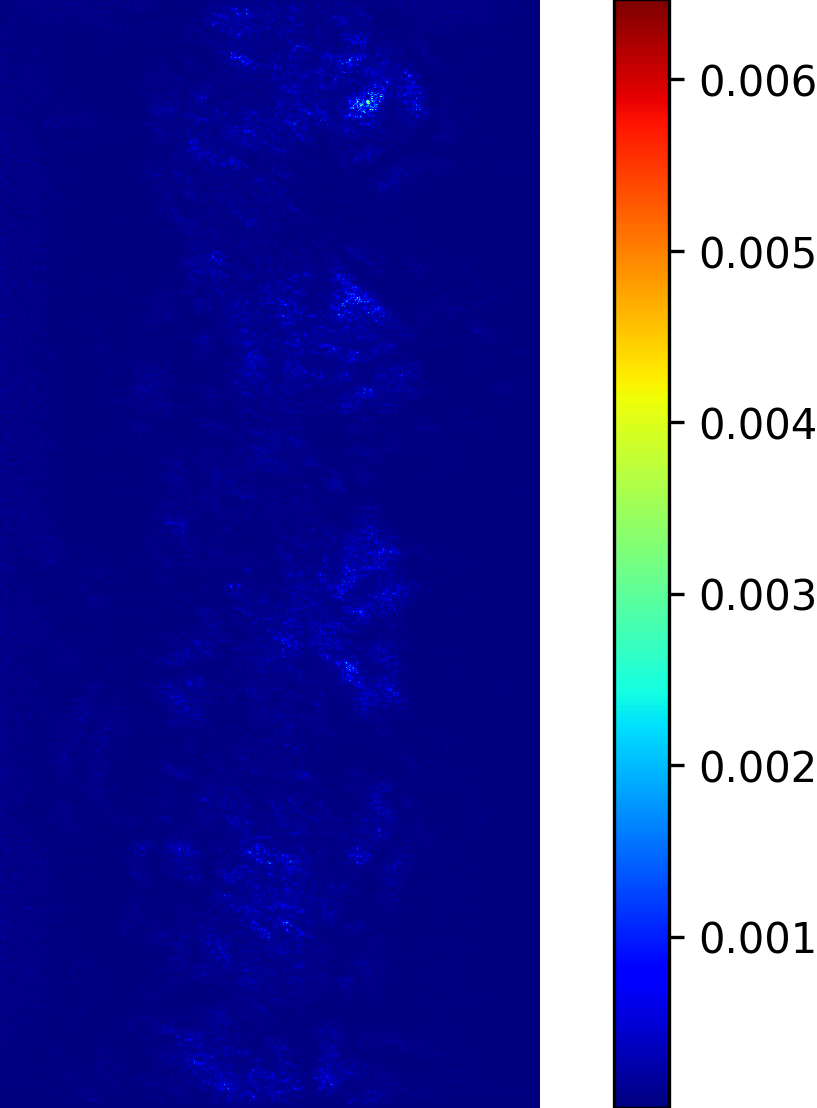} 
		\end{subfigure}
		\hfil
	\begin{subfigure}{0.1\linewidth}
		\caption*{$Reg$ \\network}
		\includegraphics[height=3cm] {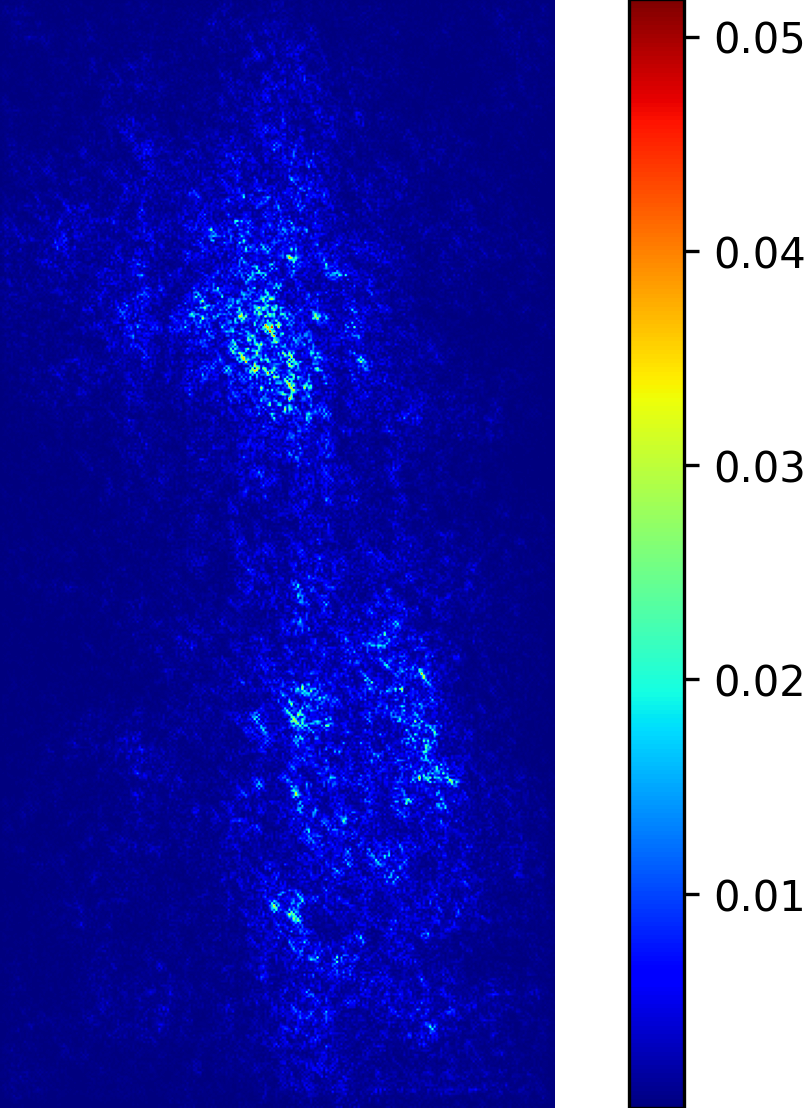}\\[3pt]
		\includegraphics[height=3cm] {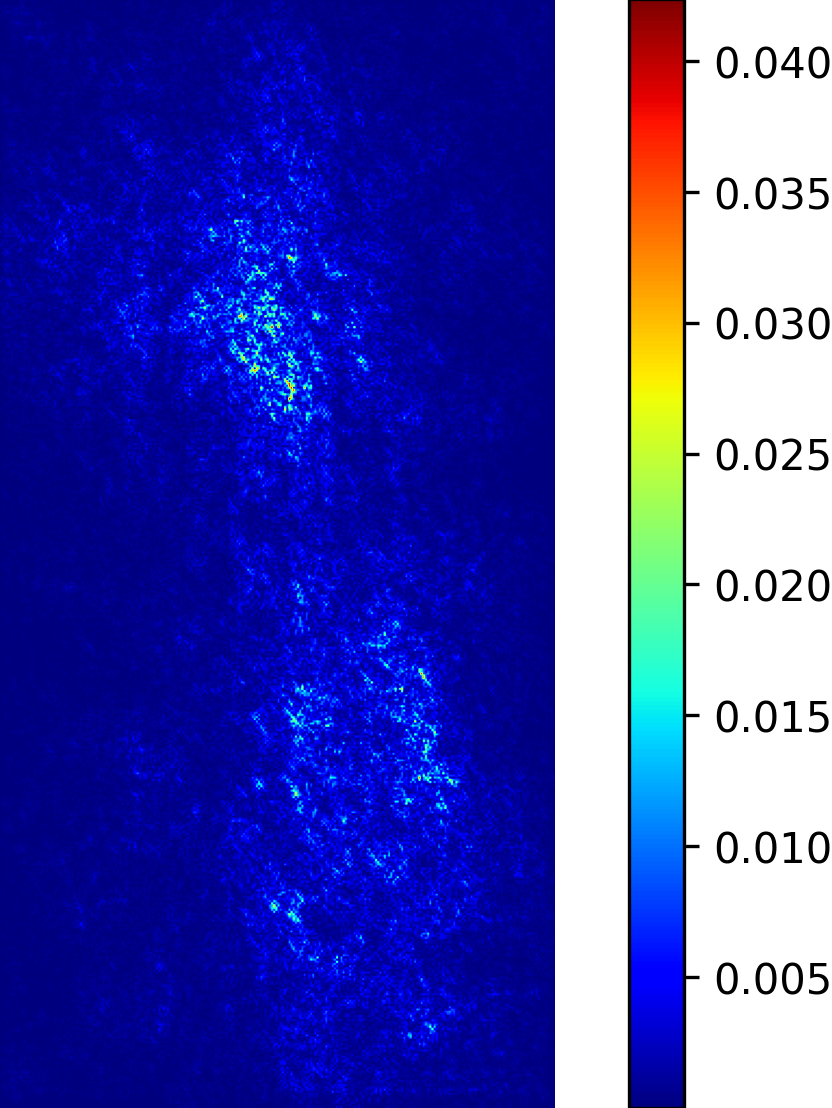} 
				\end{subfigure}
		\hfil
			\begin{subfigure}{0.1\linewidth}
		\caption*{Input}
		\includegraphics[height=3cm] {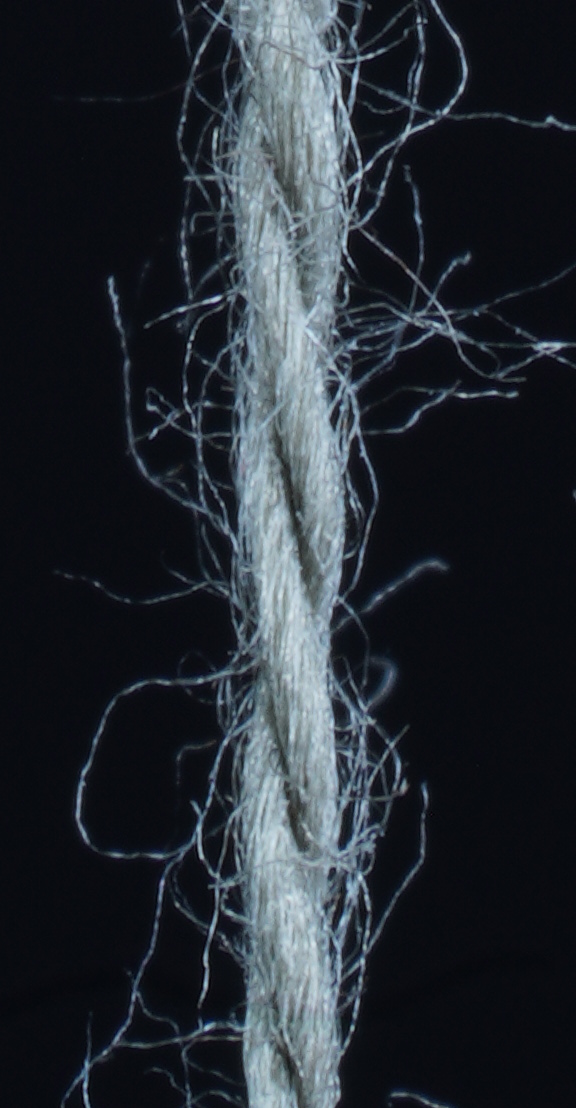}\\[3pt]
		\includegraphics[height=3cm] {figures/input_oliv.png} 
				\end{subfigure}
		\hfil
	\caption{Sailiency maps for the yarn twist pitch $\alpha_{ply}$ (top row) and the yarn radius $R_{ply}$ (bottom row).}
	\label{fig:saliency_geom}
\end{figure}

Whereas small modifications of the parameters of the basic yarn structure (i.e. without flyaways) already have a significant visual effect on the resulting yarn (see Fig.~\ref{fig:results_different_relolutions} for different values of the parameter $\alpha_{ply}$), small variations of the parameters for the flyaway characteristics do not have such a significant impact on the generated yarn since only the distribution of the flyaway characteristics has to be matched to obtain plausible flyaways.
For this reason, we compared the saliency maps from models trained for different raw yarn parameters individually (e.g., Fig.~\ref{fig:saliency_geom} shows the maps for the yarn radius and parameter $\alpha_{ply}$) and found that they differ. Motivated by the results of the individual saliency maps, we investigated the use of separate networks for the separate prediction of each of the raw yarn parameters. A similar separation of models has already been observed by Nishida et al.~\cite{Nishida18}.
We compared the previously described approach of using separate networks to predict the raw yarn parameters and the flyaway characteristics with the approach of using separate networks to predict each raw yarn parameter separately along with a network to predict the flyaway characteristics.
In this context, we leveraged further priors for some of the parameters to exploit their underlying nature.
For the parameter \emph{number of plys}, we changed the last layer from the identity function as used for regression to a softmax function, thereby framing the prediction of this discrete parameter as a classification problem.
The underlying motivation is that most knitting yarns have 2 to 6 plys and the estimation of the number of plys based a classification might be easier that based on a regression.
Furthermore, we distinguish fibers according to their elliptic cross-section characteristics into \emph{thin fibers} ($t_x = 0.01$, $t_y = 0.007$) and \emph{thick fibers} ($t_x = 0.018$, $t_y = 0.01$),
which we also frame as a classification problem since considering all intermediate states seems tricky and there seems to be no such significant perceptual difference for these intermediate states.

%% file: results.tex
\section{Experiments}
\paragraph{Training, validation and test data}
We use 4000 synthetic yarns for training and 345 synthetic yarns for validation as mentioned in Section~\ref{ssec:dataset}.
To get insights on the performance of our method for parameter inference for real yarns depicted in photographs, we tested our approach on different knitting yarns, which were made either of one type of fiber such as wool, acrylic, cotton, polyamide, etc. or of a mix of different types of fibers (Fig.~\ref{fig:results_different_yarns}, top row, second yarn).
\begin{figure*}[h]
	\center
	\includegraphics[height=2.8cm]{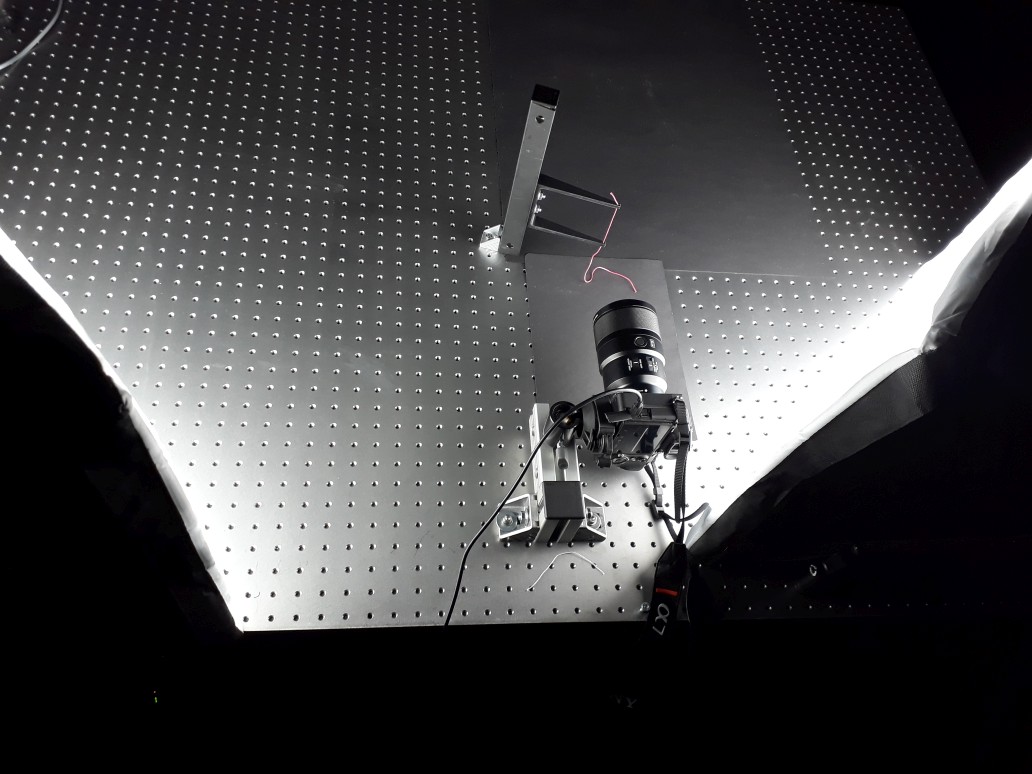}
	\includegraphics[height=3.1cm]{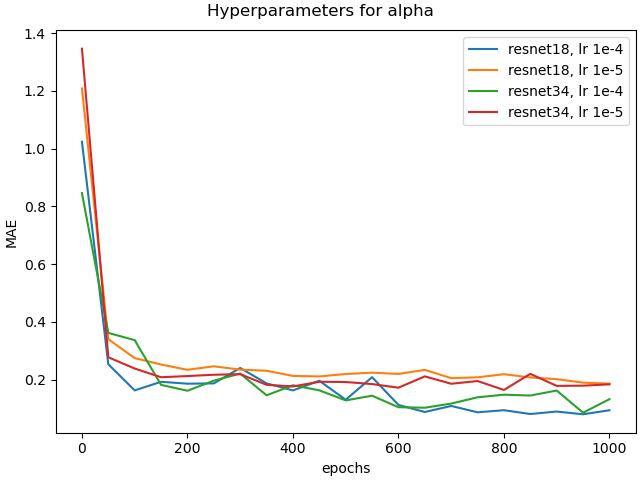}
	\includegraphics[height=3.1cm]{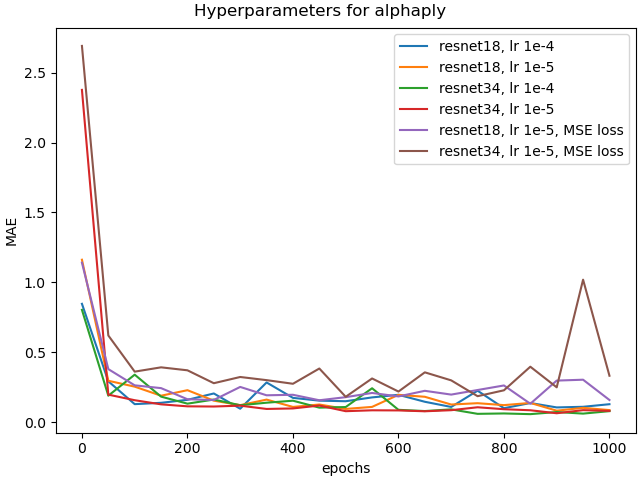}
	\includegraphics[height=3.1cm]{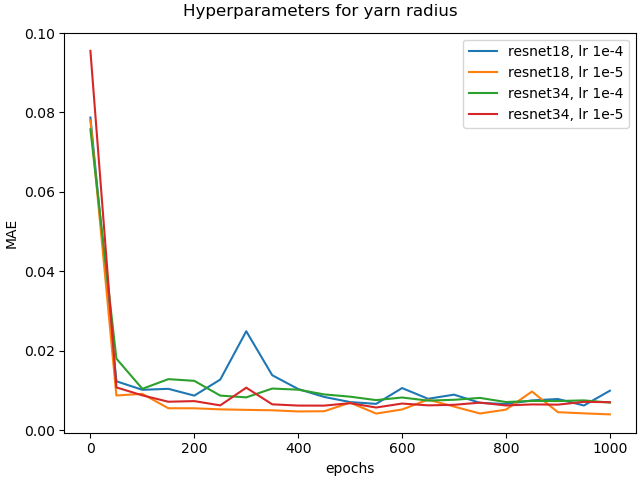}
	
		\hspace{0.1cm} a) \hspace{4cm} b) \hspace{4cm} c) \hspace{4.0cm} d)
		
	\caption{a) Our setup for capturing the test yarns. b)-d) Examples of vallidation loss comparissons for hyperparameter determination for parameters $\alpha$ (b), $\alpha_{ply}$ (c) and $R_{ply}$ (d). Based on the loss values we chose the model of ResNet18, learning rate = $1e{^{-4}}$ and epoch 850 for $\alpha$, ResNet34, learning rate = $1e{^{-4}}$ and epoch 850 for $\alpha_{ply}$ and ResNet18, learning rate = $1e{^{-5}}$ and epoch 1000 for $R_{ply}$.}
	\label{fig:hyperparams}
\end{figure*}
We took the corresponding photos of the yarns under a simple lab setup (see Fig.~\ref{fig:hyperparams}, a) with a Sony $\alpha 7$R\MakeUppercase{\romannumeral 3} camera using the makrolens Makro G OSS with FE 90 mm F2.8.
Then we cropped the photos to the size of 600 $\times$ 2000 pixels, ensuring that the yarn roughly runs through the center of the image.
These cropped photos served as an input for the parameter inference.
As our networks were trained for inputs of 1200 times 584 pixels, we again crop the previously cropped photos randomly to the required size before performing the forward pass and hence determining the parameters.
\paragraph{Details of the training process}
To improve the robustness of the trained models, we increase the variety of the training data by randomly cropping the 4000 images of a size of 2000x600 pixels to the network input size of 1200x584 pixels during each epoch.
Then we run the training for 1000 epochs with a batch size of 32 and a learning rate of 0.0001 based on the Adam optimizer~\cite{kingma-2014-adam}.
For this purpose, we used three Nvidia Titan XP GPUs, each having 12 GB of RAM.
Based on this hardware, the training for the flyaway network and the full regression network took each approx. 11 hours. When training only for one parameter, the training for the ResNet34 took ca. 4 hours, while for the ResNet18 it was 2.5 hours.
\subsection{Parameter inference on real data}
\paragraph{Validation of training process}
First, we need to validate whether the trained model shows the potential to perform well on synthetic validation data.
For this purpose, we compared the inference of yarn parameters for validation data through a set of different models against one model for all parameters.
In this scope, we compared the the approaches of using two separate networks for predicting the raw yarn parameters and flyaway characteristics and using separate parameter specific networks for predicting each individual raw yarn parameter separately together with a network for predicting the flyaway characteristics, and have chosen the best hyperparameters and the best epoch based on the validation loss computed on synthetic validation set.
Figure~\ref{fig:hyperparams} illustrates the validation losses for the twisting parameters $\alpha$, $\alpha_{ply}$ and yarn radius $R_{ply}$.
Table~\ref{tab_diffnets} shows the comparison of the best models of every case.
We can see that the loss over each parameter is bigger when training one model for all parameters of the raw yarn instead of training specific models with different hyperparameters for each parameter separately. While this indicates a better capability to infer yarn parameters on synthetic data, we did not yet analyze the generalization to images depicting real yarns, which will follow with the experiments regarding performance analysis on real data.
\begin{table}[h]
\scriptsize
\caption{Validation loss of different networks. Note that especially the important yarn twist parameters $\alpha$,  $\alpha_{ply}$ and $R_{ply}$ are better learned with parameter specific networks.}
%\label{table}
\setlength{\tabcolsep}{5pt}
\begin{tabular}{|p{20pt}|p{30pt}|p{30pt}|p{30pt}|p{30pt}|p{30pt}|}
\hline
& parameter specific networks & $Reg$ & $Reg^{ae}$&$Reg_{latent}$ & $Reg^{ae}_{latent}$ \\
\hline
$r_x$ 	&0.0080 		&0.0082 	&0.0080 	&0.0097		&0.0087\\
$r_y$ 	&0.0066 		&0.0066  &0.0074	&0.0083 		&0.0079\\
$m$ 		&12 	&12 &13 	&14		&14\\
$\alpha$ 				&0.0807 		&0.2493  &0.2587	&0.3230 		&0.3026\\
$\alpha_{ply}$ 			&0.0614 		&0.1953  &0.2101 	&0.2400 		&0.2433\\
$R_{ply}$ 	&0.00444 		&0.0082  &0.0092 	&0.0092 		&0.0095\\
\hline
\end{tabular}
\label{tab_diffnets}
\end{table}
\paragraph{Performance evaluation}
We now provide an evaluation of the performance of the different approaches of using a single network for predicting all parameters of the raw yarn and a network for the prediction of the flyaway parameters when using different loss formulations as discussed before in comparison to using also separate networks for the prediction of the raw yarn parameters. This means the model for prediction of the flyaway parameters is the same for all these approaches. In our experiments, the flyaway model with the lowest validation loss was the ResNet18 model trained with a learning rate of 0.001 and MAE loss, which we therefore use for the subsequent experiments.
Exemplary results of our experiments including a comparison between the investigated approaches can be observed in Figure~\ref{fig:results_different_yarns}.
The corresponding inferred parameters are presented in the supplemental material.
The renderings of the parameters inferred from parameter specific networks for each parameter of the raw yarn look more similar to the input image, than the renderings from the other approaches.Since we do not have the ground truth parameters for our real world yarns, we can only compare the geometry appearance of the yarns. 
Based on the appearance comparison to the input image, we conclude that the approach of the parameter specific networks is most suitable for the given task. 

\begin{figure}[h]
	\center
	\includegraphics[height=1.9cm]{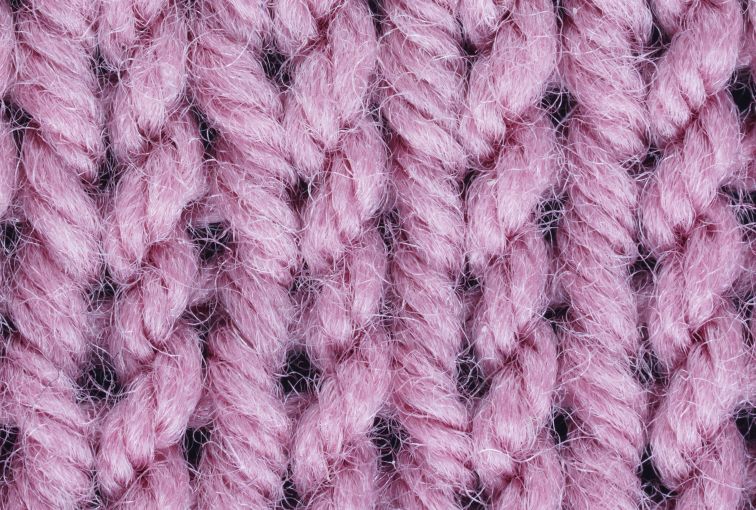}
	\includegraphics[height=1.9cm]{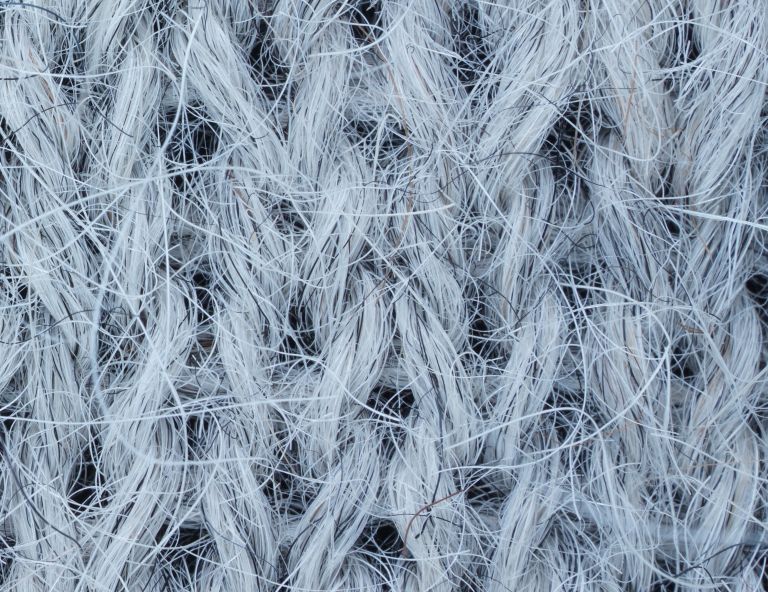}
	\includegraphics[height=1.9cm]{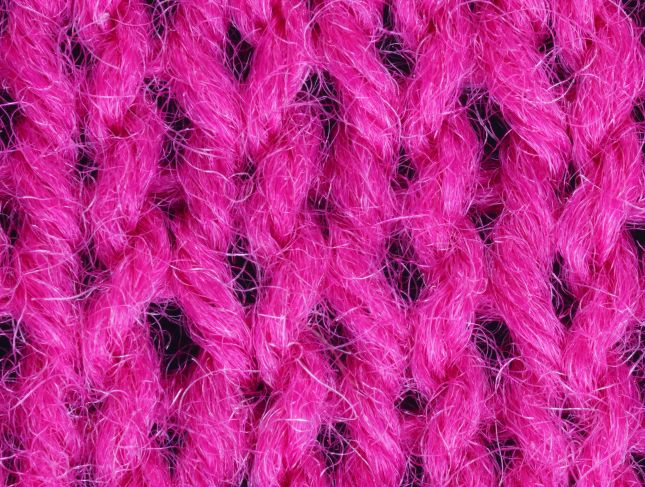}\\ \vspace{0.05cm}
	\includegraphics[height=1.9cm]{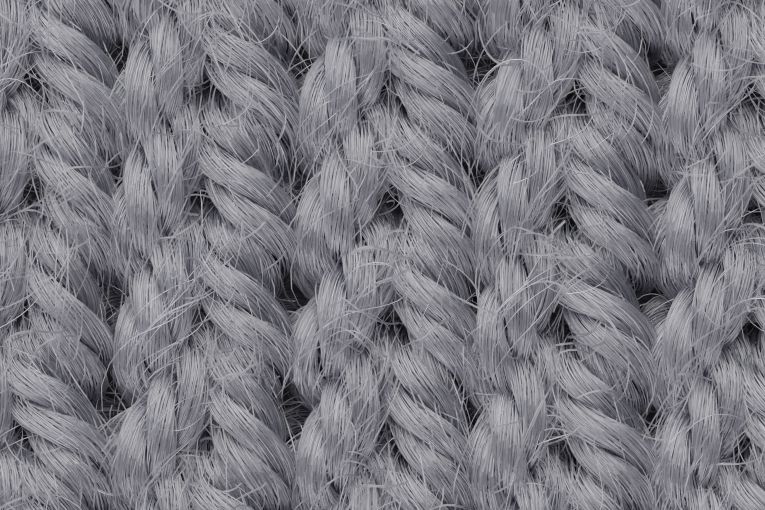}
	\includegraphics[height=1.9cm]{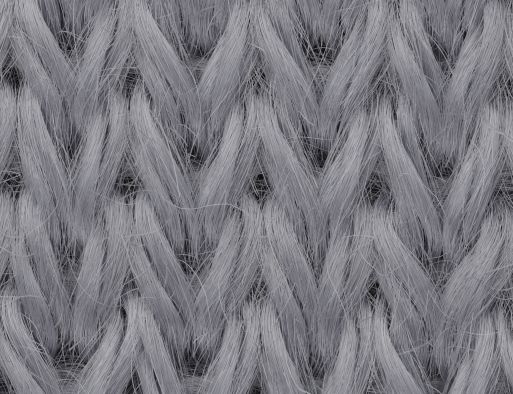}
	\includegraphics[height=1.9cm]{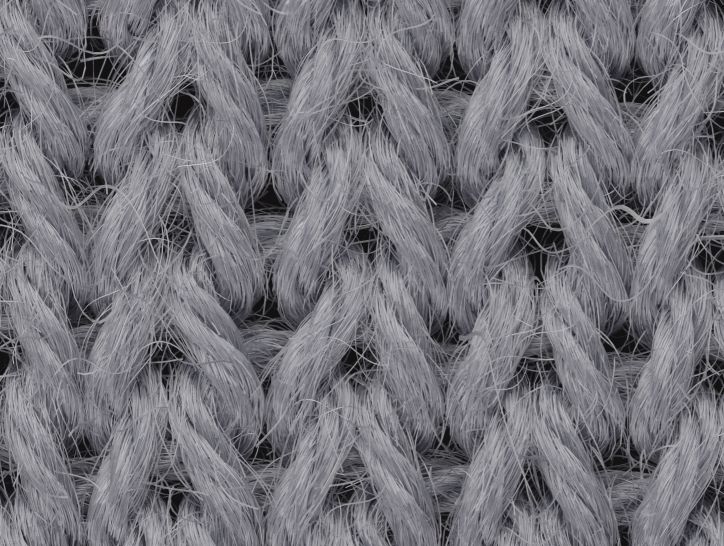}\\\vspace{0.05cm}
		\includegraphics[height=1.85cm]{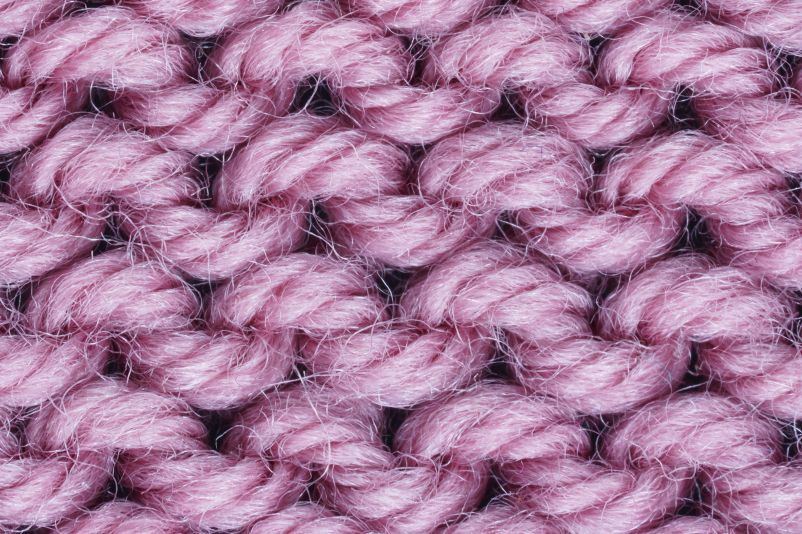}
	\includegraphics[height=1.85cm]{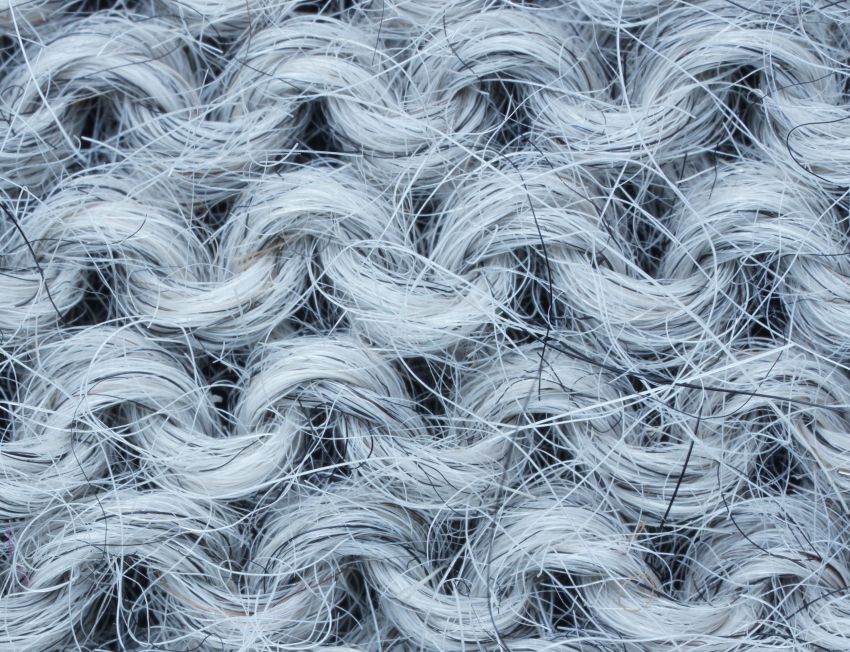}
	\includegraphics[height=1.85cm]{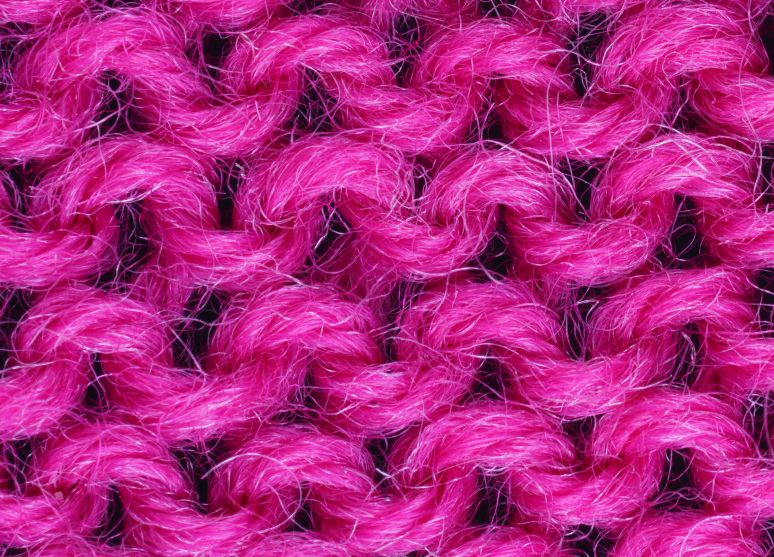}\\\vspace{0.05cm}
	\includegraphics[height=1.85cm]{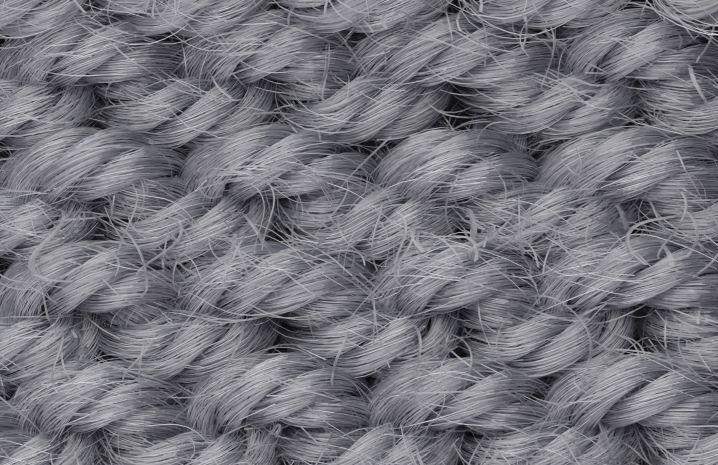}
	\includegraphics[height=1.85cm]{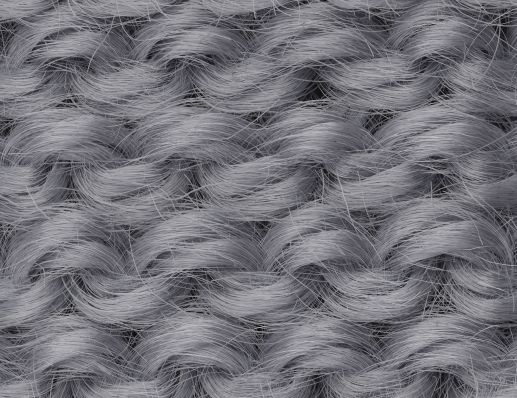}
	\includegraphics[height=1.85cm]{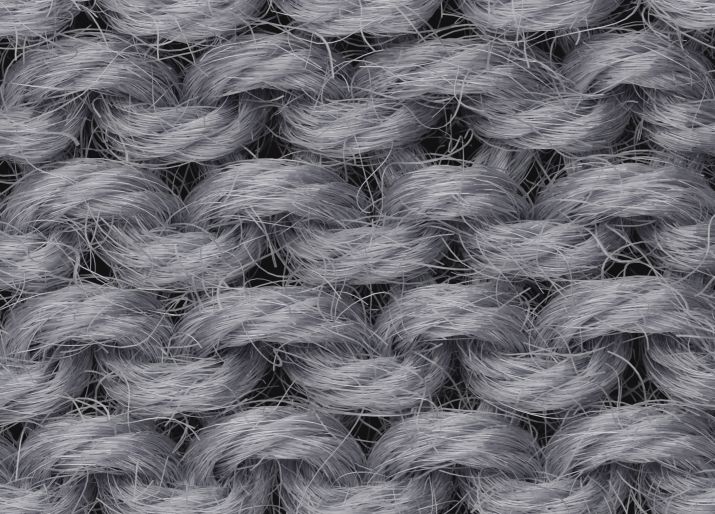}
	\caption{1st and 3rd rows: images of a real knitted cloth (made with yarns from the top row of Fig.~\ref{fig:results_different_yarns}) for the pattern consisting of knit (1st row) and purl (3rd row) stitches. 2nd and 4th rows: rendering of the same stitch pattern with the inferred yarn with default material settings.}
	\label{fig:results_patterns_knit}
\end{figure}

Additionally, we utilize the parameter inference for renderings of knitting samples.
In Figure~\ref{fig:results_patterns_knit}, we show the renderings of knitting samples, made with the three yarns of the top row from Figure~\ref{fig:results_different_yarns} with the parameters from the ensemble of per-parameter networks for the raw yarn structure.
\begin{figure*}[ht]
	\center
	\begin{subfigure}{0.05\linewidth}
    \caption*{in}
		\includegraphics[height=2.8cm] {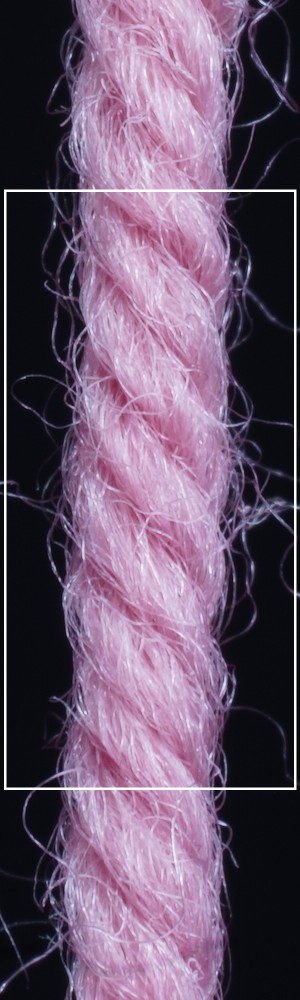} \\[3pt]
		\includegraphics[height=2.8cm] {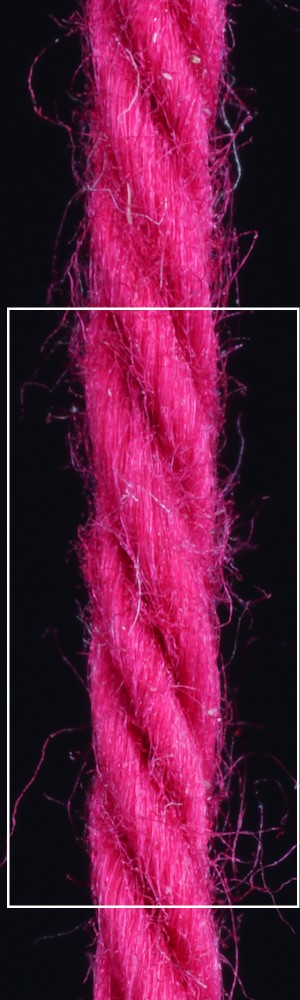} \\[3pt]
		\includegraphics[height=2.8cm] {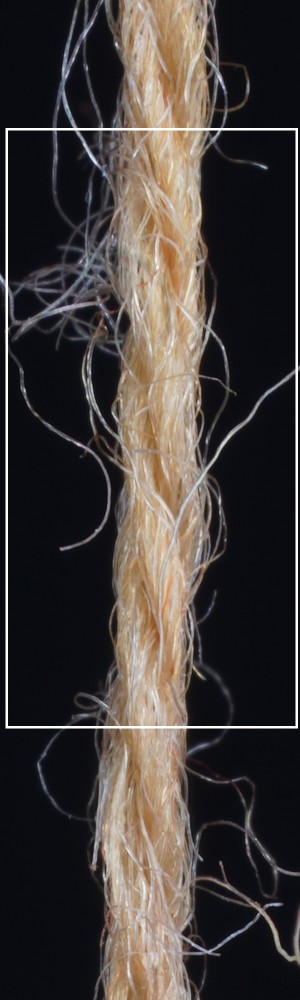} \\[3pt]
		\includegraphics[height=2.8cm] {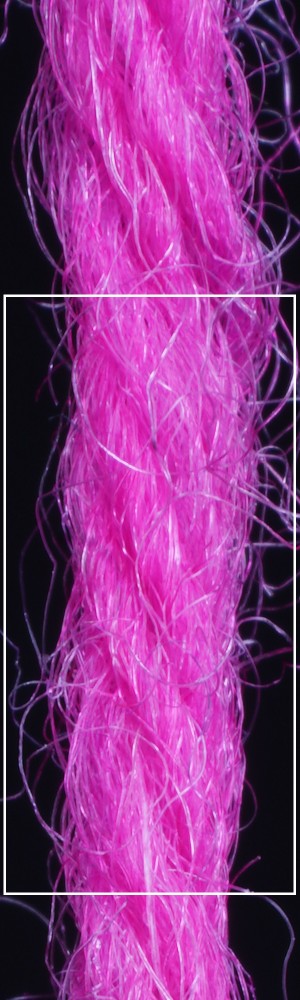} 
		\end{subfigure}
		\hfil
	\begin{subfigure}{0.05\linewidth}
		\caption*{1}
		\includegraphics[height=2.8cm] {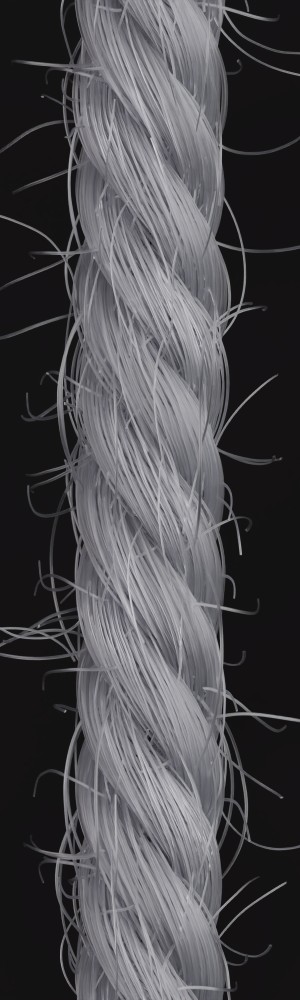}\\[3pt]
		\includegraphics[height=2.8cm] {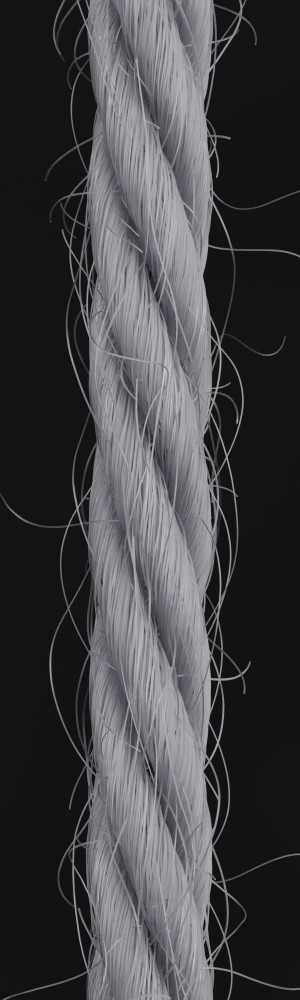}\\[3pt]
				\includegraphics[height=2.8cm] {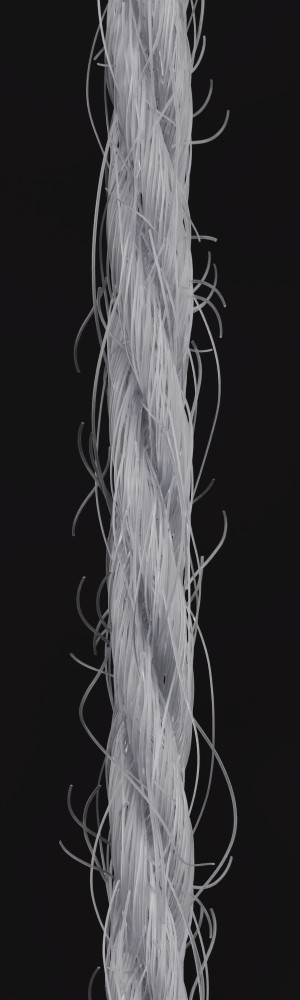}\\[3pt]
		\includegraphics[height=2.8cm] {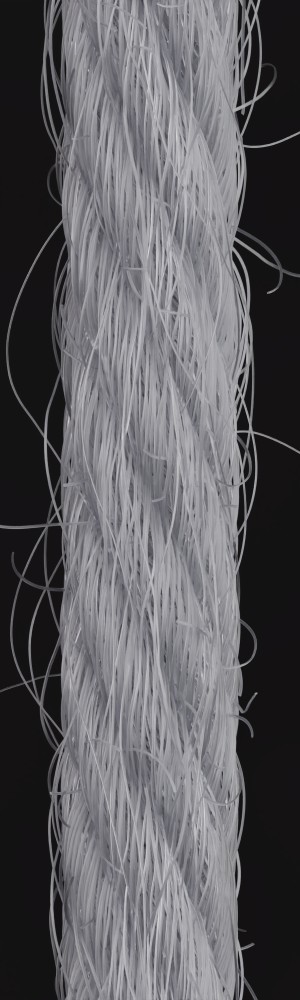}
				\end{subfigure}
		\hfil
	\begin{subfigure}{0.05\linewidth}
		\caption*{2}
				\includegraphics[height=2.8cm] {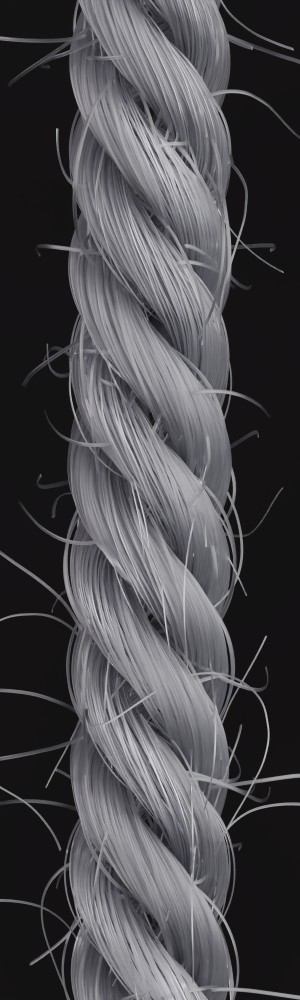} \\[3pt]
		\includegraphics[height=2.8cm] {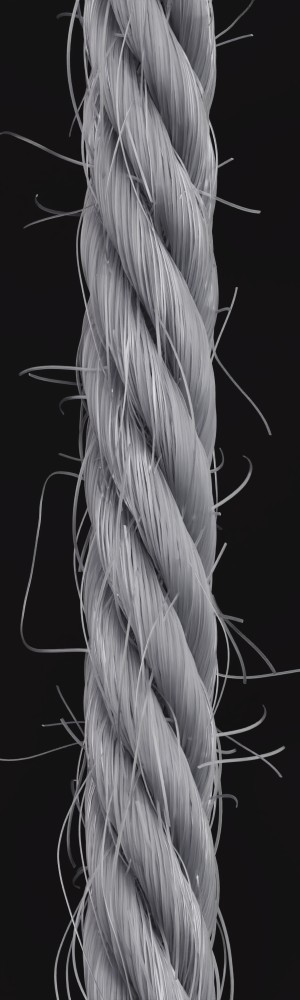}\\[3pt]
						\includegraphics[height=2.8cm] {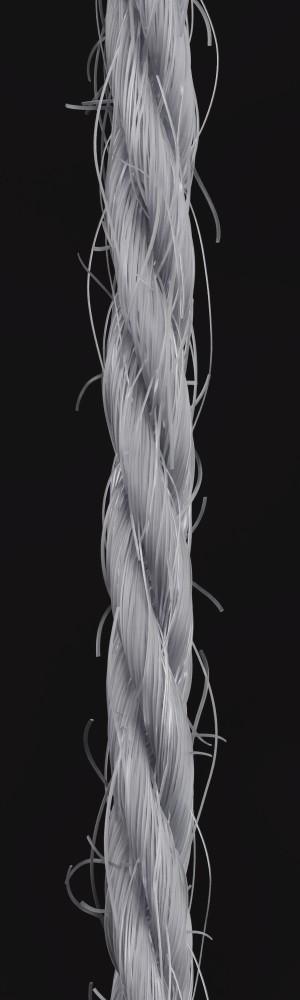} \\[3pt]
		\includegraphics[height=2.8cm] {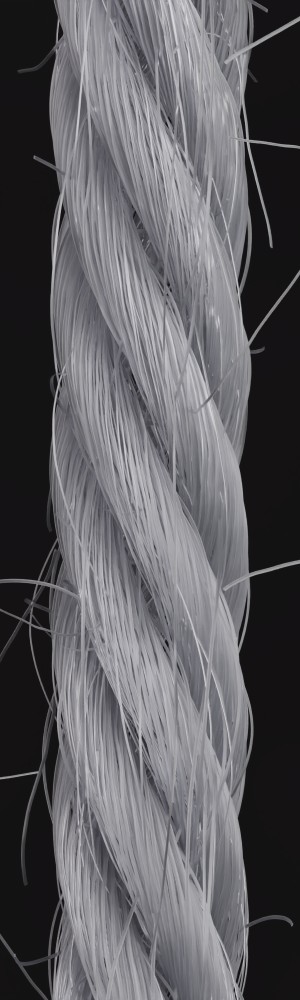}
				\end{subfigure}
		\hfil
	\begin{subfigure}{0.05\linewidth}
		\caption*{3}
		\includegraphics[height=2.8cm] {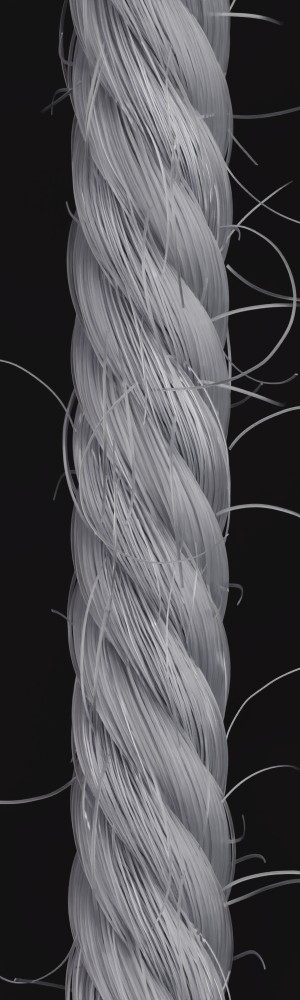}  \\[3pt]
		\includegraphics[height=2.8cm] {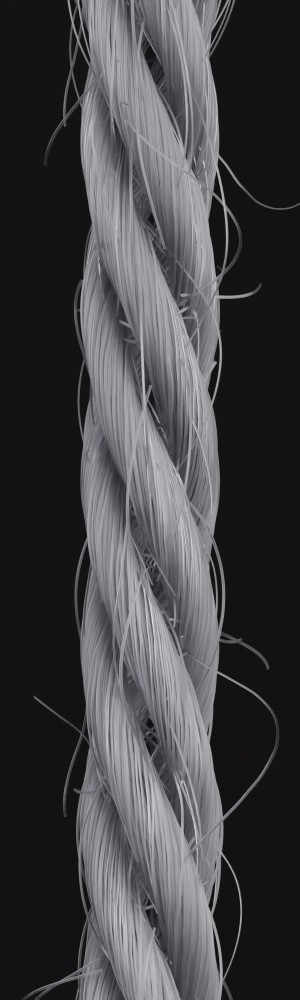}  \\[3pt]
				\includegraphics[height=2.8cm] {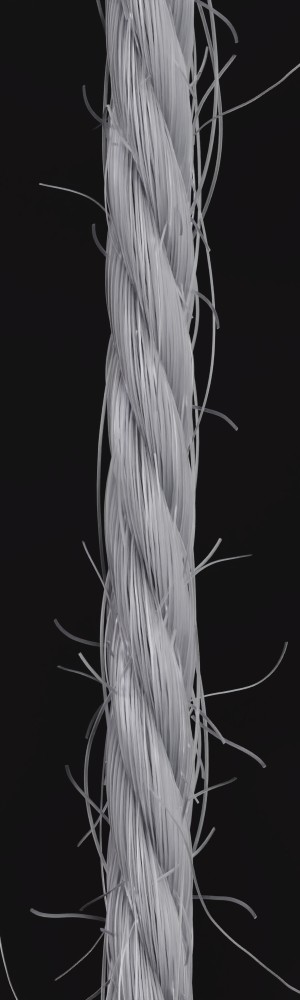}  \\[3pt]
		\includegraphics[height=2.8cm] {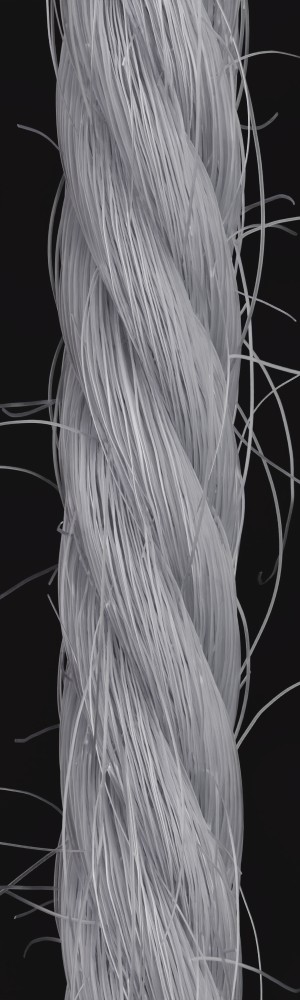}  
				\end{subfigure}
		\hfil
			\begin{subfigure}{0.05\linewidth}
		\caption*{4}
				\includegraphics[height=2.8cm] {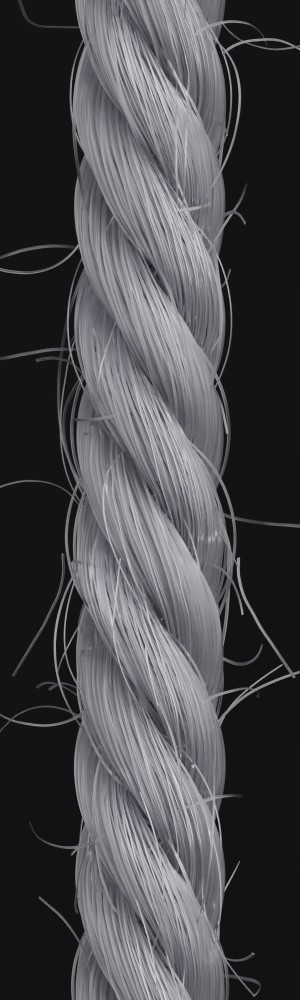} \\[3pt]
		\includegraphics[height=2.8cm] {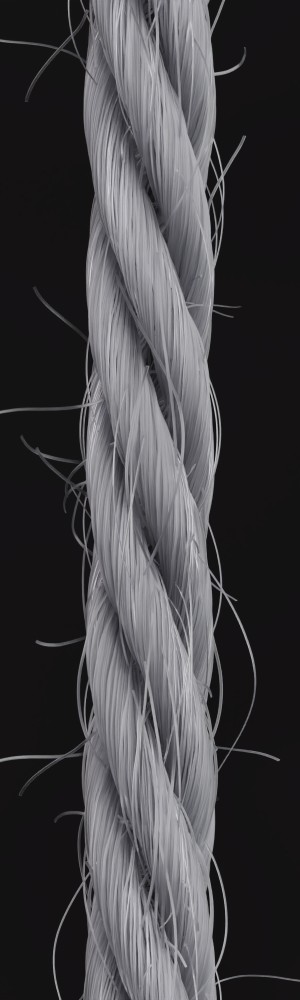} \\[3pt]
						\includegraphics[height=2.8cm] {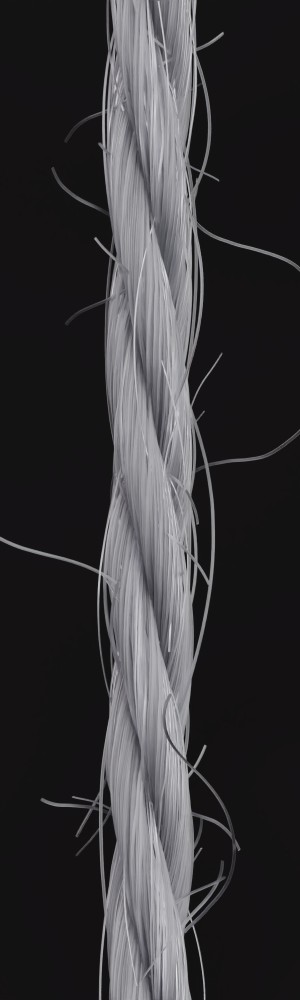} \\[3pt]
		\includegraphics[height=2.8cm] {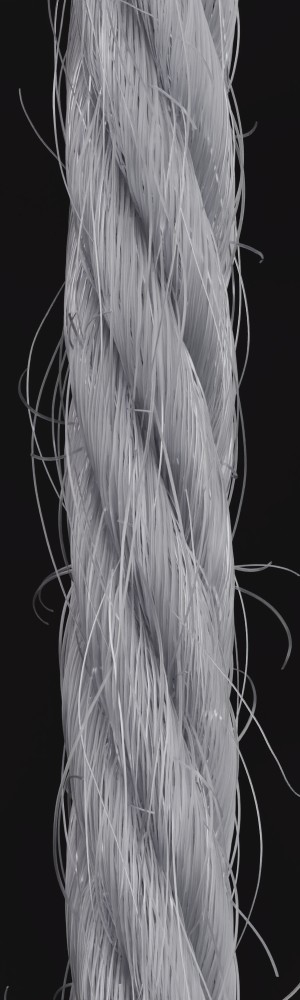} 
				\end{subfigure}
		\hfil
	\begin{subfigure}{0.05\linewidth}
		\caption*{5}
				\includegraphics[height=2.8cm] {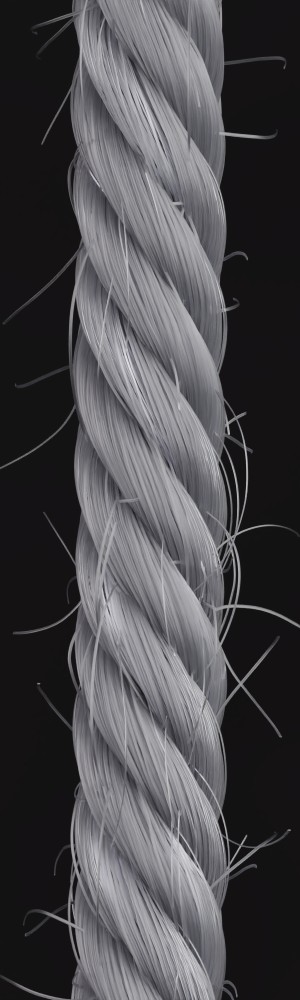}  \\[3pt]
		\includegraphics[height=2.8cm] {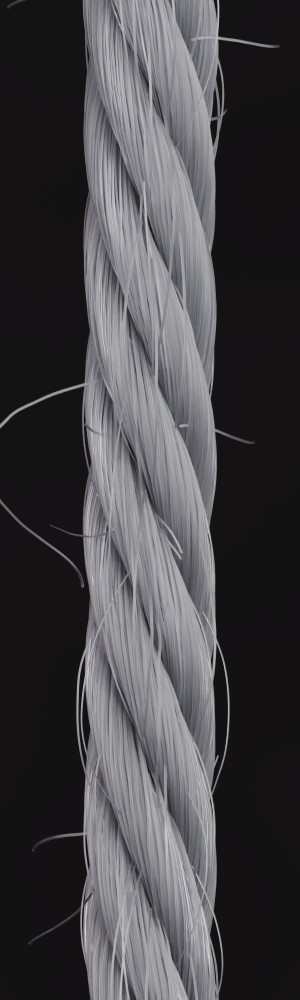} \\[3pt]
						\includegraphics[height=2.8cm] {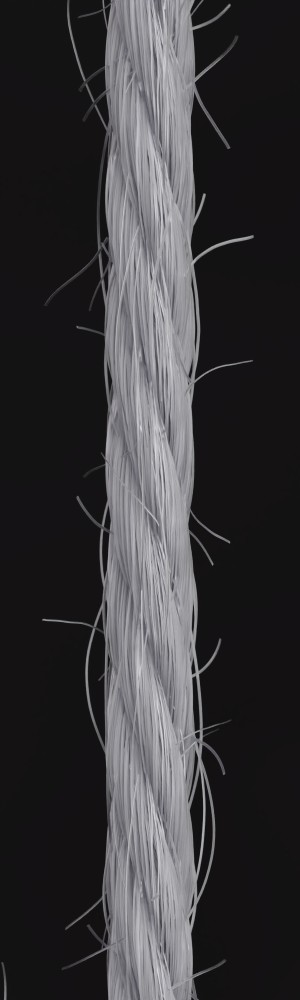}  \\[3pt]
		\includegraphics[height=2.8cm] {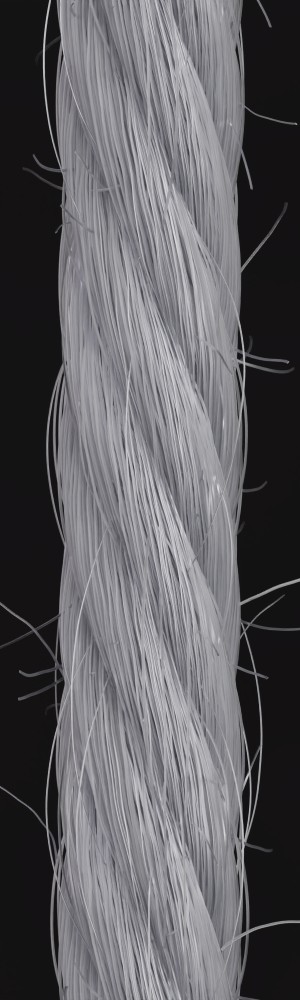} 
				\end{subfigure}
	\hspace{0.025cm}	\hfil
		\begin{subfigure}{0.05\linewidth}
    \caption*{in}
		\includegraphics[height=2.8cm] {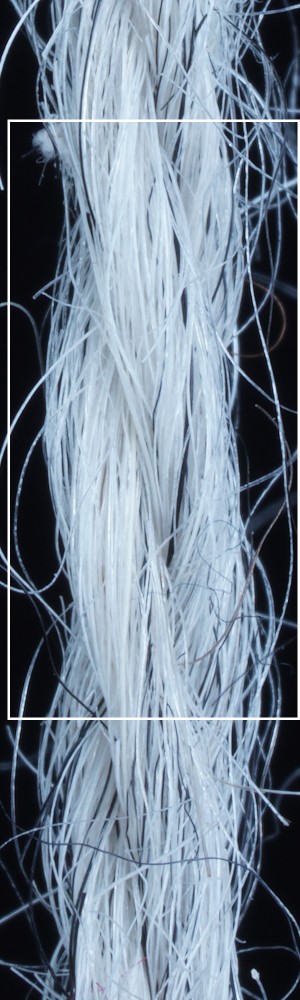} \\[3pt]
		\includegraphics[height=2.8cm] {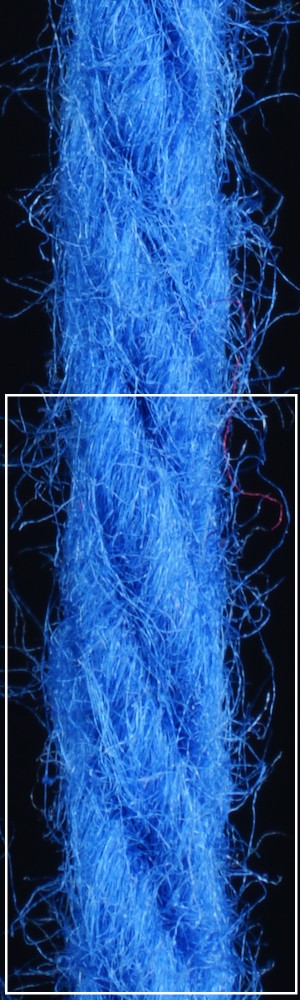} \\[3pt]
				\includegraphics[height=2.8cm] {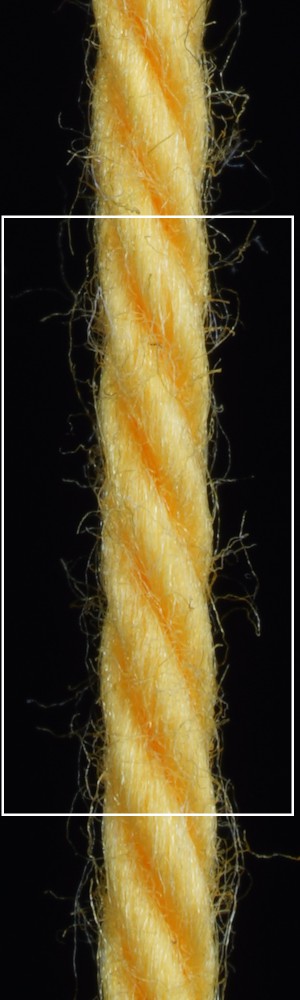} \\[3pt]
		\includegraphics[height=2.8cm] {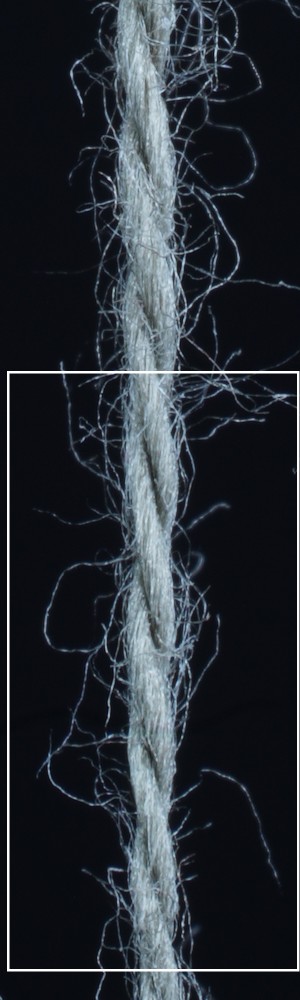} 
		\end{subfigure}
		\hfil
	\begin{subfigure}{0.05\linewidth}
		\caption*{1}
		\includegraphics[height=2.8cm] {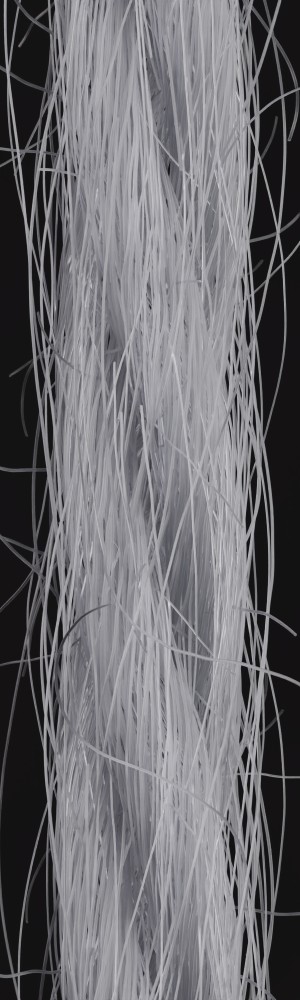}\\[3pt]
		\includegraphics[height=2.8cm] {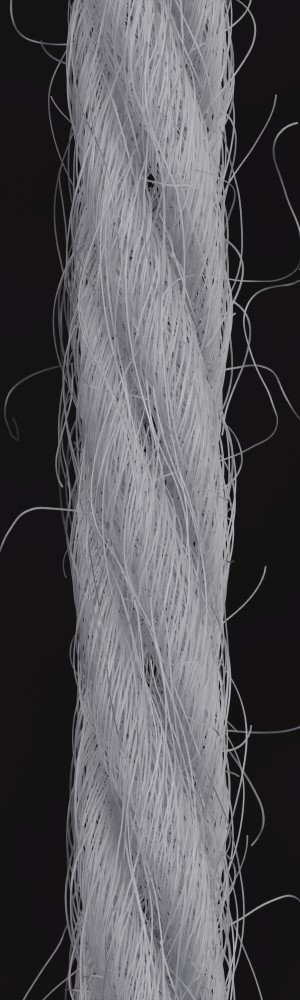}\\[3pt]
				\includegraphics[height=2.8cm] {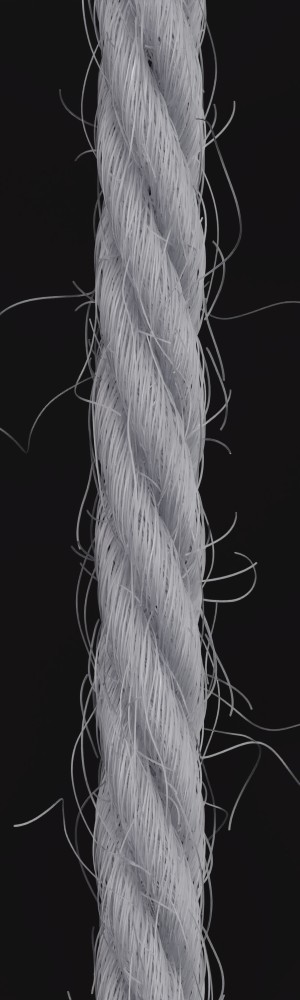}\\[3pt]
		\includegraphics[height=2.8cm] {figures/yarns/2er_oliv_diffnets_small.jpg}
				\end{subfigure}
		\hfil
	\begin{subfigure}{0.05\linewidth}
		\caption*{2}
				\includegraphics[height=2.8cm] {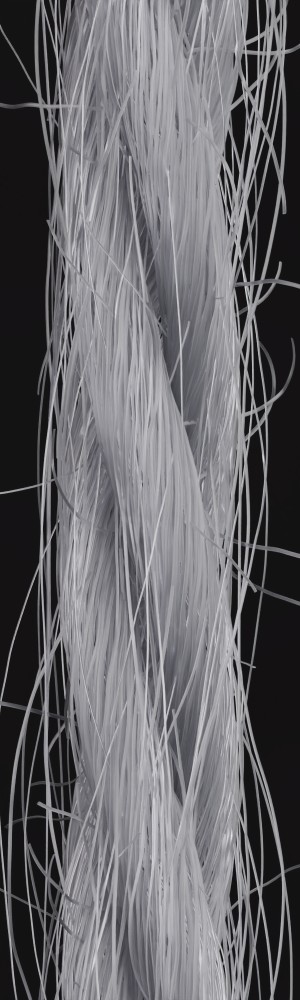} \\[3pt]
		\includegraphics[height=2.8cm] {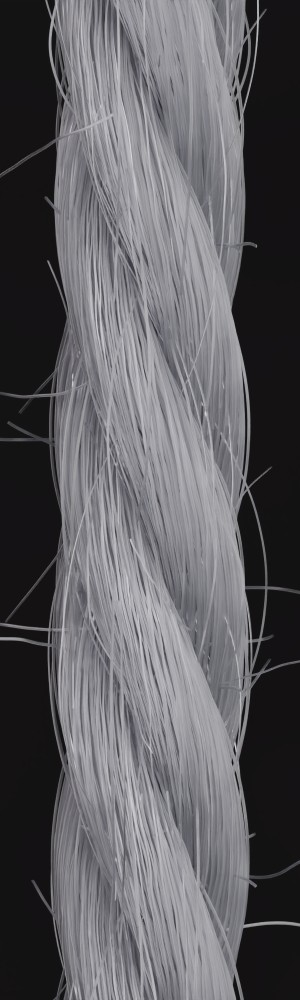}\\[3pt]
						\includegraphics[height=2.8cm] {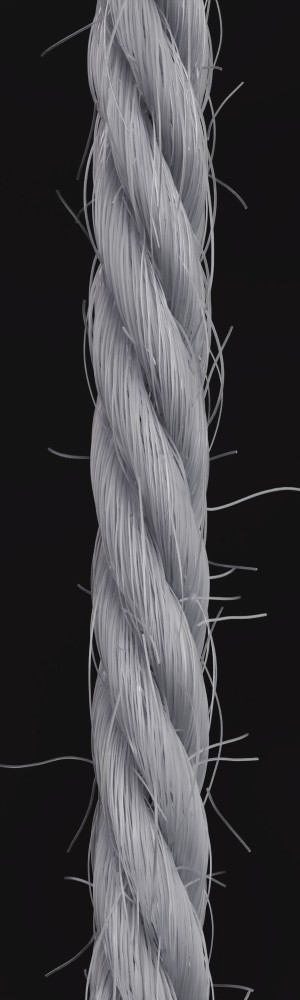} \\[3pt]
		\includegraphics[height=2.8cm] {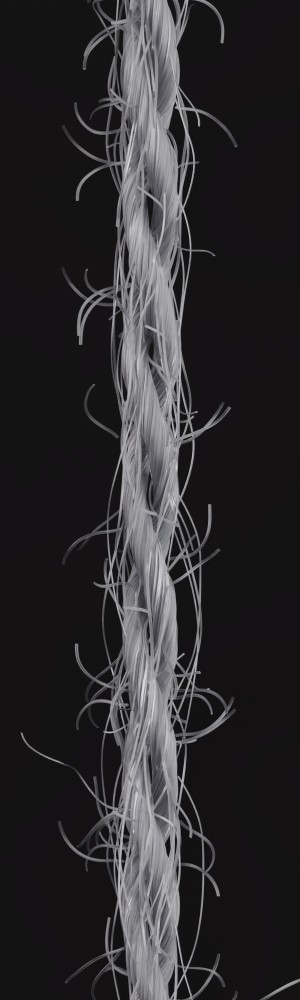}
				\end{subfigure}
		\hfil
	\begin{subfigure}{0.05\linewidth}
		\caption*{3}
		\includegraphics[height=2.8cm] {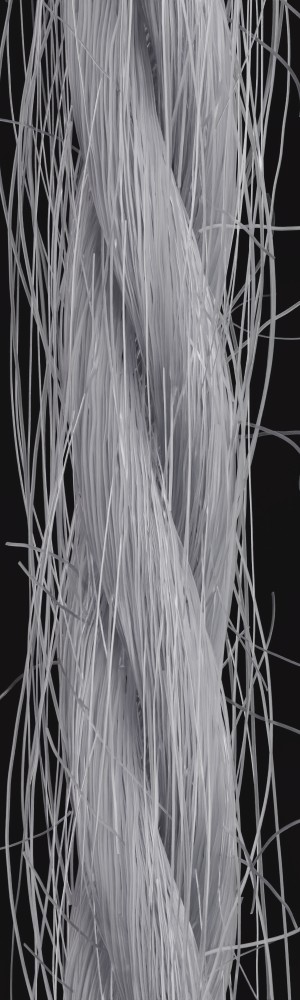}  \\[3pt]
		\includegraphics[height=2.8cm] {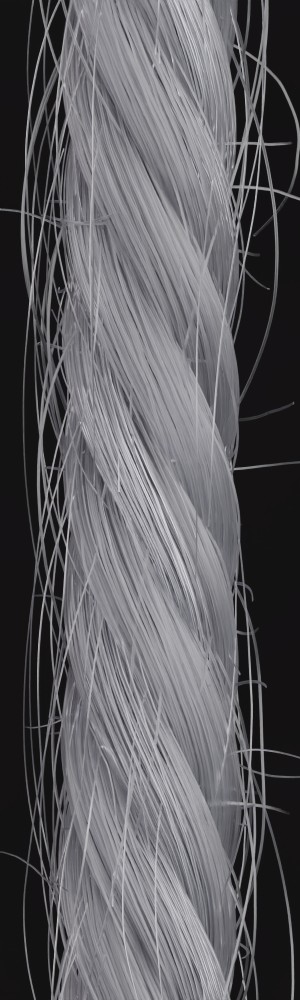}   \\[3pt]
				\includegraphics[height=2.8cm] {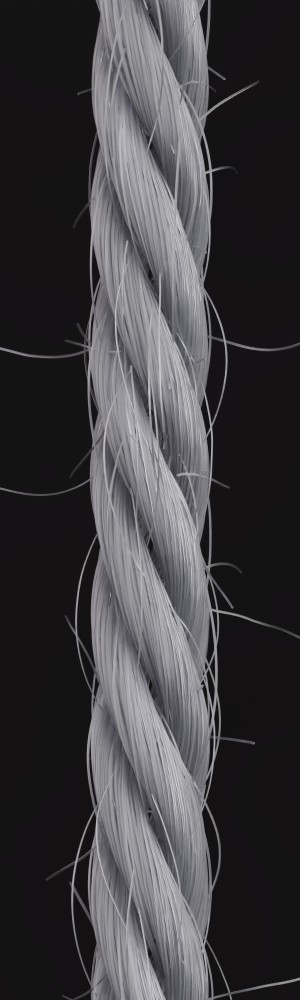}  \\[3pt]
		\includegraphics[height=2.8cm] {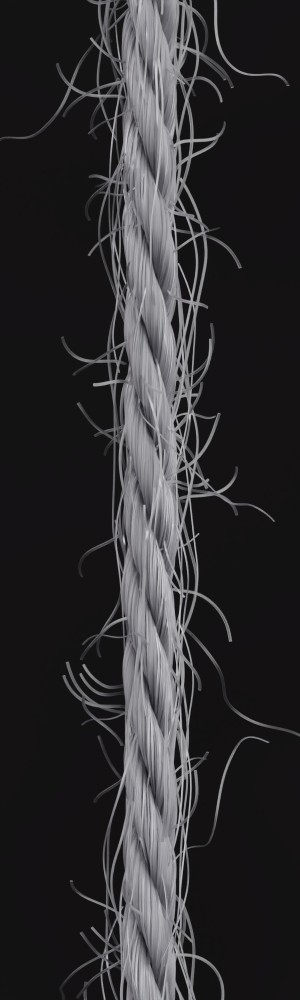}
				\end{subfigure}
		\hfil
			\begin{subfigure}{0.05\linewidth}
		\caption*{4}
				\includegraphics[height=2.8cm] {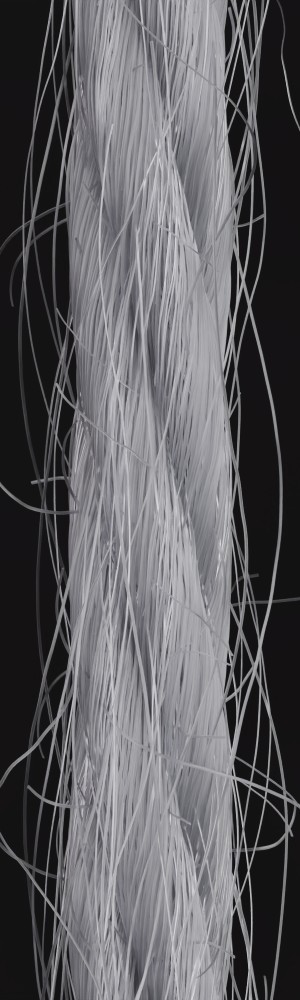} \\[3pt]
		\includegraphics[height=2.8cm] {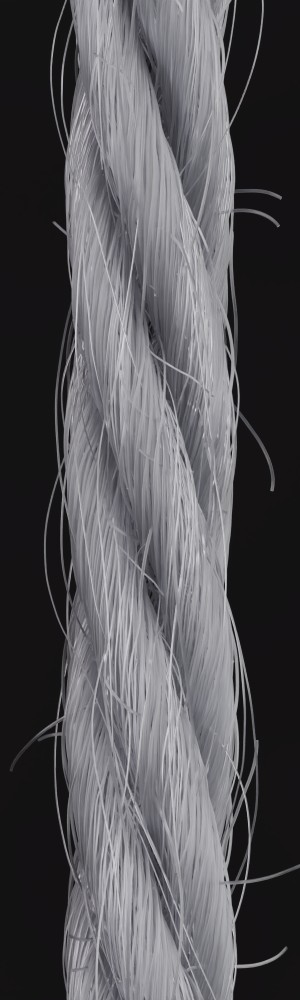}  \\[3pt]
						\includegraphics[height=2.8cm] {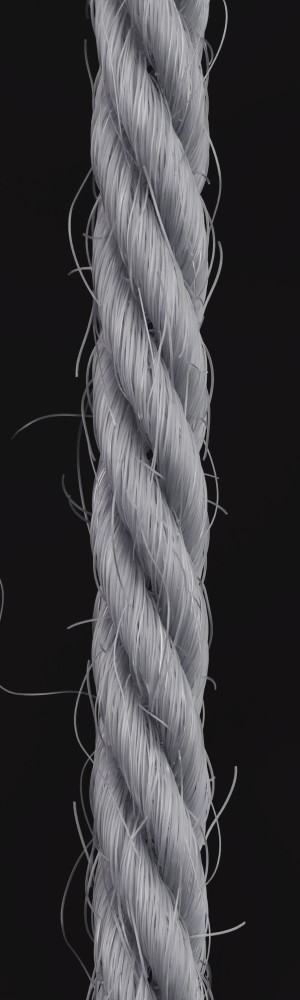} \\[3pt]
		\includegraphics[height=2.8cm] {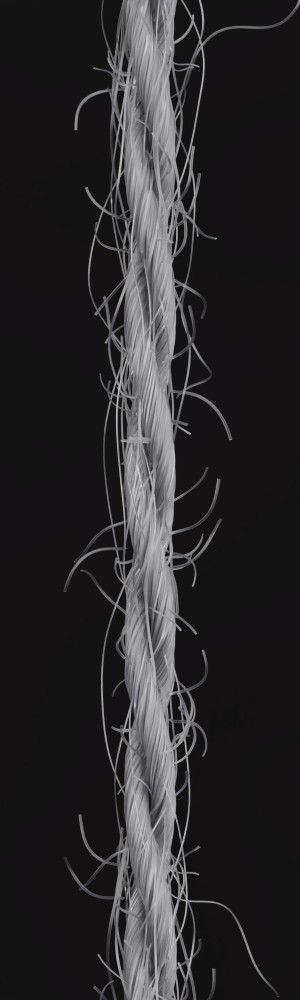} 
				\end{subfigure}
		\hfil
	\begin{subfigure}{0.05\linewidth}
		\caption*{5}
				\includegraphics[height=2.8cm] {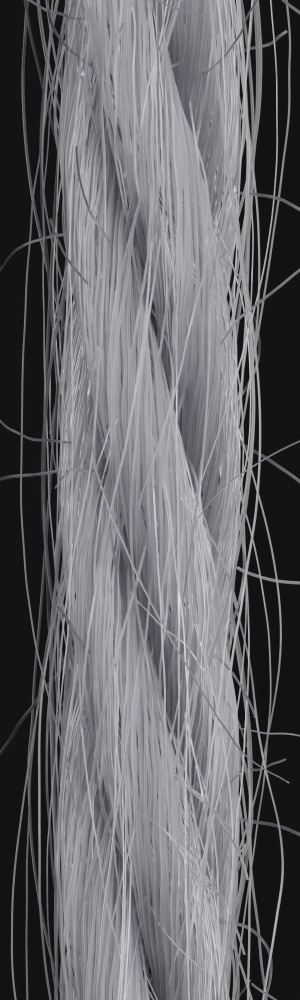}  \\[3pt]
		\includegraphics[height=2.8cm] {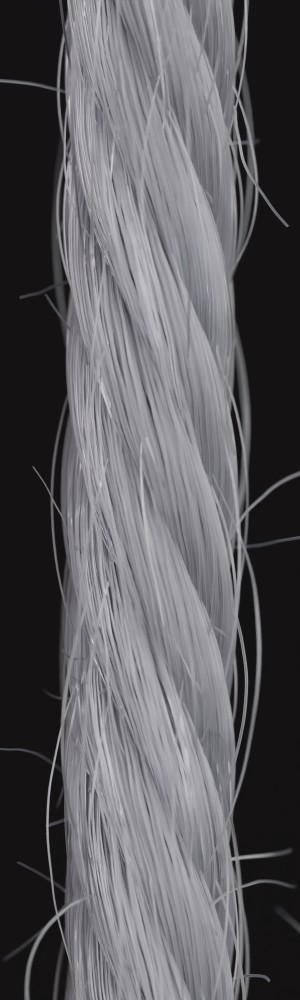}  \\[3pt]
						\includegraphics[height=2.8cm] {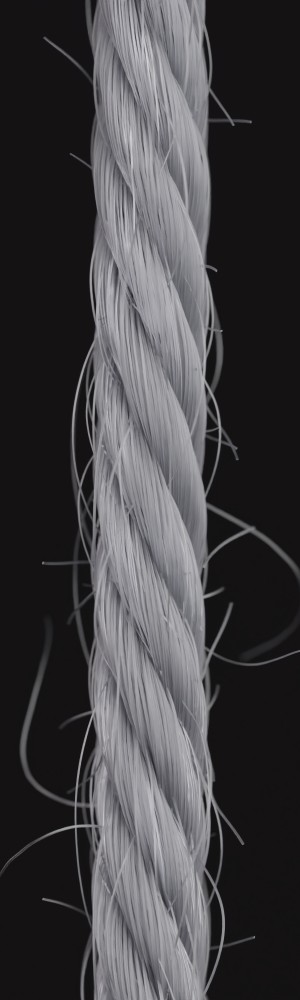}  \\[3pt]
		\includegraphics[height=2.8cm] {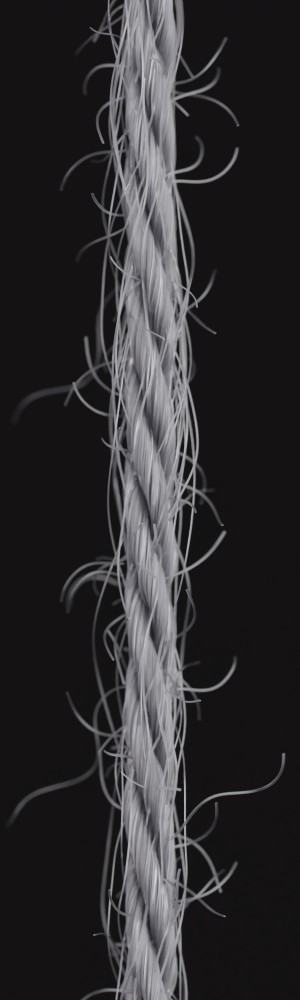} 
				\end{subfigure}
	\hspace{0.025cm}	\hfil
		\begin{subfigure}{0.05\linewidth}
    \caption*{in}
		\includegraphics[height=2.8cm] {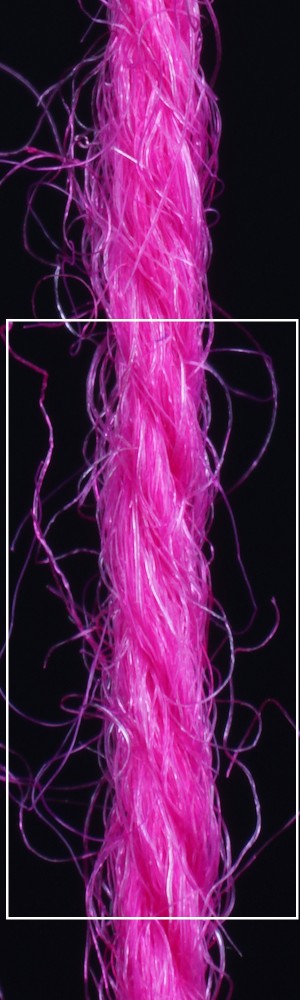} \\[3pt]
		\includegraphics[height=2.8cm] {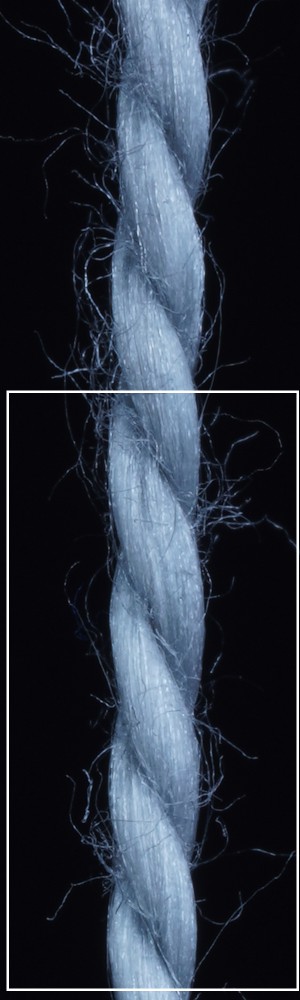}  \\[3pt]
				\includegraphics[height=2.8cm] {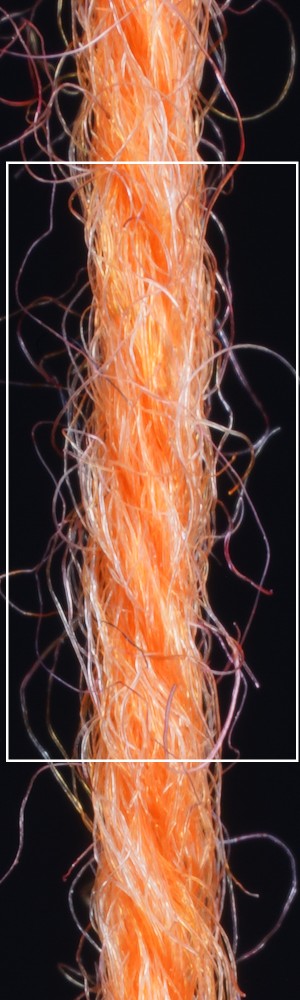} \\[3pt]
		\includegraphics[height=2.8cm] {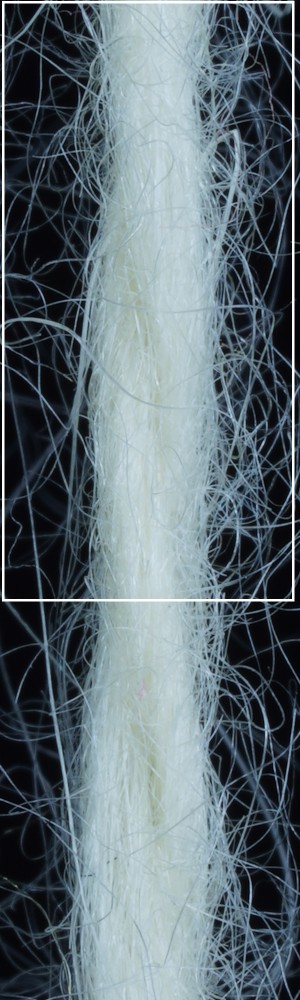} 
		\end{subfigure}
		\hfil
	\begin{subfigure}{0.05\linewidth}
		\caption*{1}
		\includegraphics[height=2.8cm] {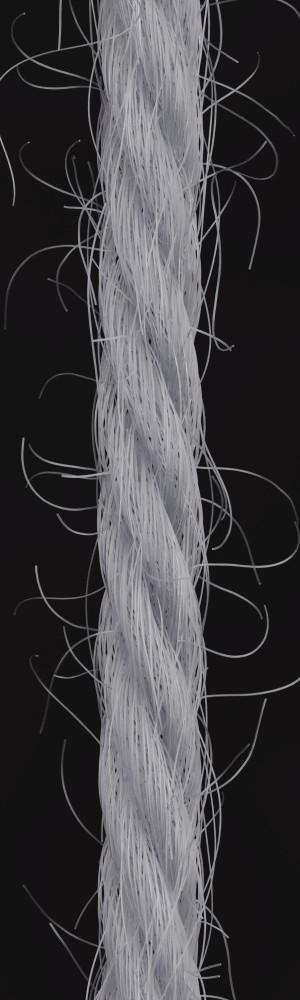}\\[3pt]
		\includegraphics[height=2.8cm] {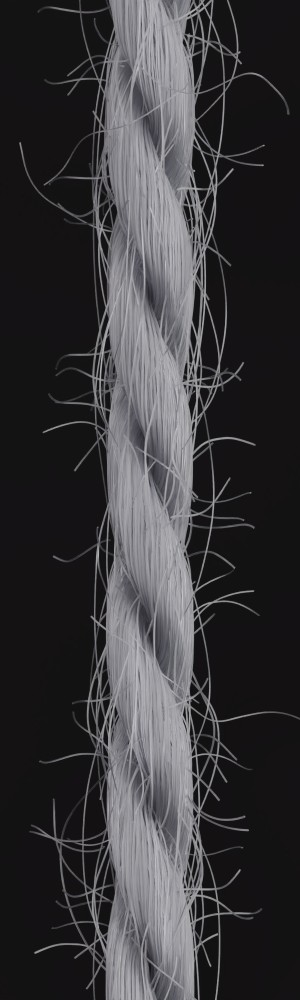}\\[3pt]
				\includegraphics[height=2.8cm] {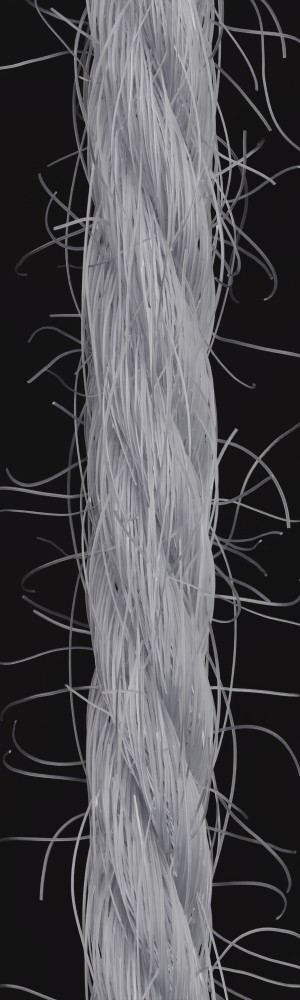}\\[3pt]
		\includegraphics[height=2.8cm] {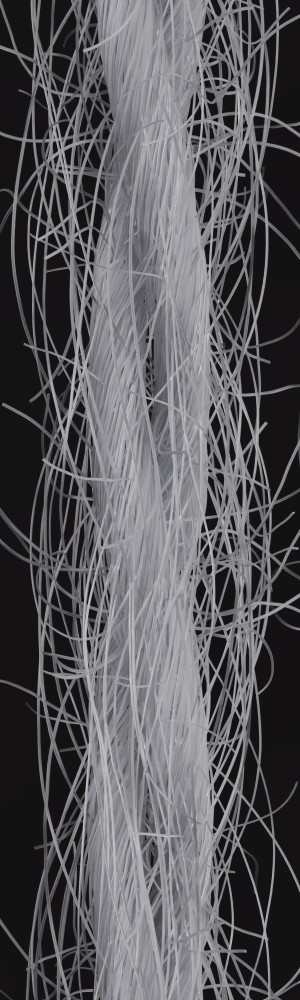}
				\end{subfigure}
		\hfil
	\begin{subfigure}{0.05\linewidth}
		\caption*{2}
				\includegraphics[height=2.8cm] {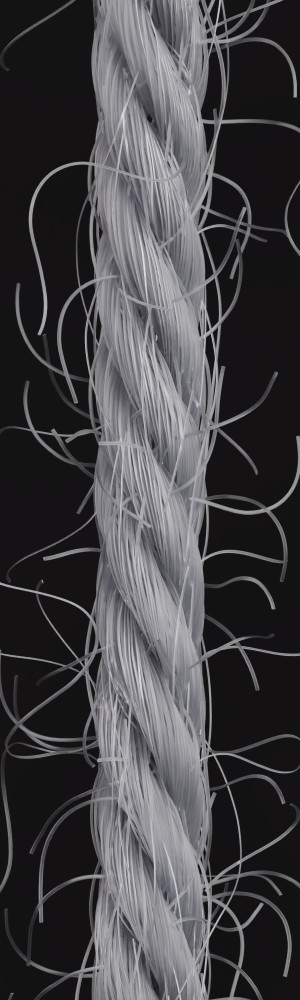} \\[3pt]
		\includegraphics[height=2.8cm] {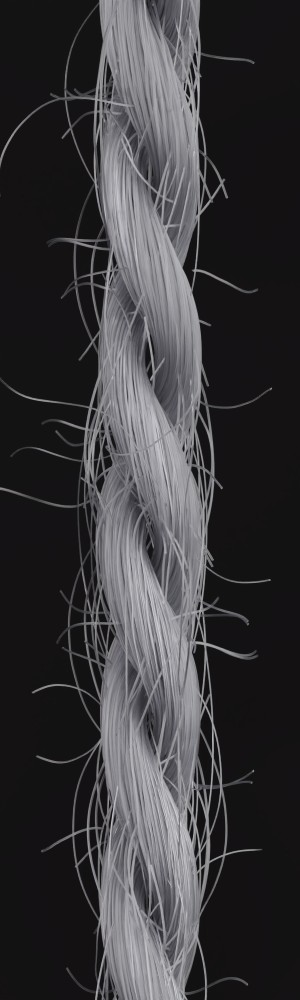}\\[3pt]
						\includegraphics[height=2.8cm] {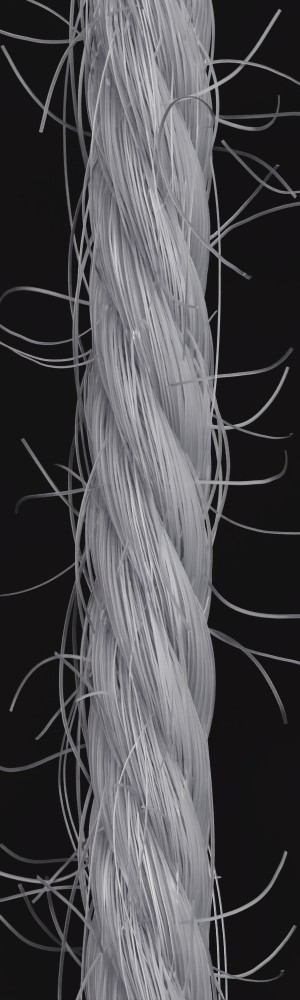} \\[3pt]
		\includegraphics[height=2.8cm] {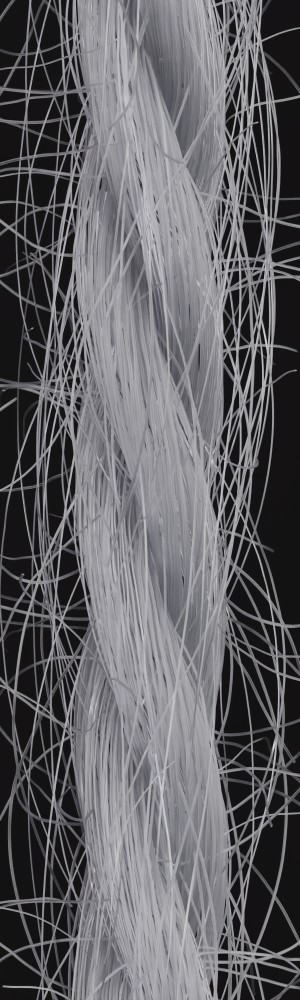}
				\end{subfigure}
		\hfil
	\begin{subfigure}{0.05\linewidth}
		\caption*{3}
		\includegraphics[height=2.8cm] {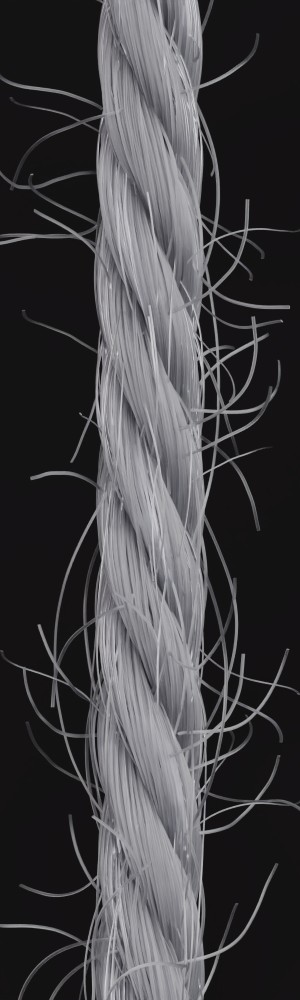}  \\[3pt]
		\includegraphics[height=2.8cm] {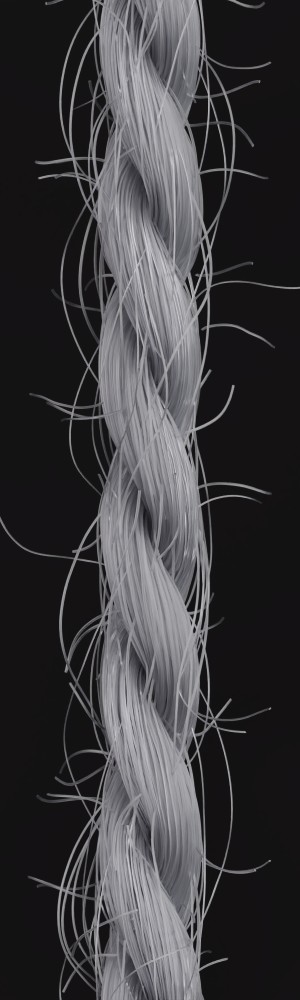}  \\[3pt]
				\includegraphics[height=2.8cm] {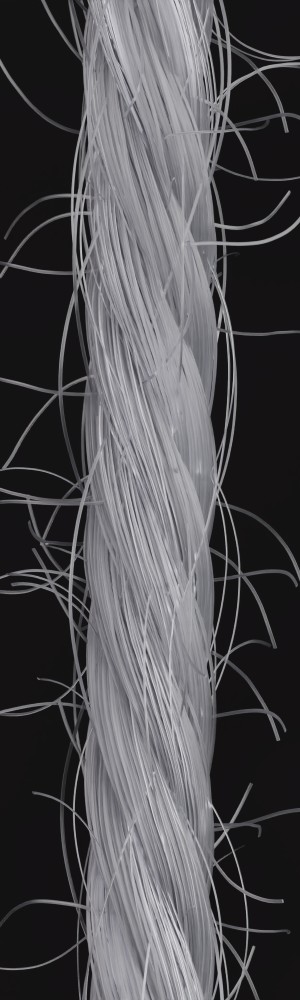}  \\[3pt]
		\includegraphics[height=2.8cm] {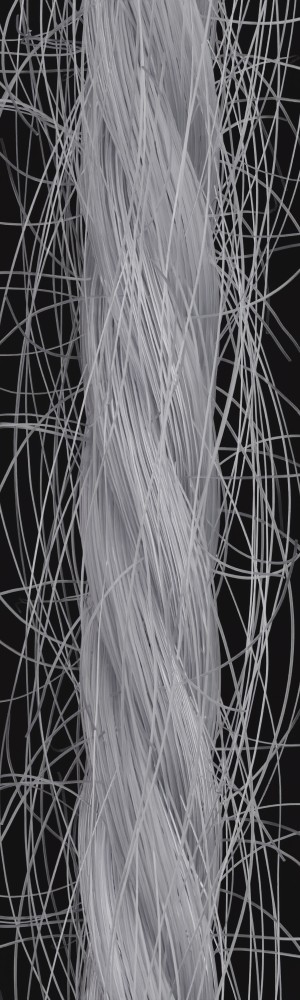} 
				\end{subfigure}
		\hfil
			\begin{subfigure}{0.05\linewidth}
		\caption*{4}
				\includegraphics[height=2.8cm] {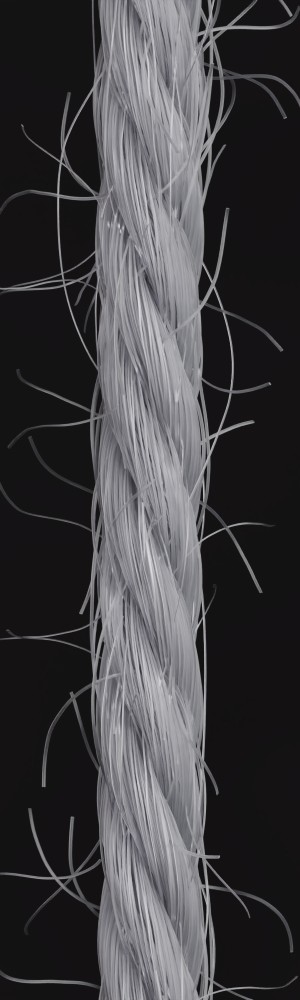} \\[3pt]
		\includegraphics[height=2.8cm] {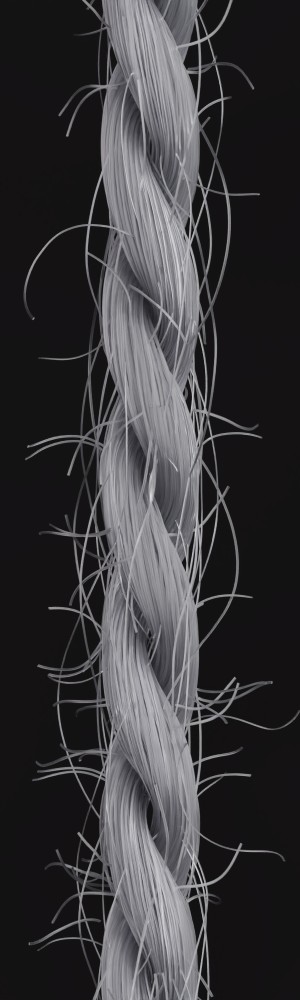} \\[3pt]
						\includegraphics[height=2.8cm] {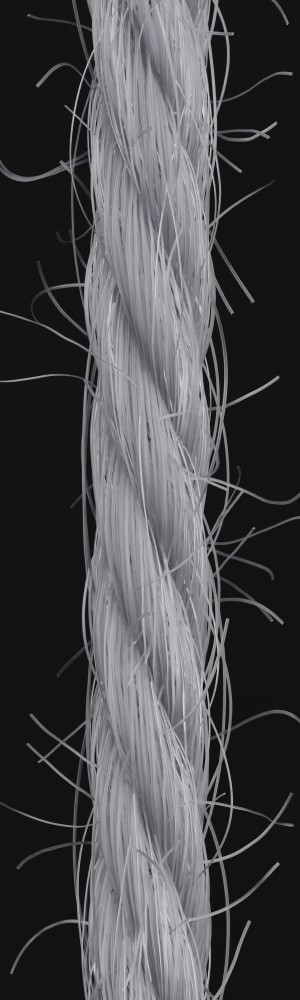} \\[3pt]
		\includegraphics[height=2.8cm] {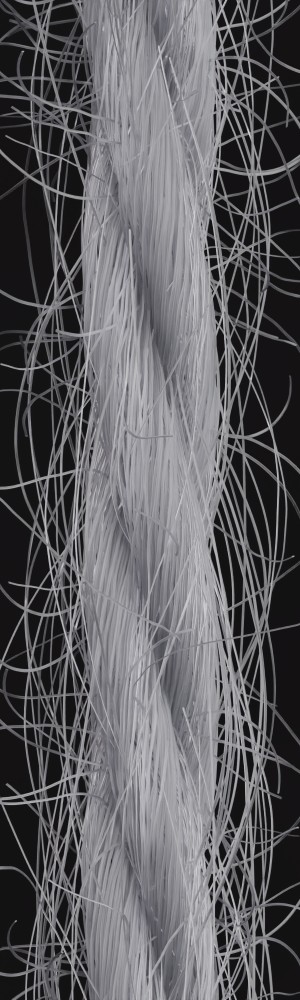} 
				\end{subfigure}
		\hfil
	\begin{subfigure}{0.05\linewidth}
		\caption*{5}
				\includegraphics[height=2.8cm] {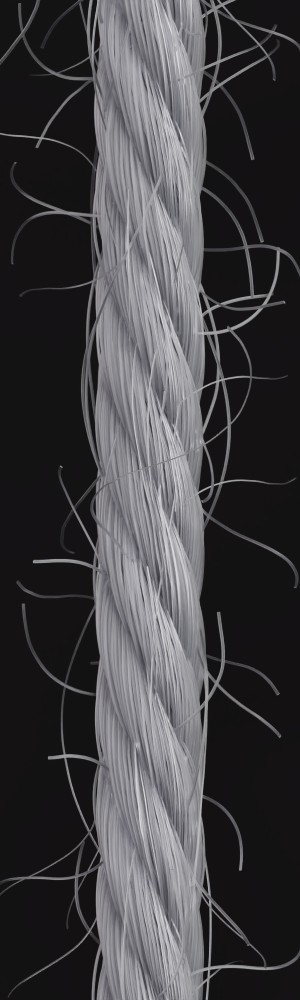}  \\[3pt]
		\includegraphics[height=2.8cm] {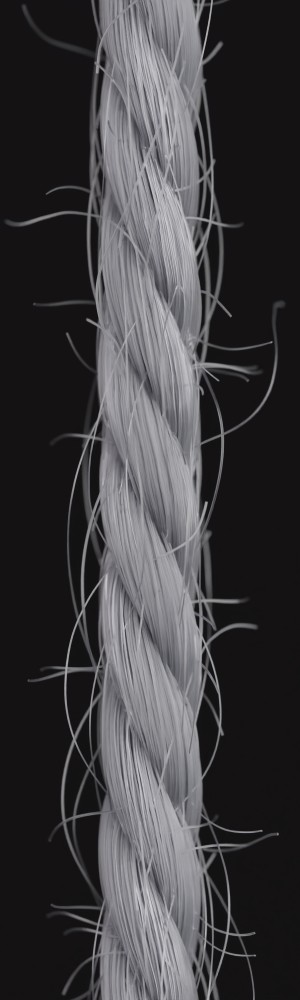} \\[3pt]
						\includegraphics[height=2.8cm] {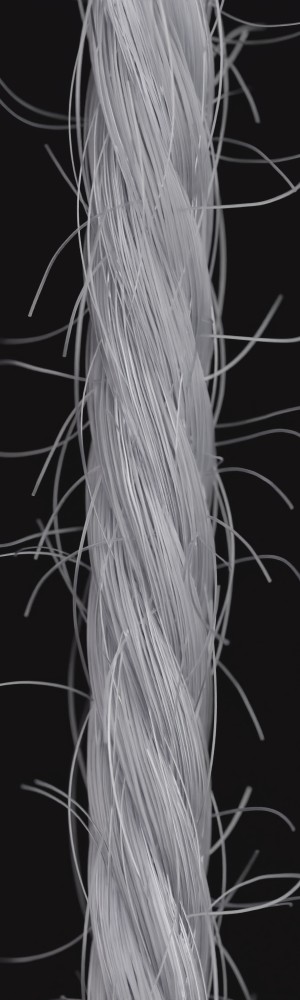}  \\[3pt]
		\includegraphics[height=2.8cm] {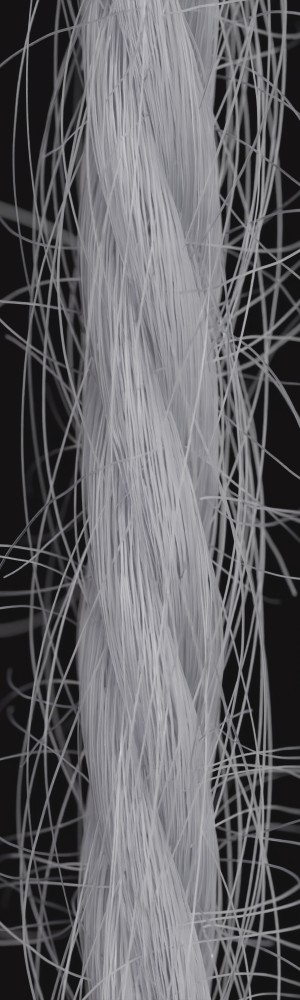} 
				\end{subfigure}
	%\hfil
	 \caption{\label{fig:results_different_yarns} in = input image, 1 = reconstruction image from parameter specific models, trained for each yarn parameter separately, 2 = Reg, 3 = $Reg_{latent}$, 4 = $Reg^{ae}$, 5 = $Reg_{latent}^{ae}$. The rectangle region shows the input image, which was randomly cropped from the whole image.}
\end{figure*}
We observe that yarns with different geometry lead to entirely different appearances of the same pattern.
Furthermore, we can see that if the inferred yarn looks similar to the yarn in the image, the pattern rendered with the inferred yarn will also look similar to the pattern knitted with the real yarn.

\begin{figure*}[h]
	\center
	\includegraphics[width=4cm]{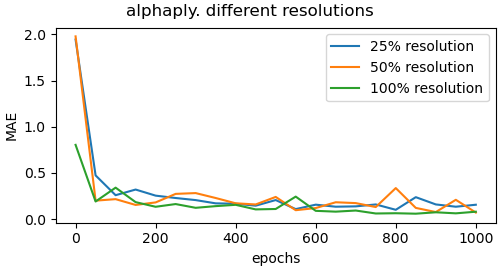}
	\includegraphics[width=4cm]{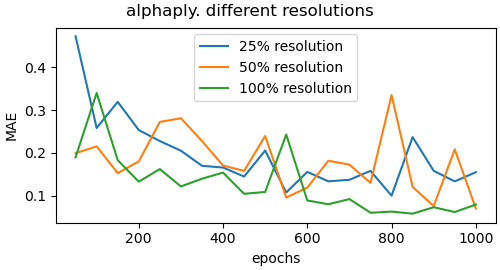}
	\includegraphics[width=4cm]{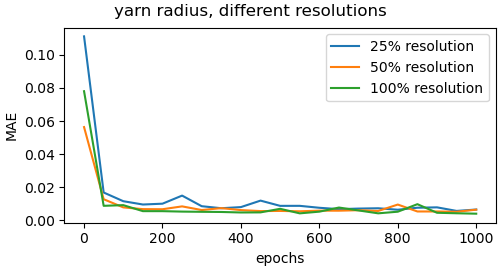}
	\includegraphics[width=4cm]{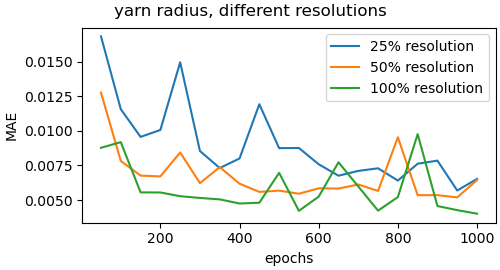}
	\caption{Validation loss comparisons for trainings with different resolution of input images. 1st and 3rd columns: full loss curves, 2nd and 4th columns: loss curves without the first element.}
	\label{fig:resolution}
\end{figure*}

\begin{figure*}[t]
	\center
	\includegraphics[width=4cm]{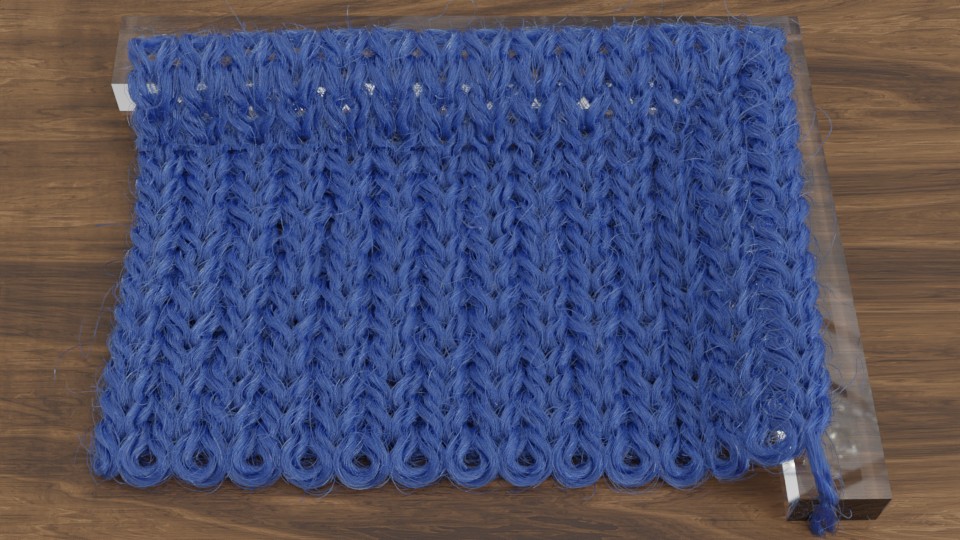}\hspace{0.1pt}
	\includegraphics[width=4cm]{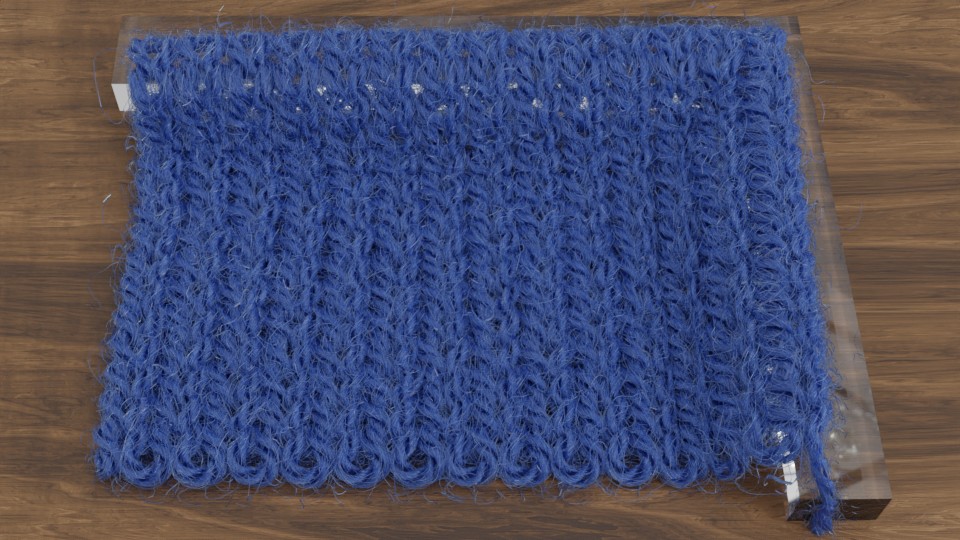}\hspace{0.1pt}
	\includegraphics[width=4cm]{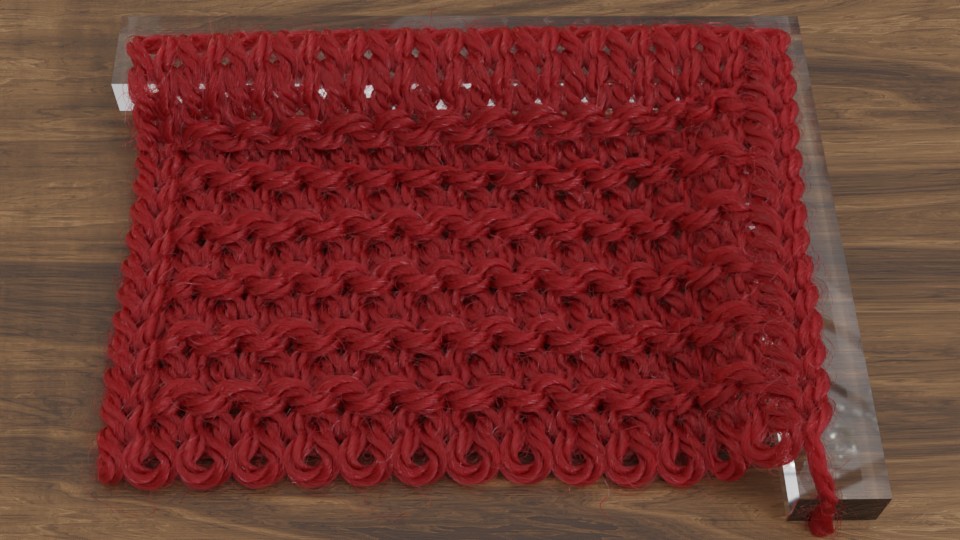}\hspace{0.1pt}
	\includegraphics[width=4cm]{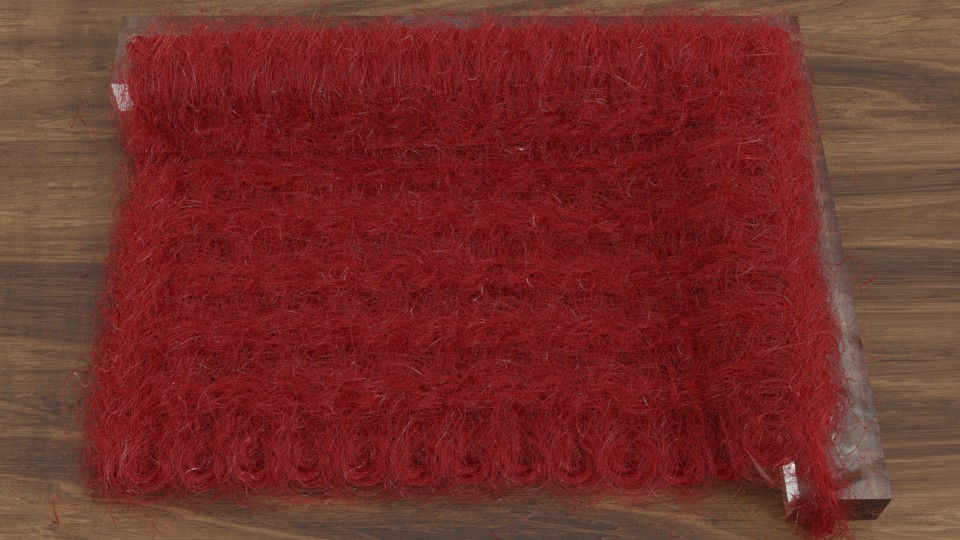}\\
	\includegraphics[height=3.1cm, angle =90]{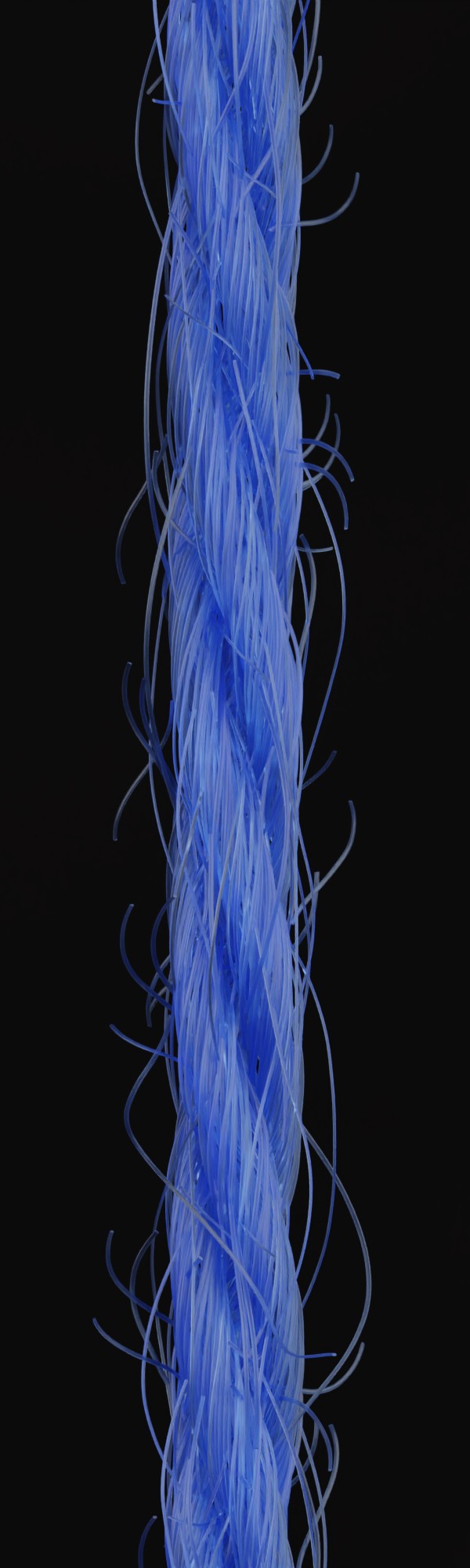} \hspace{1cm}
	\includegraphics[height=3.1cm, angle =90]{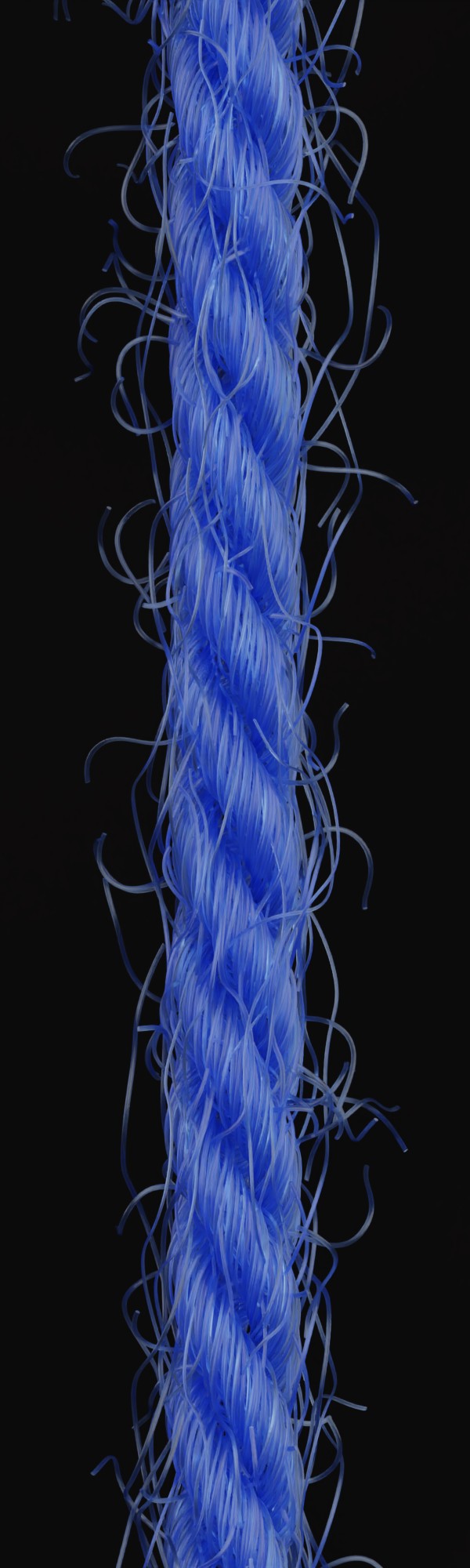}\hspace{1cm}
	\includegraphics[height=3.1cm, angle =90]{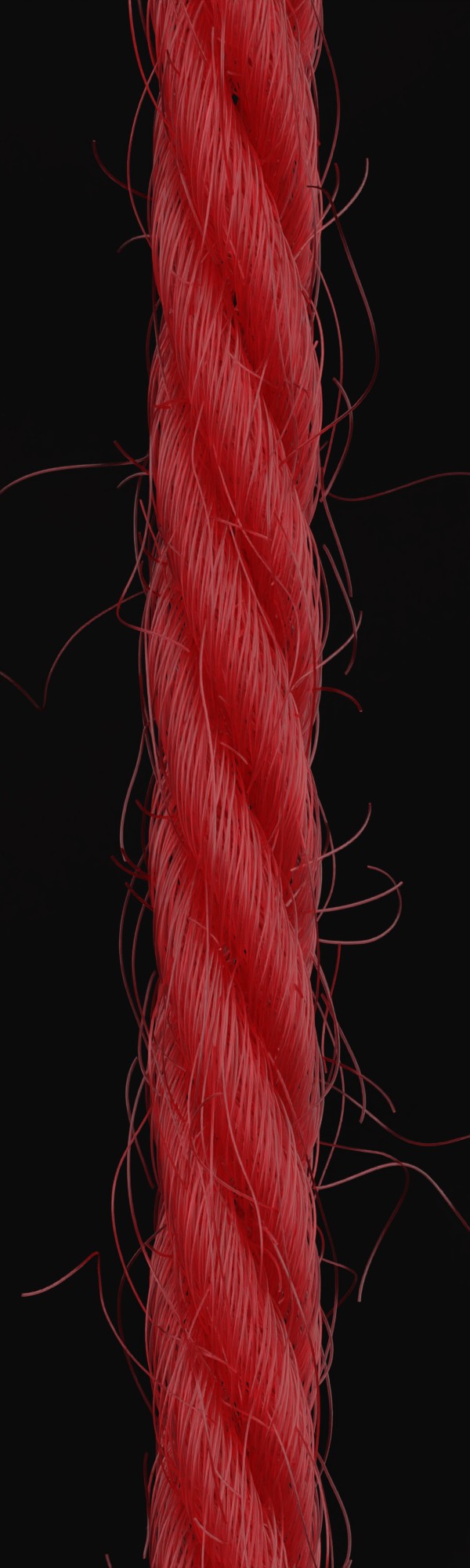}\hspace{1cm}
	\includegraphics[height=3.1cm, angle =90]{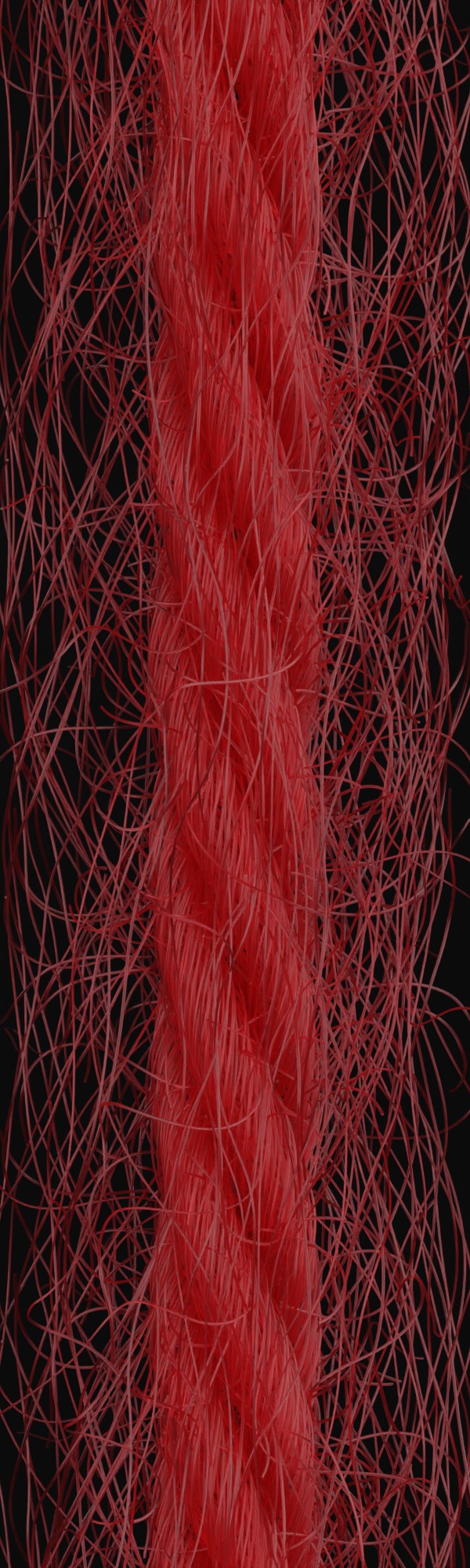}
	\caption{Two examples of editing operations for yarns with original inferred parameters and the edited ones together with corresponding renderings of knitted patches. Reflectance parameters were not part of the inference but chosen arbitrarily for demonstration. 1st column: golden yarn from Figure~\ref{fig:results_different_yarns} in the 3rd row, left. 2nd column: the same yarn but with both pitch parameters $\alpha$ and $\alpha_{ply}$ divided by 2. 3rd column: yellow yarn from Figure~\ref{fig:results_different_yarns} in the 3rd row. 4th column: the same yarn but with parameters for flyaway amount and length multiplied by 2.}
	\label{fig:editing}
\end{figure*}

\begin{figure}
	\center
	\begin{subfigure}{0.1\linewidth}
    \caption*{in}
		\includegraphics[height=3.1cm] {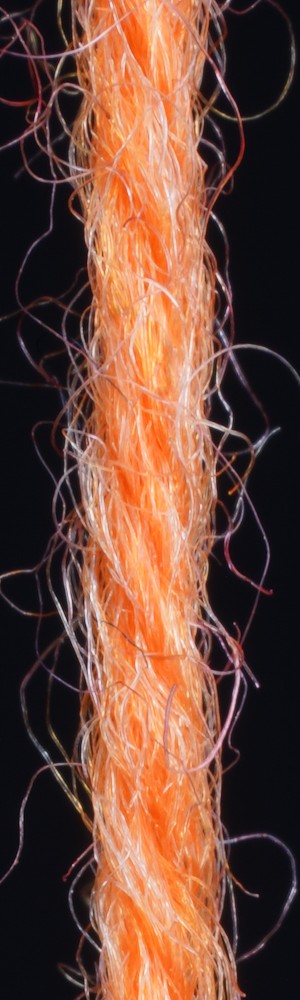} \\[3pt]
		\includegraphics[height=3.1cm] {figures/yarns/2er_oliv_input_small.jpg} 
		\end{subfigure}
		\hfil
	\begin{subfigure}{0.1\linewidth}
		\caption*{1}
		\includegraphics[height=3.1cm] {figures/yarns/4er_orange_diffnets_small.jpg}\\[3pt]
		\includegraphics[height=3.1cm] {figures/yarns/2er_oliv_diffnets_small.jpg} 
				\end{subfigure}
		\hfil
	\begin{subfigure}{0.1\linewidth}
		\caption*{2}
				\includegraphics[height=3.1cm] {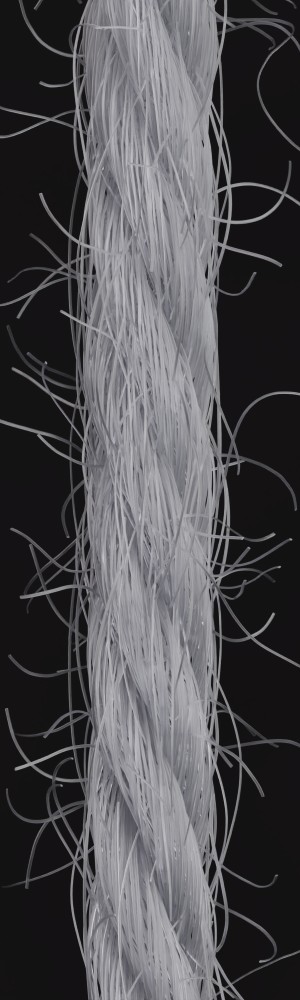} \\[3pt]
		\includegraphics[height=3.1cm] {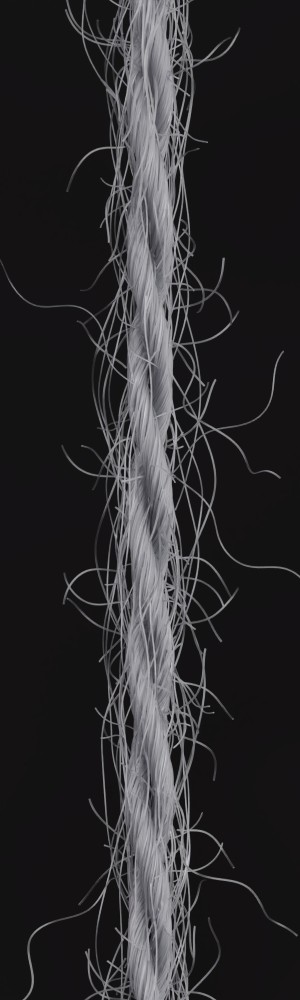} 

				\end{subfigure}
		\hfil
	\begin{subfigure}{0.1\linewidth}
		\caption*{3}
				\includegraphics[height=3.1cm] {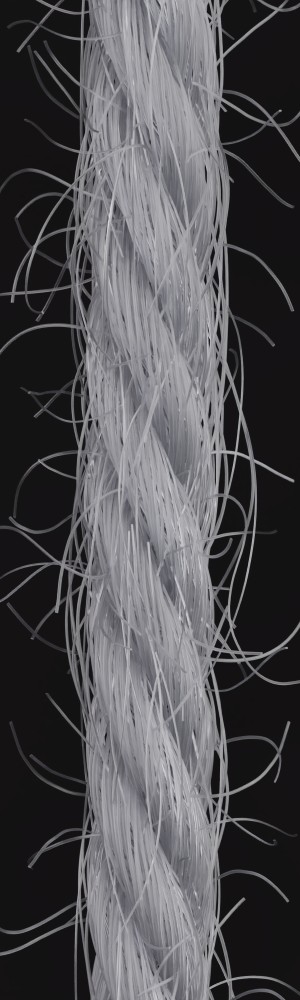} \\[3pt]
		\includegraphics[height=3.1cm] {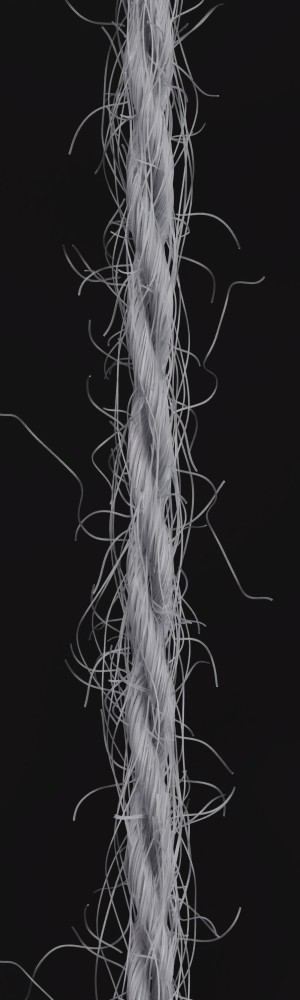}
				\end{subfigure}
				\hspace{0.03cm}				\hfil
		\begin{subfigure}{0.1\linewidth}
    \caption*{in}
		\includegraphics[height=3.1cm] {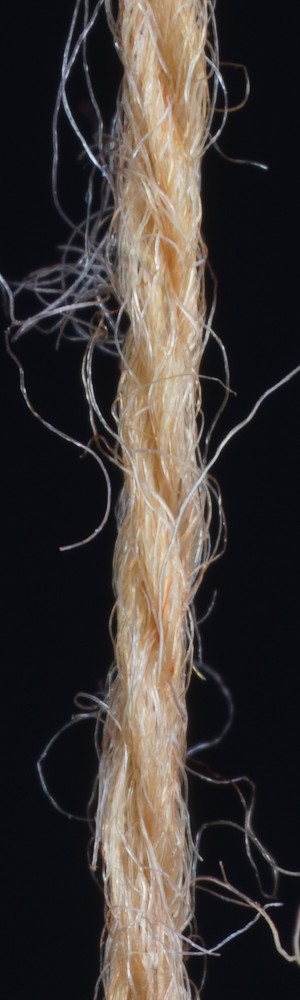} \\[3pt]
		\includegraphics[height=3.1cm] {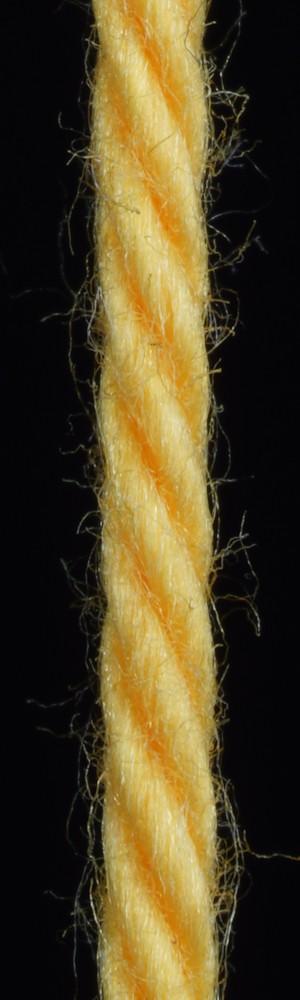} 
		\end{subfigure}
		\hfil
	\begin{subfigure}{0.1\linewidth}
		\caption*{1}
				\includegraphics[height=3.1cm] {figures/yarns/3er_gelb_diffnets_small.jpg} \\[3pt]
		\includegraphics[height=3.1cm] {figures/yarns/4er_gelb_diffnets_small.jpg} 
				\end{subfigure}
		\hfil
	\begin{subfigure}{0.1\linewidth}
		\caption*{2}
		\includegraphics[height=3.1cm] {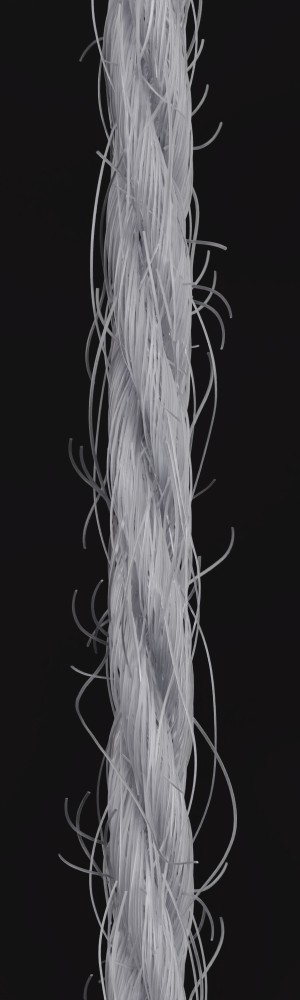} \\[3pt]
		\includegraphics[height=3.1cm] {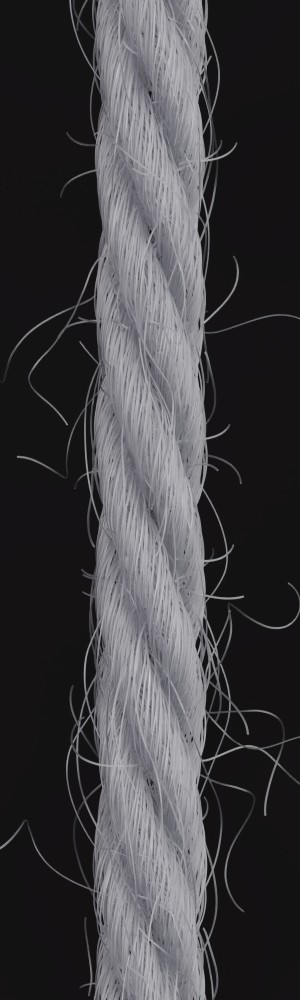} 
				\end{subfigure}
		\hfil
	\begin{subfigure}{0.1\linewidth}
		\caption*{3}
		\includegraphics[height=3.1cm] {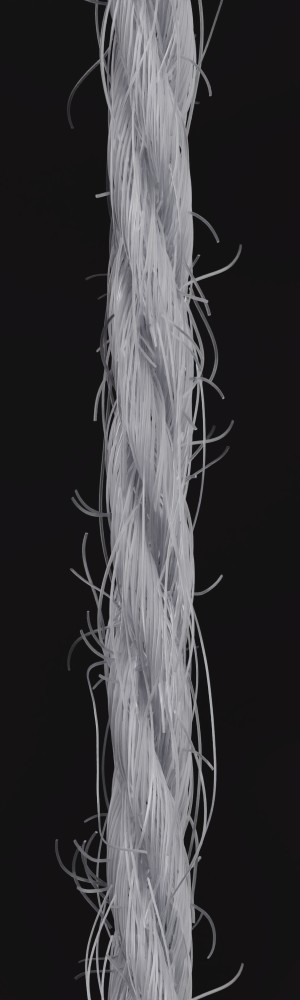} \\[3pt]
		\includegraphics[height=3.1cm] {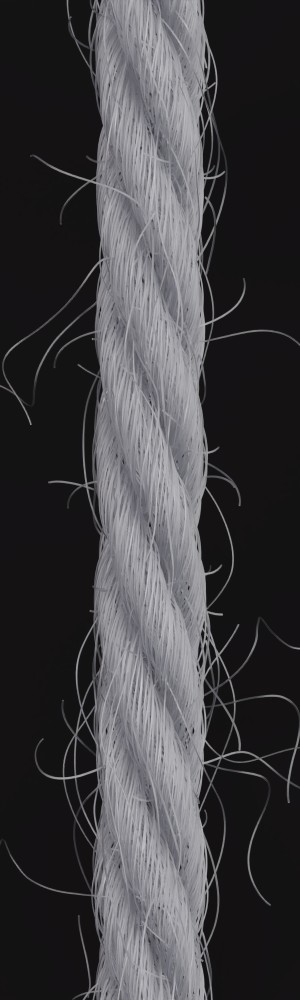} 
				\end{subfigure}
	 \caption{\label{fig:results_different_relolutions} in = input image, 1 = reconstruction image from different models trained for each yarn parameter separately, where the $\alpha_{ply}$ parameter was trained on images with full resolution, 2 = $\alpha_{ply}$ parameter was trained on images with $50\%$ of the full resolution, 3 = $\alpha_{ply}$ parameter was trained on images with $25\%$ of the full resolution.}
\end{figure}

Once the parameters are inferred, we can use them also for editing and for the creation of new yarns. Some examples for modification of the twist parameters $\alpha$ and $\alpha_{ply}$ as well as some flyaways parameters are depicted in Figure \ref{fig:editing}.
\paragraph{Ablation study regarding effect of resolution}\label{ssec:study}
To get insights on the effect of the resolution of the input images, we trained the networks for the different yarn parameters on images of significantly lower resolution. We experimented with the reduction to 50 and 25 percent of the original resolution of 1200 times 584 pixels. Figure~\ref{fig:resolution} shows the validation loss plots for the alphaply and yarn radius parameters for different resolutions. Figure~\ref{fig:results_different_relolutions} shows some visual comparisons of reconstructed yarns with the corresponding parameters for alphaply. As can be observed, the achieved accuracy decreases with decreasing image resolution. We expect this to be a result of the lower quality of the depiction of the individual fiber arrangements that can be seen in terms of a blurring of the yarn structure. 
In order to demonstrate the robustness of our approach to different exposure times we made exposure series of the input yarns and tested images with different exposure times. The results show that as long as the images are not too dark or over-exposed, the inferred parameters vary only insignificantly and the reconstructions are very similar. %\mwrevision{}{belongs to ablation study}
\subsection{Limitations}
In addition to the dependence on the quality of the depicted fiber arrangements (as shown in the previous section), our approach depends on having the variations to be expected in the test data included in the training data.
Note that our dataset includes only yarns with a normal (helix-like) fiber twisting.
However, other fiber twisting-types could also occur as shown with the example in Figure~\ref{fig:discussion}.
The depiction shows a reconstruction that exhibits a high similarity to the input yarn.
The thickness on both ply- and yarn-level as well as the number of twists closely follow the original structure.%
Since we did not consider this type of ply-twist in our yarn generator, there is also some deviation.
We expect that such deviations might be handled by further extending the dataset regarding further types of yarn variations.

Furthermore, despite the fact that our yarn generator also supports the fourth hierarchical level (i.e., where thinner yarns are twisted into thicker yarns, see Figure~\ref{fig:discussion} d)-e)), we only included yarns represented based on the first three levels, which limits our approach to the prediction of the characteristics up to the third level. However, the extension to the level of also twisting yarns is straightforward and we leave it for future work.

%% file: conclusions.tex
\section{Conclusions}
We presented an investigation of different neural inverse procedural modeling methods with different architectures and loss formulations to infer procedural yarn parameters from a single yarn image.
The key to our approach was the accurate hierarchical parametric modeling of yarns, enhanced by handling elliptic fiber cross-sections, as occurring in many types of natural hair fibers, as well as more accurately handling flyaway characteristics and the twisting axis and the respective generation of synthetic yarns that are realistic enough so that the trained model can extrapolate to the real yarn inputs.
Our experiments indicate that that the complexity of yarn structures in terms of twisting and migration characteristics of the involved fibers can be better encountered in terms of ensembles of networks that focus on individual characteristics than in terms of a single neural network that estimates all parameters.
In addition, we demonstrated that carefully designed parametric, procedural yarn models in combination with respective neural architectures and respective loss functions even allow robust parameter inference based on models trained on purely synthetic data.
\begin{figure}
	\center
	\includegraphics[height=3cm]{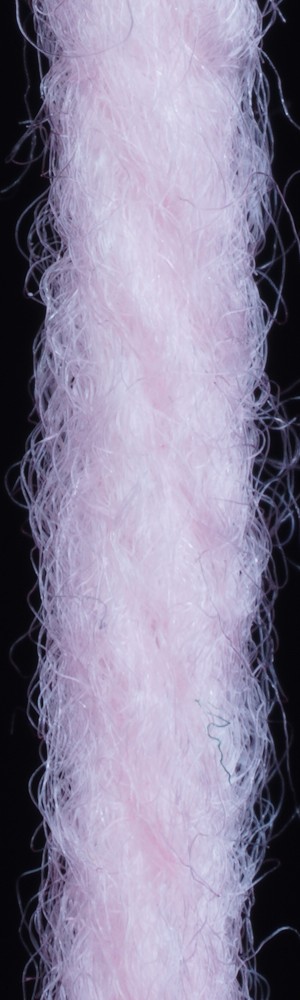}
	\includegraphics[height=3cm]{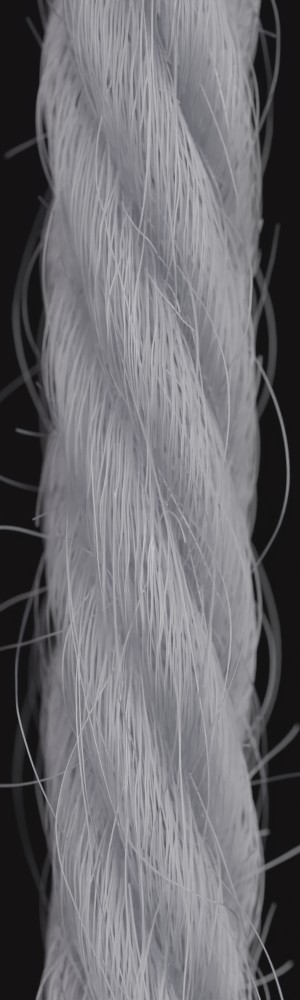}
	\includegraphics[height=3cm]{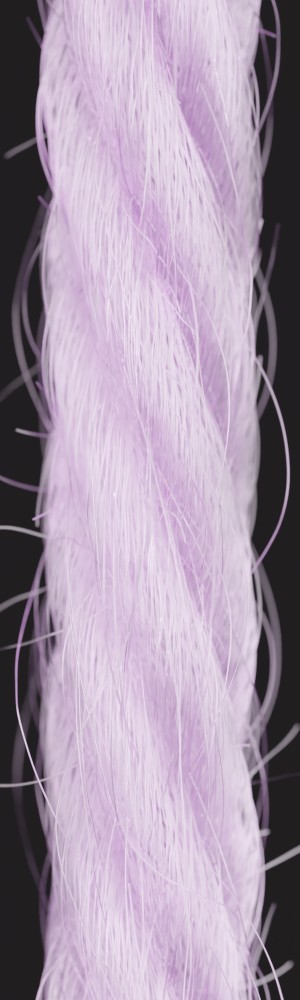} \hspace{1cm}
	\includegraphics[height=3cm]{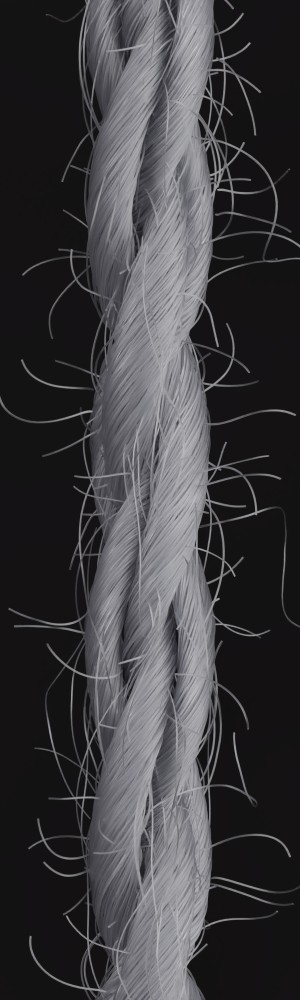}
	\includegraphics[height=3cm]{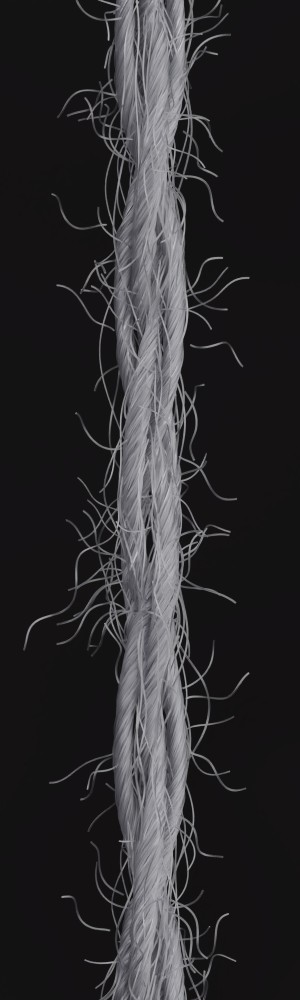}
	
	\hspace{0.3cm} a) \hspace{0.6cm} b) \hspace{0.6cm} c) \hspace{1.5cm} d)  \hspace{0.6cm} e)
	
	\caption{a) Input image of a yarn made by unusual (non-helical) fiber twisting procedure. b) Rendering of a yarn with infered parameters with default material. c) Rendering with color. d) and e) Examples of yarns of fourth level, where two thinner yarns are twisted into one to make it thicker and better suitable for knitting: d) Two yarns of the thin grey yarn from Figure~\ref{fig:results_different_yarns}, fourth row, e) Two yarns of the thick grey yarn from second row of Figure~\ref{fig:results_different_yarns}}
	\label{fig:discussion}
\end{figure}
In the scope of this paper we focused solely on the geometric fiber arrangement including migration characteristics (i.e. flyaways) and left the prediction of the reflectance characteristics of knitting yarns for future work.
Further developments may also consider a further hierarchical level of yarns, i.e. thinner yarns twisted to thicker yarns. Whereas we did not focus on inferring parameters for this kind of yarns, our yarn generator would be able to produce the respective characteristics and might allow enriching the dataset accordingly in future work.

%% file: supplemental.tex
\twocolumn[  
    \begin{@twocolumnfalse}
        \begin{center}

             \textbf{\LARGE{Neural inverse procedural modeling of knitting yarns from images (supplemental material).}}

         \end{center}
     \end{@twocolumnfalse}
]
		
\linenumbers

\section{Yarn sampler}
In the course of our experiments, we heuristically determined parameter sampling intervals that produced natural-looking yarns. These intervals are presented below. For some of the parameters, we defined sampling intervals directly, while for others, the sampling was implemented through auxiliary variables. 

The intervals for fiber thickness were chosen as follows:
\begin{linenomath}
\begin{align}
&t_y = [0.006, 0.01] \\
&t_x = [t_y, 2.5 \cdot t_{y}]
\end{align}
\end{linenomath}
For the number $n$ of plies, we sampled integers between 2 and 6, while for the number $m$ of fibers, we sampled integers between 40 and 200. All other raw yarn parameters were sampled indirectly using auxiliary variables.

Let $r_{frac} = \frac{r_y}{r_x}$ be the parameter that reflects how much a ply has been squeezed during the twisting and how much it deviates from its original circular cross-section (Fig. 1 in the paper). The fewer plies there are in the yarn, the less they resemble a circle after twisting:

\begin{linenomath}
\begin{align}
&n = 2 \Rightarrow r_{frac} = [0.67, 0.9] \\
&n = 3 \Rightarrow r_{frac} = [0.72, 0.91] \\
&n > 3 \Rightarrow r_{frac} = [0.85, 0.95]
\end{align}
\end{linenomath}
Let the parameter $area_{frac}^{ply}$ represent different fiber densities in a ply:
\begin{linenomath}
\begin{equation}
area_{frac}^{ply} = \frac{m\cdot t_x \cdot t_y \cdot \pi}{r_x \cdot r_y \cdot \pi} = \frac{m\cdot t_x \cdot t_y }{r_x^2 \cdot r_{frac} }
\end{equation}
\end{linenomath}

We sample it from the following interval:
\begin{linenomath}
\begin{equation}
area_{frac}^{ply} = [0.035, 0.215]
\end{equation}
\end{linenomath}

Then the parameters $r_x$ and $r_y$ can be computed as
\begin{linenomath}
\begin{align}
&r_x = \sqrt{\frac{m \cdot t_{x}\cdot t_{y}}{area_{frac}^{ply} \cdot r_{frac}}}  \\
&r_y = r_{frac} \cdot r_x
\end{align}
\end{linenomath}

The auxiliary variable $area_{frac}^{yarn}$ depicts the ply density in the yarn:
\begin{linenomath}
\begin{equation}
area_{frac}^{yarn} = \frac{m\cdot t_x \cdot t_y \cdot \pi}{r_x \cdot r_y \cdot \pi} = \frac{m\cdot t_x \cdot t_y }{r_x^2 \cdot r_{frac} }
\end{equation}
\end{linenomath}

We sample from the following interval:
\begin{linenomath}
\begin{equation}
area_{frac}^{yarn} = [0.55, 0.82]
\end{equation}
\end{linenomath}

Furthermore, we sample both the auxiliary variable $\gamma$ of the helix angle of the fiber twist in a ply and the auxiliary variable $\gamma_{ply}$ of the helix angle of the ply twist in the yarn as follows (both angles are represented in radians):
\begin{linenomath}
\begin{align}
&\gamma_{ply} = [50, 80] \cdot \frac{\pi}{180}\\
&\gamma = [50, 80] \cdot \frac{\pi}{180}
\end{align}
\end{linenomath}

Then, following the helix formula, we compute the parameters for the pitch $\alpha$ of the ply helix and the pitch $\alpha_{ply}$ and radius $R_{ply}$ of the yarn helix:
\begin{linenomath}
\begin{align}
&R_{ply} = \sqrt{\frac{n \cdot r_{frac} (\frac{r_x}{sin(\gamma_{ply})})^{2}}{area_{frac}^{yarn}}} - r_{frac}\frac{r_x}{sin(\gamma_{ply})} \\
&\alpha_{ply} = 2 \pi R_{ply} \cdot tan(\gamma_{ply})\\
&\alpha = -1 \cdot 2 \pi r_{x} \cdot tan(\gamma)
\end{align}
\end{linenomath}

Note the opposite signs for the clockwise and counterclockwise directions of the twist of the ply and yarn. This is not true for all existing yarns, but it is true for all knitting yarns that we have observed. If necessary, yarns with other combinations of clockwise and counterclockwise twist can be added to the database.

The overall intervals of all learnable yarn parameters are shown in Table~\ref{tab_params}.

\begin{table}
\centering
\scriptsize
\caption{Value intervals of the learnable parameters in our synthetic yarn database.}
\setlength{\tabcolsep}{7pt}
\begin{tabular}{|p{30pt}|p{50pt}|}
\hline
Parameter& Value interval\\
\hline
$m$ & [20,200] \\
$t_x$ & [0.006,0.01] \\
$t_y$ & [0.006,0.02] \\
$\alpha$ & [-25.778, -0.476] \\
\hline
$n$ & [2,6] \\
$r_x$& [0.029,0.789] \\
$r_y$ & [0.042,0.830] \\
$\alpha_{ply}$ &[0.639,31.655] \\
$R_{ply}$ &[0.053,1.486] \\
\hline
$j$ & [0,0.3] \\
$j_{xy}$ & [0,0.03]\\
$g$ & [30,300]\\
$p$ & [0.35,0.65] \\
$\beta$& [0.050,1.571]\\
$l_{hair}$& [0.222,14.5]\\
$s$& [0,1]\\
$l_{loop}$& [0.407,34.627]\\
$d_{mean}$& [0.394,30.469]\\
$d_{std}$& [0.007,5]\\
\hline
\end{tabular}
\label{tab_params}
\end{table}

\section{Inferred yarn parameters}
In Chapter 5 of our paper, we presented exemplary results of our experiments on yarn parameter inference, including a comparison between all investigated approaches (Figure 12).
The corresponding inferred parameters are presented in the Tables~\ref{tab_params_results} (raw yarn parameters) and~\ref{tab_params_fly} (flyaway parameters).
\begin{table*}
\begin{center}
\scriptsize
\caption{Inferred raw yarn parameters for the yarns of the Figure 12 from top to bottom and from left to right. For each yarn there are 5 rows, each corresponding to parameter detection from different experiments: from top to bottom: parameter specific models, $Reg$, $Reg^{ae}$, $Reg_{latent}$, $Reg_{latent}^{ae}$}
%\label{table}
\setlength{\tabcolsep}{5pt}
\begin{tabular}{|p{22pt}|p{3pt}|p{20pt}|p{15pt}|p{20pt}|p{20pt}|p{20pt}|p{10pt}|p{30pt}|}
\hline
yarn&$n$&$\alpha_{ply}$ & $R_{ply}$ & $\alpha$ & $r_x$ & $r_y$ & $m$ & thickness \\
\hline
\multirow{5}{*}{rose} &4 & 4.515 & 0.555 & -3.611& 0.329 &0.283 &72& 2\\
&4 & 4.591 & 0.614 & -3.217& 0.305 &0.285 &102& 0.021,0.008\\
&5 & 5.186 & 0.603 & -3.989& 0.327 &0.286 &107& 0.020,0.008\\
&4 & 5.053 & 0.546 & -3.545& 0.332 &0.295 &100& 0.018,0.008\\
&5 & 5.110 & 0.581 & -3.474& 0.304 &0.274 &118& 0.018,0.007\\
\hline
\multirow{5}{*}{red} &4 & 7.815 & 0.449 & -2.442& 0.297 &0.240 &131& 1\\
&5 & 6.636 & 0.518 & -2.647& 0.250 &0.225 &106& 0.018,0.008\\
&4 & 7.760 & 0.520 & -3.273& 0.282 &0.249 &132& 0.017,0.008\\
&4 & 7.911 & 0.477 & -3.033& 0.282 &0.246 &139& 0.018,0.008\\
&5 & 7.663 & 0.504 & -3.786& 0.280 &0.238 &135& 0.016,0.008\\
\hline
\multirow{5}{*}{golden} &3 & 6.474 & 0.289 & -2.392& 0.241 &0.195 &43& 2\\
&3 & 5.823 & 0.291 & -2.326& 0.236 &0.200 &62& 0.022,0.008\\
&4 & 5.708 & 0.292 & -3.175& 0.228 &0.192 &87& 0.020,0.008\\
&3 & 6.714 & 0.261 & -3.223& 0.238 &0.206 &81& 0.019,0.008\\
&4 & 6.021 & 0.280 & -3.150& 0.227 &0.194 &72& 0.017,0.007\\
\hline
\multirow{5}{*}{pink 6ply} &6 & 10.679 & 0.672 & -3.288& 0.390 &0.350 &64& 2\\
&5 & 9.884 & 0.758 & -4.457& 0.358 &0.373 &86& 0.021,0.008\\
&5 & 9.079 & 0.669 & -5.512& 0.399 &0.363 &92& 0.021,0.008\\
&5 & 11.173 & 0.636 & -3.868& 0.371 &0.328 &106& 0.018,0.008\\
&6 & 11.000 & 0.688 & -6.077& 0.377 &0.332 &113& 0.017,0.007\\
\hline
\multirow{5}{*}{mixed} &3 & 14.961 & 0.639 & -6.152& 0.521 &0.423 &99& 2\\
&3 & 13.159 & 0.598 & -5.445& 0.472 &0.415 &112& 0.022,0.008\\
&3 & 11.965 & 0.572 & -6.160& 0.478 &0.409 &129& 0.020,0.008\\
&4 & 14.522 & 0.540 & -4.994& 0.441 &0.386 &135& 0.019,0.008\\
&4 & 15.316 & 0.635 & -6.440& 0.433 &0.391 &124& 0.018,0.007\\
\hline
\multirow{5}{*}{blue} &4 & 11.009 & 0.602 & -2.866& 0.406 &0.298 &74& 1\\
&4 & 10.108 & 0.663 & -6.308& 0.458 &0.407 &151& 0.017,0.008\\
&5 & 7.545 & 0.631 & -7.674& 0.395 &0.359 &120& 0.017,0.008\\
&4 & 11.868 & 0.606 & -4.054& 0.383 &0.335 &114& 0.019,0.008\\
&6 & 9.505 & 0.638 & -6.250& 0.366 &0.325 &123& 0.017,0.007\\
\hline
\multirow{5}{*}{yellow} &4 & 6.398 & 0.393 & -2.023& 0.278 &0.236 &65& 1\\
&4 & 6.176 & 0.425 & -2.838& 0.272 &0.237 &92& 0.015,0.007\\
&4 & 6.087 & 0.435 & -3.384& 0.254 &0.222 &110& 0.014,0.007\\
&4 & 5.897 & 0.424 & -1.953& 0.261 &0.233 &103& 0.014,0.007\\
&5 & 5.996 & 0.415 & -3.054& 0.247 &0.217 &105& 0.014,0.007\\
\hline
\multirow{5}{*}{grey thin} &2 & 3.348 & 0.139 & -1.175& 0.146 &0.115 &82& 1\\
&2 & 3.230 & 0.179 & -1.329& 0.155 &0.122 &105& 0.017,0.008\\
&3 & 3.005 & 0.171 & -1.430& 0.147 &0.118 &97& 0.017,0.007\\
&2 & 3.648 & 0.154 & -1.320& 0.162 &0.130 &84& 0.016,0.008\\
&3 & 3.456 & 0.168 & -1.306& 0.134 &0.108 &102& 0.015,0.007\\
\hline
\multirow{5}{*}{pink 4ply} &4 & 5.605 & 0.364 & -2.419& 0.249 &0.230 &49& 2\\
&4 & 4.867 & 0.400 & -2.529& 0.248 &0.222 &67& 0.021,0.008\\
&4 & 5.206 & 0.367 & -3.360& 0.264 &0.222 &91& 0.020,0.008\\
&4 & 5.381 & 0.359 & -3.160& 0.251 &0.218 &72& 0.019,0.008\\
&5 & 5.704 & 0.381 & -3.189& 0.251 &0.219 &83& 0.018,0.007\\
\hline
\multirow{5}{*}{grey thick} &2 & 3.825 & 0.254 & -2.525& 0.334 &0.232 &148& 1\\
&2 & 4.142 & 0.308 & -3.109& 0.322 &0.250 &188& 0.014,0.007\\
&2 & 3.630 & 0.284 & -3.015& 0.342 &0.274 &163& 0.016,0.008\\
&2 & 3.770 & 0.291 & -3.057& 0.326 &0.246 &173& 0.016,0.008\\
&3 & 4.126 & 0.296 & -3.328& 0.317 &0.260 &176& 0.014,0.007\\
\hline
\multirow{5}{*}{orange} &4 & 6.995 & 0.440 & -3.181& 0.309 &0.275 &52& 2\\
&5 & 6.275 & 0.447 & -3.745& 0.270 &0.246 &66& 0.021,0.008\\
&5 & 6.762 & 0.454 & -5.513& 0.299 &0.234 &70& 0.020,0.008\\
&4 & 6.932 & 0.405 & -3.506& 0.312 &0.276 &71& 0.018,0.008\\
&5 & 6.921 & 0.440 & -5.403& 0.290 &0.256 &69& 0.018,0.007\\
\hline
\multirow{5}{*}{light} &2 & 10.705 & 0.369 & -3.654& 0.380 &0.318 &93& 2\\
&3 & 8.471 & 0.477 & -5.574& 0.433 &0.380 &101& 0.017,0.008\\
&4 & 7.926 & 0.465 & -6.81& 0.369 &0.324 &101& 0.016,0.008\\
&3 & 10.433 & 0.409 & -4.501& 0.410 &0.357 &113& 0.017,0.008\\
&4 & 9.913 & 0.472 & -5.625& 0.363 &0.319 &90& 0.014,0.007\\
\hline
\end{tabular}
\label{tab_params_results}
\end{center}
\end{table*}

\begin{table*}
\begin{center}
\scriptsize
\caption{Inferred flyaway parameters for the Figure 12.}
\setlength{\tabcolsep}{5pt}
\begin{tabular}{|p{30pt}|p{7pt}|p{10pt}|p{10pt}|p{10pt}|p{15pt}|p{15pt}|p{15pt}|p{10pt}|p{15pt}|p{10pt}|}
\hline
yarn&$m$&$p$&$l_{hair}$&$\beta$&$s$&$l_{loop}$& $d _{mean}$ & $d_{std}$&$j_{xy}$& $j$ \\
\hline
red&151&0.48&2.41&0.86&0.65&4.75&7.77&2.35&0.014&0.20\\
golden&123&0.50&2.26&0.77&0.44&4.48&4.44&2.08&0.014&0.19\\
light 3ply&160&0.53&4.71&0.77&0.55&12.38&11.10&2.68&0.019&0.19\\
orange&163&0.54&3.50&1.07&0.46&5.14&7.83&1.78&0.015&0.20\\
rose &147&0.51&2.61&0.83&0.49&4.40&8.30&1.53&0.014&0.21\\
grey-blue&177&0.44&1.93&0.97&0.43&4.03&5.42&1.70&0.010&0.16\\
pink 4ply&154&0.56&3.1&0.82&0.45&4.66&4.69&2.35&0.018&0.23\\
blue&182&0.45&2.91&0.85&0.73&7.20&10.61&1.68&0.014&0.19\\
grey&200&0.42&1.72&0.85&0.36&3.22&3.38&1.52&0.015&0.17\\
yellow&183&0.35&1.78&0.88&0.52&4.14&7.79&1.45&0.011&0.16\\
pink 6ply&175&0.54&3.70&0.97&0.62&7.04&10.74&2.14&0.016&0.22\\
light 2ply&223&0.47&4.75&1.12&0.48&8.14&13.12&2.08&0.011&0.23\\
\hline
\end{tabular}
\label{tab_params_fly}
\end{center}
\end{table*}